\def\year{2022}\relax
\documentclass[letterpaper]{article} %
\usepackage{titletoc}

\usepackage{aaai22}  %
\usepackage{times}  %
\usepackage{helvet}  %
\usepackage{courier}  %
\usepackage{hyperref}       %
\usepackage{graphicx} %
\usepackage{natbib}  %
\usepackage{caption} %
\DeclareCaptionStyle{ruled}{labelfont=normalfont,labelsep=colon,strut=off} %
\usepackage{amsfonts}       %
\usepackage{nicefrac}       %
\usepackage{microtype}      %
\usepackage{xcolor}         %
\usepackage{tabularx}

\usepackage{graphicx}

\usepackage{amsmath,amssymb,amsfonts}
\usepackage[ruled,vlined]{algorithm2e}
\usepackage{amsthm}
\usepackage{array}
\usepackage{float}
\usepackage{subcaption}
\usepackage{bbding}
\usepackage{threeparttable}
\usepackage{multirow}

\usepackage{newfloat}
\usepackage{makecell}
\usepackage{listings}
\floatstyle{ruled}
\newfloat{listing}{tb}{lst}{}
\floatname{listing}{Listing}
\newtheorem{assumption}{Assumption}
\newtheorem{theorem}{Theorem}
\newtheorem{corollary}{Corollary}
\newtheorem{lemma}{Lemma}

\allowdisplaybreaks

\title{Demystifying Why Local Aggregation Helps: Convergence Analysis of\\ Hierarchical SGD}

\author {
    Jiayi Wang\textsuperscript{\rm 1},
    Shiqiang Wang\textsuperscript{\rm 2},
    Rong-Rong Chen\textsuperscript{\rm 1},
    Mingyue Ji\textsuperscript{\rm 1}
}
\affiliations {
    \textsuperscript{\rm 1} Department of Electrical \& Computer Engineering, University of Utah, Salt Lake City, UT, USA\\
    \textsuperscript{\rm 2} IBM T. J. Watson Research Center, Yorktown Heights, NY, USA\\
    jiayi.wang@utah.edu, wangshiq@us.ibm.com, rchen@ece.utah.edu, mingyue.ji@utah.edu
}

\usepackage{bibentry}

\begin{document}

\maketitle

\begin{abstract}
Hierarchical SGD (H-SGD) has emerged as a new distributed SGD algorithm for multi-level communication networks. In H-SGD, before each global aggregation, workers send their updated local models to local servers for aggregations. Despite recent research efforts, the effect of local aggregation on global convergence still lacks theoretical understanding. In this work, we first introduce a new notion of ``upward'' and ``downward'' divergences. We then use it to conduct a novel analysis to obtain a worst-case convergence upper bound for two-level H-SGD with non-IID data, non-convex objective function, and stochastic gradient. By extending this result to the case with random grouping, we observe that this convergence upper bound of H-SGD is between the upper bounds of two single-level local SGD settings, with the number of local iterations equal to the local and global update periods in H-SGD, respectively. We refer to this as the ``sandwich behavior''. Furthermore, we extend our analytical approach based on ``upward'' and ``downward'' divergences to study the convergence for the general case of H-SGD with more than two levels, where the ``sandwich behavior'' still holds. Our theoretical results provide key insights of why local aggregation can be beneficial in improving the convergence of H-SGD.
\end{abstract}

\section{Introduction}

Stochastic gradient descent (SGD) is a widely used optimization technique in machine learning applications. 
In distributed SGD, all $n$ workers collaboratively learn a global model $\mathbf{w}$ by minimizing the empirical loss with their local data:
\begin{align}
\label{eq:initialObjective}
\min_{\mathbf{w}\in \mathbb{R}^d} f(\mathbf{w}) := 
\frac{1}{n}\sum_{j=1}^n F_{j}(\mathbf{w}),    
\end{align}
where $F_j(\cdot)$ is the local loss function of worker $j$. Traditionally, workers send their models or gradients to the central server after one local iteration, which is inefficient due to frequent model aggregations. This may cost high communication latency in practice.
Recently, a form of distributed SGD was proposed to 
reduce communication cost by allowing multiple local iterations during one communication round \cite{McMahan2017a}. 
This is referred to as local SGD \cite{stich2018local,Lin2020Don't}. 
However, since data on workers can be non-IID, to avoid model divergence, the number of local iterations $P$ cannot be too large, which limits its advantage of reducing communication cost.

In practical scenarios, networks often have a hierarchical structure in nature, such as edge computing systems \cite{8270639} and software defined networks (SDN) \cite{6461195}. In these networks, workers often directly communicate with their local server rather than global server. Motivated by this hierarchical structure, a few works \cite{castiglia2020multi,liu2020client} proposed hierarchical SGD (H-SGD). In H-SGD, workers are partitioned into $N$ groups, where each group has a local server. To reduce communication cost, workers send their models to local servers to do several local aggregations before communicating with global server. They perform $I$ local iterations (referred to as local period) between local aggregations, and $G$ (referred to as global period) local iterations ($G>I$) between global aggregations.
In this paper, we call the connection between local servers and workers as ``downward'' network and the connection between local servers and the global server as ``upward'' network.

Recently, there have been a few works analyzing the convergence behavior of H-SGD.  A structure where the upward network is a peer-to-peer network while downward network is a server-worker network was considered by \citet{castiglia2020multi}. However, it only considers IID data. The work by \citet{liu2020client} considers non-IID data but it uses full-batch (non-stochastic) gradient descent and the convergence bound is an exponential function of $G$. In both works, a comprehensive comparison between H-SGD and local SGD is missing and none of them analyzes the effect of local aggregation on overcoming data heterogeneity, which is the key merit of H-SGD. Therefore, a more general and tighter analysis for H-SGD with non-convex objective function, non-IID data and stochastic gradient descent is needed. The effect of local aggregation on overcoming data heterogeneity lacks theoretical understanding.  

\begin{table*}[t]
\centering
\caption{A summary of convergence bounds in the literature.}
\label{tab: comparison}
\small
\renewcommand{\arraystretch}{2}
\begin{tabular}{
>{\centering\arraybackslash}p{0.25\linewidth}|
>{\centering\arraybackslash}p{0.36\linewidth} | >{\centering\arraybackslash}p{0.05\linewidth} |
>{\centering\arraybackslash}p{0.03\linewidth} |
>{\centering\arraybackslash}p{0.08\linewidth} |
>{\centering\arraybackslash}p{0.08\linewidth}}
\hline
Paper&Convergence Bound& \!\!\!Non-IID\!\!\! & \!\!SGD\!\! & Type & %
\!\!\!Assumption\!\!\!\\ 
\hline
\citet{yu2019linear}& $O\left(\frac{1+\sigma^2}{\sqrt{nT}}+\frac{n}{T}(P\sigma^2+P^2\tilde{\epsilon}^2) \right)$ & \CheckmarkBold &\CheckmarkBold & \!\!\! Local SGD \!\!\!& $N=1$  \\ 
 \hline
\citet{liu2020client} & $O\left(\frac{1+B^G\tilde{\epsilon}^2}{\sqrt{nT}}\right)$ & \CheckmarkBold & \XSolidBrush & H-SGD & $\sigma^2=0$ \\ 
 \hline
\!\!\!\citet{castiglia2020multi}\!\!\!  &   $O\left(\frac{1+\sigma^2}{\sqrt{nT}}+ \frac{n}{T}\frac{G^2}{I}\sigma^2\right)$ & \XSolidBrush & \CheckmarkBold &  H-SGD & $\tilde{\epsilon}^2=0$ \\ 
 \hline
 Ours  &
$
O\left(\frac{1+\sigma^2}{\sqrt{nT}}+ \frac{(N-1)(G\sigma^2+G^2\tilde{\epsilon}^2) +(n-N)(I\sigma^2 + I^2\tilde{\epsilon}^2)}{T}\right)
    $
 & \CheckmarkBold & \CheckmarkBold &  H-SGD & None  \\
\hline
\end{tabular}
 \begin{tablenotes}
        \footnotesize
        \item \textsuperscript{1} $P$: aggregation period in local SGD; $G$: global aggregation period in H-SGD; $I$: local aggregation period in H-SGD; $N$: number of groups in H-SGD; $\tilde{\epsilon}^2$: global divergence (Assumption~\ref{assumption:divergence-sf-sgd}); $\sigma^2$: stochastic gradient noise
        \item \textsuperscript{2} $B$ is a constant and $B>2$. $n$ is the number of nodes and $N$ is the number of groups. $T$ is the total number of local iterations. 
        \item \textsuperscript{3} Our bound can reduce to the IID case by setting $\tilde{\epsilon}^2=0$. 
 \end{tablenotes}
\end{table*}

In this paper, to provide a better theoretical understanding to H-SGD, 
we devise a novel characterization of the data heterogeneity of H-SGD by ``upward'' divergence and ``downward'' divergence, respectively. Furthermore, we show that the global divergence can be partitioned into upward and downward divergences. 
H-SGD can be seen as performing distributed SGD both within a group and across groups. Within a group, workers perform multiple local iterations (on each worker) before local aggregation (within each group). Across groups, each group performs multiple local aggregations  
before global aggregation.
With this characterization, we conduct a novel convergence analysis for H-SGD with non-IID data, non-convex objective function and stochastic gradients. 
Furthermore,
we show that although data can be highly non-IID, local aggregation can help global convergence even when grouping is random.
Then by a detailed comparison with local SGD, we show that our convergence upper bound for H-SGD lies in between the convergence upper bounds of two single-level local SGD settings with aggregation periods of $I$ and $G$, respectively. This is referred to as the ``sandwich'' behavior, which reveals the fundamental impact of local aggregation.

Our convergence analysis shows that better convergence of H-SGD can be achieved with local aggregation when the number of groups, together with the global and local periods, are chosen appropriately to control the combined effect of upward  and downward divergences. 
In general, we show that since local aggregation is more frequent, grouping strategies with a smaller upward divergence can strengthen the benefit of local aggregation. To reduce the communication cost while maintaining similar or better convergence, it can be beneficial to increase the global period $G$ and decrease the local period $I$.
We also extend our results to multi-level cases where there are multiple levels of local servers, where our characterization with upward and downward divergences can be applied to each level, from which we derive the convergence bound for general multi-level cases.

A comparison of our result with existing results is shown in Table~\ref{tab: comparison}.
When setting $N=1$ and $P=I=G$, our result recovers the well-known result for single-level local SGD by \citet{yu2019linear}. We can also see that choosing $I<G=P$ for H-SGD gives a smaller convergence upper bound, which shows the benefit of the hierarchical structure.
The result by \citet{liu2020client} only considers full gradient descent. In this case, the stochastic noise $\sigma^2 = 0$. Even when setting $\sigma^2=0$, our result is tighter. The result by \citet{castiglia2020multi} only considers IID case where the global divergence $\tilde{\epsilon}^2 =0$. Setting $\tilde{\epsilon}^2=0$, our result is still tighter than theirs since $I<G$. It can be seen that our result is the most general and tightest.

\textbf{Main Contributions.} 
\begin{itemize}
    \item We introduce the new notion of ``upward'' and ``downward'' divergences to characterize data heterogeneity of H-SGD. We show that it can be extended %
    to multi-level cases.
    \item We derive a general convergence bound for two-level H-SGD with non-IID data, non-convex objective functions and stochastic gradient descent.
    \item We provide a novel convergence analysis for random grouping and show how local aggregation helps global convergence by a ``sandwich'' behavior.  
    \item We extend our analysis to multi-level H-SGD and the result shows similar properties as the two-level case. To our knowledge, this is the first analysis which can be extended to multi-level cases. 
\end{itemize}
We also conduct experiments on CIFAR-10, FEMNIST, and CelebA datasets. The results of experiments validate our theoretical results. 

\section{Related Works}
Traditional distributed SGD is introduced by \citet{zinkevich2010parallelized}, which can be seen as a special case of local SGD with only one local iteration during one communication round.  
There have been a large volume of works analyzing the convergence of local SGD, for convex objective functions \citep{Li2020On,WangJSAC2019}, non-convex objective functions \citep{NEURIPS2019_c17028c9,YuAAAI2019}, and their variants \citep{karimireddy2020scaffold,li2019federated,reddi2020adaptive,CooperativeSGD,yu2019linear}.
Local SGD can be regarded as a special case of H-SGD with only one level.
Our theoretical results for H-SGD recover results for both local SGD and traditional distributed SGD. There have been a few works analyzing the convergence of H-SGD, including \citet{castiglia2020multi,zhou2019distributed} for IID data, and \citet{liu2020client} for non-IID data with full-batch gradient descent, as described earlier. There are also works on system design for H-SGD without theoretical guarantees \citep{abad2020hierarchical,9127160}. %
In addition, 
there are works on decentralized SGD \citep{CooperativeSGD,bellet2021d}, where workers exchange their models based on a doubly stochastic mixing matrix. However, the analysis for decentralized SGD cannot be applied to H-SGD, since the second largest eigenvalue of the mixing matrix in the case of H-SGD is one and the decentralized SGD analysis requires this second largest eigenvalue to be strictly less than one.

There are also some works focusing on practical aspects such as model compression and sparsification \citep{han2020adaptive,Peng2018,jiang2019model,Konecny2016} and partial worker participation \citep{Bonawitz2019,chen2020fedcluster}.
These algorithms and techniques are orthogonal to our work and may be applied together with H-SGD.

\section{H-SGD Setup}
In two-level H-SGD, all workers are partitioned into $N$ groups $\mathcal{V}_1,\mathcal{V}_2,\ldots ,\mathcal{V}_N$. The number of workers in each group is denoted by $n_i := |\mathcal{V}_i|$ ($i=1,2,\ldots ,N$). Then we have $n = \sum_{i=1}^N n_i$.
With this grouping, the objective function (\ref{eq:initialObjective}) is equivalent to
\begin{align}\label{eq:objective}
\min_{\mathbf{w}\in \mathbb{R}^d} f(\mathbf{w}) := 
\sum_{i=1}^N \frac{n_i}{n} f_{i}(\mathbf{w}), 
\end{align}
where 
$f_i(\cdot)$ is the averaged loss function of workers in group $i$ that is defined as follows:
\begin{align}
f_i(\mathbf{w}):=\frac{1}{n_i}\sum_{j \in \mathcal{V}_i} F_{j}(\mathbf{w}).    
\end{align}
During each local iteration $t$, each worker $j$ updates its own model using SGD:
\begin{align}
\mathbf{w}_{j}^{t+1} = \mathbf{w}_{j}^t - \gamma \mathbf{g}(\mathbf{w}_{j}^t, \zeta^t_{j}), 	
\end{align} 
where $\gamma$ is the learning rate, $\mathbf{g}(\mathbf{w}_{j}^t, \zeta^t_{j})$ is the stochastic gradient of $F_{j}(\mathbf{w})$, and $\zeta^t_{j}$ represents random data samples from the local dataset $\mathcal{D}_{j}$ at worker $j$. 
We assume that $\mathbb{E}_{\zeta^t_{j}\sim \mathcal{D}_{j}}[\mathbf{g}(\mathbf{w}_{j}^t, \zeta^t_{j})] = \nabla F_{j}(\mathbf{w}_{j}^t)$. 

During one communication round, local models are first averaged within group $i$ ($i = 1, 2,\ldots , N$) after every $I_i$ local iterations.
In particular, at local iteration $t \in \{I_i,2I_i,3I_i,\ldots \}$, we compute $\bar{\mathbf{w}}_i^{t}:=\frac{1}{n_i} \sum_{j \in \mathcal{V}_i} \mathbf{w}_{j}^{t}$. This can be done at a \textit{local server} (e.g., a computational component in close proximity) for group $i$. 
After several rounds of ``intra-group'' aggregations, models from the $N$ groups are averaged globally. Let a global aggregation be performed for every  $G$ local iterations.
Then, at local iteration $t$, we have $\bar{\mathbf{w}}^{t}:=\frac{1}{N} \sum_{i=1}^{N} \bar{\mathbf{w}}_i^{t}$ for $t=G,2G,3G,\ldots $. 
Note that we assume that all workers perform synchronous updates and let $G$ be a common multiple of $\{I_1, \ldots , I_N\}$. Therefore, distributed SGD is conducted both at the local level within a group  and the global level across groups.  We summarize the algorithm in Algorithm~\ref{alg:HF-SGD}, where $a\mid b$ (or $a\nmid b$) denotes that $a$ divides (or does not divide) $b$, i.e., $b$ is (or is not) an integer multiple of $a$.

In order to understand the fundamental convergence behavior of H-SGD, we will mainly focus on the two-level model introduced above. We will extend our analysis to %
more than two levels in a later section.

\begin{algorithm}[t]
\small
 \caption{Hierarchical SGD (H-SGD)}
 \label{alg:HF-SGD}
\KwIn{$\gamma$, $\bar{\mathbf{w}}^0$, $G$, $\{\mathcal{V}_i : i\in \{1,2,\ldots ,N\}\}$, $\{I_i : i\in \{1,2,\ldots ,N\}\}$}
\KwOut{Global aggregated model $\mathbf{\bar w}^{T}$}
\For{$t=0$ to $T-1$}{
\For{Each group $i\in \{1,2,\ldots ,N\}$, in parallel}{
\For{Each worker $j \in \mathcal{V}_i$, in parallel}{
Compute $\mathbf{g}(\mathbf{w}_{j}^t,\zeta^t_{j})$;\\
$\mathbf{w}_{j}^{t+1} \leftarrow \mathbf{w}_{j}^t - \gamma \mathbf{g}(\mathbf{w}_{j}^t,\zeta^t_{j})$;\\
}
\If{$I_i \mid t+1$}{
Local aggregate: $\mathbf{\bar w}_i^{t+1} \leftarrow \frac{1}{n_i} \sum_{j \in \mathcal{V}_i} \mathbf{w}_{j}^{t+1}$; \\
\If{$G \nmid t+1$}{
    Distribute: $\mathbf{w}_{j}^{t+1} \leftarrow \mathbf{\bar w}_i^{t+1}, \forall j\in\mathcal{V}_i$;
}
}
}
\If{$G \mid t+1$}{
Global aggregate: $\mathbf{\bar w}^{t+1} \leftarrow \frac{1}{N} \sum^N_{i=1} \mathbf{\bar w}_i^{t+1} $;\\
Distribute: $\mathbf{w}_{j}^{t+1} \leftarrow \mathbf{\bar w}^{t+1}, \forall j\in\mathcal{V}$;
}
}
\end{algorithm}

\section{Convergence Analysis for Two-level H-SGD}
\label{sec: Convergence Analysis}
We begin with a description of the assumptions made in our convergence analysis. Then, we present our results for general two-level H-SGD with fixed groupings. The number of workers and the number of local period during one communication round can vary among groups. 

\subsection{Assumptions}
We make the following minimal set of assumptions that are common in the literature.

\begin{assumption}
\label{assumption:hf-sgd}
For H-SGD, we assume the following.

a) \textbf{Lipschitz gradient} 
\begin{align}
\|\nabla F_{j}(\mathbf{w}) - \nabla F_{j}(\mathbf{w}')\| \le L\|\mathbf{w}-\mathbf{w}' \|, \forall j, \mathbf{w}, \mathbf{w}'
\end{align}
where $L$ is some positive constant. 

b) \textbf{Bounded variance}
\begin{align}
\mathbb{E}_{\zeta^t_j\sim \mathcal{D}_{j} }\|\mathbf{g}(\mathbf{w};\zeta^t_j)-\nabla F_{j}(\mathbf{w})\|^2 \le \sigma^2, \forall j,\mathbf{w}
\end{align}
where $\mathcal{D}_{j}$ is the dataset at worker $j\in\mathcal{V}$.

c) \textbf{Bounded upward divergence (H-SGD)}
\begin{align}
\label{eq: div global}
	\sum_{i=1}^N\frac{n_i}{n}\|\nabla f_i(\mathbf{w}) - \nabla f(\mathbf{w})\|^2 \le \epsilon^2, \forall \mathbf{w} 
\end{align}

d) \textbf{Bounded downward divergence (H-SGD)}
\begin{align}
\label{eq: bounded second-level divergence}
		\frac{1}{n_i}\sum_{j\in \mathcal{V}_i}\|\nabla F_{j}(\mathbf{w}) - \nabla f_i(\mathbf{w})\|^2 \le \epsilon_i^2,\forall i, \mathbf{w} 
\end{align}  
\end{assumption}

Note that the Lipschitz gradient assumption also applies to the group objective $f_i(\mathbf{w})$ and overall objective $f(\mathbf{w})$. For example, $\|\nabla f_i(\mathbf{w}) - \nabla f_i(\mathbf{w}')\| 
= \|\frac{1}{n_i}\sum_{j\in \mathcal{V}_i}\nabla F_{j}(\mathbf{w}) -\frac{1}{n_i}\sum_{j\in \mathcal{V}_i}\nabla F_{j}(\mathbf{w}') \| 
\le  \frac{1}{n_i} \sum_{j\in \mathcal{V}_i}\|\nabla F_{j}(\mathbf{w}) - \nabla F_{j}(\mathbf{w}')\| 
\le  L\|\mathbf{w}-\mathbf{w}'\|$. Global divergence is often used to describe the data heterogeneity of single-level local SGD \cite{yu2019linear}, which is as the following.
\begin{assumption}{\textbf{Bounded global divergence}}
\label{assumption:divergence-sf-sgd}
\begin{align}
 \frac{1}{n}\sum_{j=1}^n\|\nabla F_{j}(\mathbf{w}) - \nabla f (\mathbf{w}) \|^2 \le \tilde{\epsilon}^2 ,\forall \mathbf{w}.
\label{eq:div_SF}
\end{align}

\end{assumption} 
Note that when $N=1$, H-SGD reduces to local SGD and downward divergence becomes the global divergence while upward divergence becomes zero. When $N=n$, H-SGD also reduces to local SGD but upward divergence becomes the global divergence while downward divergence becomes zero. Here, we will use global divergence to show our results for H-SGD can encompass the results for local SGD. 

Note for the global divergence, it is easy to show that
\begin{align}\label{eq:divergence partition}
\frac{1}{n}\sum_{j=1}^n &\|\nabla F_{j}(\mathbf{w}) \!-\! \nabla f (\mathbf{w}) \|^2 =\sum_{i=1}^N \frac{n_i}{n}\|\nabla f_i(\mathbf{w}) \!-\! \nabla f(\mathbf{w})\|^2 \notag \\
&+ \sum_{i=1}^N\frac{n_i}{n} \frac{1}{n_i}\sum_{j\in \mathcal{V}_i}\|\nabla F_{j}(\mathbf{w}) - \nabla f_i(\mathbf{w})\|^2.
\end{align}
In (\ref{eq:divergence partition}), we see that the upward and downward divergences are in fact a partition of the global divergence, which implies that upward and downward divergences do not increase at the same time. It is the hierarchical structure that makes this partition possible. Later we will show that this partition can exactly explain the benefits of local aggregation. We will discuss more about the relationship between upward/downward  and global divergences in a later section.

\subsection{Convergence Analysis}
\textbf{Technical Challenge.} The main challenge of the analysis is that workers only perform local iterations before global aggregation in local SGD, whereas in H-SGD, workers not only perform local iterations but also do aggregations with other workers in the same group. If we directly apply the analysis for local SGD in upward part, the effect of local aggregations would be neglected and the resulted bound would be loose. To address this, our key idea is to construct the local parameter drift $\|\mathbf{w}_j - \bar{\mathbf{w}}_i \|, j \in \mathcal{V}_i$ in the analysis of global parameter drift $\|\bar{\mathbf{w}}_i - \mathbf{\bar{w}}\|$, so that the analysis for the downward part can be incorporated in the analysis for the upward part.
\begin{theorem}\label{thm1}
Consider the problem in (\ref{eq:objective}). For any fixed worker grouping that satisfies Assumption~\ref{assumption:hf-sgd}, if  the learning rate in Algorithm $1$ satisfies $\gamma < \frac{1}{2\sqrt{6}GL}$, 
then for any $T\ge 1$,
we have 
\begin{subequations}
\begin{align}
&\frac{1}{T} \sum_{t=0}^{T-1}\mathbb{E}\|\nabla f(\mathbf{\bar w}^t) \|^2 \le \frac{2(f^0-f^*)}{\gamma T} +  \gamma L\frac{1}{n}\sigma^2\label{eq1a} \\
&\quad +2C\gamma^2L^2G\frac{N-1}{n}\sigma^2
+3C\gamma^2L^2G^2\epsilon^2
 \label{eq1b1} \\
&\quad +2C\gamma^2L^2\sigma^2 \sum_{i=1}^N \frac{(n_i-1)I_i}{n} + 3C\gamma^2 L^2\sum_{i=1}^N\frac{n_i}{n} I_i^2\epsilon_i^2 , 
\label{eq1d}
\end{align}
\end{subequations}
where $C=40/3$. 
\end{theorem} 

\textbf{Remark 1.}
Note that the bound (\ref{eq1a})--(\ref{eq1d}) can be partitioned into three parts. The terms in (\ref{eq1a}) are the original SGD part. If we set $N=n_i=1$ ($\epsilon$ and $\epsilon_i$ become zero), then only (\ref{eq1a}) will remain, which is the same as convergence bound for non-convex SGD in \cite{bottou2018optimization}. The terms in (\ref{eq1b1}) are the upward part, which consists of noise and upward divergence associated with $G$. The terms in (\ref{eq1d}) are the downward part, which has a similar form as (\ref{eq1b1}) associated with $I_i$. Divergence plays a more important role than noise since the coefficients in front of the SGD noise $\sigma^2$ include $G$ and $I_i$, while the coefficients in front of the divergences $\epsilon$ and $\epsilon_i$ include $G^2$ and $I_i^2$. Since $G>I_i,\forall i$, the upward part has a stronger influence on convergence.

\textbf{Remark 2.}
Note that all the divergences of H-SGD can be written as $O(\gamma^2G^2\epsilon^2 + \gamma^2 \sum_{i=1}^N \frac{n_i}{n} I_i^2 \epsilon_i^2)$  while the corresponding part in local SGD is $O(\gamma^2 G^2 \tilde{\epsilon}^2)$ \cite{yu2019linear}. This exactly shows how local aggregation overcomes data heterogeneity: global divergence is partitioned into two parts and local aggregation weakens the effects of the downward part. This also brings us some insights on how to group workers. %
The grouping strategy should have the smallest upward divergence since this can make full use of benefits of local aggregation. We will validate this property in the experiments.

\textbf{Remark 3.}
Let $\gamma = \sqrt{\frac{n}{T}} $ with $T\ge \frac{1}{24G^2L^2\sqrt{n}}$, when $T$ is sufficiently large, $\frac{1}{T} \sum_{t=0}^{T-1}\mathbb{E}\|\nabla f(\mathbf{\bar w}^t) \|^2 
= O\left(\sqrt{\frac{1}{nT}}\right)+ O\left(\frac{1}{T}\right)$,
which achieves a linear speedup in $n$.

In the following corollary, we show that our results can encompass local SGD with local period $P$.
\begin{corollary}\label{corollary:single-level}
\textbf{(Degenerate to local SGD)}
Let $N=1$ and $\gamma \le \frac{1}{2\sqrt{6}PL}$, 
from Theorem~\ref{thm1}, we obtain 
\begin{align}
&\frac{1}{T} \sum_{t=0}^{T-1}\mathbb{E}\|\nabla f(\mathbf{\bar w}^t) \|^2 \le \frac{2(f^0-f^*)}{\gamma T} + \frac{\gamma L\sigma^2}{n} \notag\\
&\quad + 2C\gamma^2L^2\sigma^2 \left(1-\frac{1}{n}\right)P
+  3C\gamma^2L^2 P^2 \tilde{\epsilon}^2 \notag \\
&= O\left(\frac{1}{\gamma T}\right) + O\left(\frac{\gamma \sigma^2}{n}\right) \notag\\
&\quad + O\left(\gamma^2 P \sigma^2\left(1-\frac{1}{n}\right)\right)+ O\left(\gamma^2 P^2 \tilde{\epsilon}^2\right).  
\label{eq: SF_SGD 1}
\end{align}
\end{corollary} 
While (\ref{eq: SF_SGD 1}) is similar to the bound  by \citet{yu2019linear}, %
 we note that the third term $O\left(\gamma^2 P \sigma^2\left(1-\frac{1}{n}\right)\right) $ in the last equality in (\ref{eq: SF_SGD 1}) has an additional term of $\left(1-\frac{1}{n}\right)$ compared to the bound by \citet{yu2019linear}. This term can potentially make our bound tighter. Another important observation is that the techniques used to obtain this term is the key to make our H-SGD bound in Theorem~\ref{thm1} encompass original SGD cases.

\subsection{H-SGD with Random Grouping}
\label{sec: HF-SGD random grouping}
We now state our convergence results for H-SGD with random grouping. 
We will show that the convergence upper bound of H-SGD with random grouping takes a value that is between the convergence upper bounds of two single-level local SGD settings with local and global periods of $I$ and $G$, respectively. This will provide insights on when and why local aggregation helps.

For worker grouping, we consider all possible grouping strategies with the constraint that $n_i = n/N$, $\forall i$. 
Then, we uniformly select one grouping strategy at random. 
Let the random variable $\mathsf{S}$ denote the uniformly random grouping strategy. This means that each realization of $\mathsf{S}$ corresponds to one grouping realization. First, we introduce two key lemmas for our convergence analysis.

\begin{lemma} \label{lemma1}
Using the uniformly random grouping strategy $\mathsf{S}$, for any $\mathbf{w}$, the average upward divergence is %
\begin{align}
\label{eq: lemma 1}
\mathbb{E}_{\mathsf{S}} \left[\frac{1}{N}\sum_{i=1}^N \left\| \nabla f(\mathbf{w}) \!-\! \nabla f_i(\mathbf{w}) \right\|^2\right] 
\le  \left(\frac{N\!-\!1}{n\!-\!1}\right) \tilde{\epsilon}^2,
\end{align}
where $\tilde{\epsilon}$ is given in (\ref{eq:div_SF}).
\end{lemma}

\begin{lemma} \label{lemma2}
Using the uniformly random grouping strategy $\mathsf{S}$, for any $\mathbf{w}$, the average downward divergence is %
\begin{align}
\label{eq: lemma 2}
\mathbb{E}_{\mathsf{S}}\left[\frac{1}{n}\sum_{i=1}^N \sum_{\mathsf{k} \in \mathcal{V}_i} \left\|\nabla f_i(\mathbf{w}) \!-\! \nabla F_{\mathsf{k}}(\mathbf{w})\right\|^2 \right] \le \left(1\!-\!\frac{N\!-\!1}{n\!-\!1}\right)\tilde{\epsilon}^2.
\end{align}
\end{lemma}

Similar to (\ref{eq:divergence partition}), Lemma~\ref{lemma1} and Lemma~\ref{lemma2} show that the sum of upper bounds for upward and downward divergences is equal to the global divergence. Furthermore, when the number of groups increases, upward divergence becomes larger while downward divergence becomes smaller. When grouping is random, $N$ is the key parameter that determines how global divergence is partitioned. For simplicity, we let each group have the same local period $I_i =I, \forall i$. Then, we can obtain the following theorem. 

\begin{theorem}
\label{thm2}
Using the uniformly random grouping strategy $\mathsf{S}$, let $\gamma \le \frac{1}{2\sqrt{6}GL}$, then we have  
\begin{align}
&\mathbb{E}_{\mathsf{S}}\left[\frac{1}{T} \sum_{t=0}^{T-1} \mathbb{E}\|\nabla f(\mathbf{\bar w}^t) \|^2\right]
\le \frac{2(f^0-f^*)}{\gamma T} + \frac{\gamma L\sigma^2}{n}  \notag\\
&\quad+2C\gamma^2L^2\left[\left(\frac{N-1}{n}\right)G+\left(1-\frac{N}{n}\right)I\right]\sigma^2 \notag\\
&\quad+3C \gamma^2L^2\!\left[\!\left(\frac{N\!-\!1}{n\!-\!1}\right)\!G^2\! +\! \left(\!1-\frac{N\!-\!1}{n\!-\!1}\right)\!I^2\right] \tilde{\epsilon}^2,
\label{eq:thm2}
\end{align}
where $C = 40/3$. 
\end{theorem}
\textbf{Remark 4.} Note that both multiplicative factors of the noise term with $\sigma^2$ and the divergence term with $\tilde{\epsilon}^2$ are composed of two parts, where the upward part is ``modulated'' by $G$ while the downward part is ``modulated'' by $I$. As $N$ becomes larger, the upward part has a stronger influence on the convergence bound since both $\frac{N-1}{n}$ and $\frac{N-1}{n-1}$ become larger. Note that the convergence upper bound of H-SGD can be sandwiched by the convergence upper bounds of two local SGD. To see this, 
we consider the following three scenarios: 1) local SGD with    aggregation period $P=G$, 2) local SGD with aggregation period $P=I$, and 3) H-SGD with local aggregation period $I$ and global aggregation period $G$. 
We let them all start with the same $\mathbf{\bar w}^0 $ and choose learning rate $\gamma$ such that $\gamma \le \frac{1}{2\sqrt{6}GL}$. We can see that all these three cases have the same first two terms in (\ref{eq:thm2}). However, for the third and fourth terms in (\ref{eq:thm2}), we have
\begin{align}
\left(1-\frac{1}{n}\right)&I 
\le \left(\frac{N\!-\! 1}{n}\right)G + \left(1-\frac{N}{n}\right)I 
\le \left(1-\frac{1}{n}\right)G,    \label{eq: remark 21} \\
I^2  \le &\left(\frac{N-1}{n-1}\right)G^2 + \left(1-\frac{N-1}{n-1}\right)I^2 \le G^2.  \label{eq: remark 22}
\end{align}
Equations~(\ref{eq: remark 21}) and (\ref{eq: remark 22}) show that the convergence upper bound of H-SGD has a value between the convergence upper bounds of local SGD with aggregation periods of $P=I$ and $P=G$, respectively. The grouping approach can explicitly characterize how much the convergence bound moves towards the best case, i.e., local SGD with $P=I$. However, %
this case incurs the highest global communication cost, so grouping can adjust the trade-off between convergence and communication cost.

\textbf{Remark 5.} Even with a large $G$, if we make $I$ sufficiently small, the convergence bound of H-SGD can be improved. To see this, consider the last two terms of (\ref{eq:thm2}). 
Suppose $G = mI, m=1,2, \cdots$. For a non-trivial worker grouping, i.e., $1<N<n$, if we increase $G$ to $G' = lG, 1<l<\sqrt{\frac{1}{m^2}\frac{n-N}{N}+1}$ and decrease $I$ to $I' = q I, q\le \sqrt{1-m^2(l^2-1)\frac{N}{n-N}}$, then one can  show that the bound (\ref{eq:thm2}) using $G'$ and $I'$ can be lower than %
that using $G$ and $I$. 
A similar behavior can also be seen for fixed grouping, as shown empirically in Figure~\ref{fig:largeG} in the experiments.

\section{H-SGD with More Than Two Levels}
\label{sec:multi-level}
\begin{figure}[t]
\centering
\includegraphics[width=1\columnwidth]{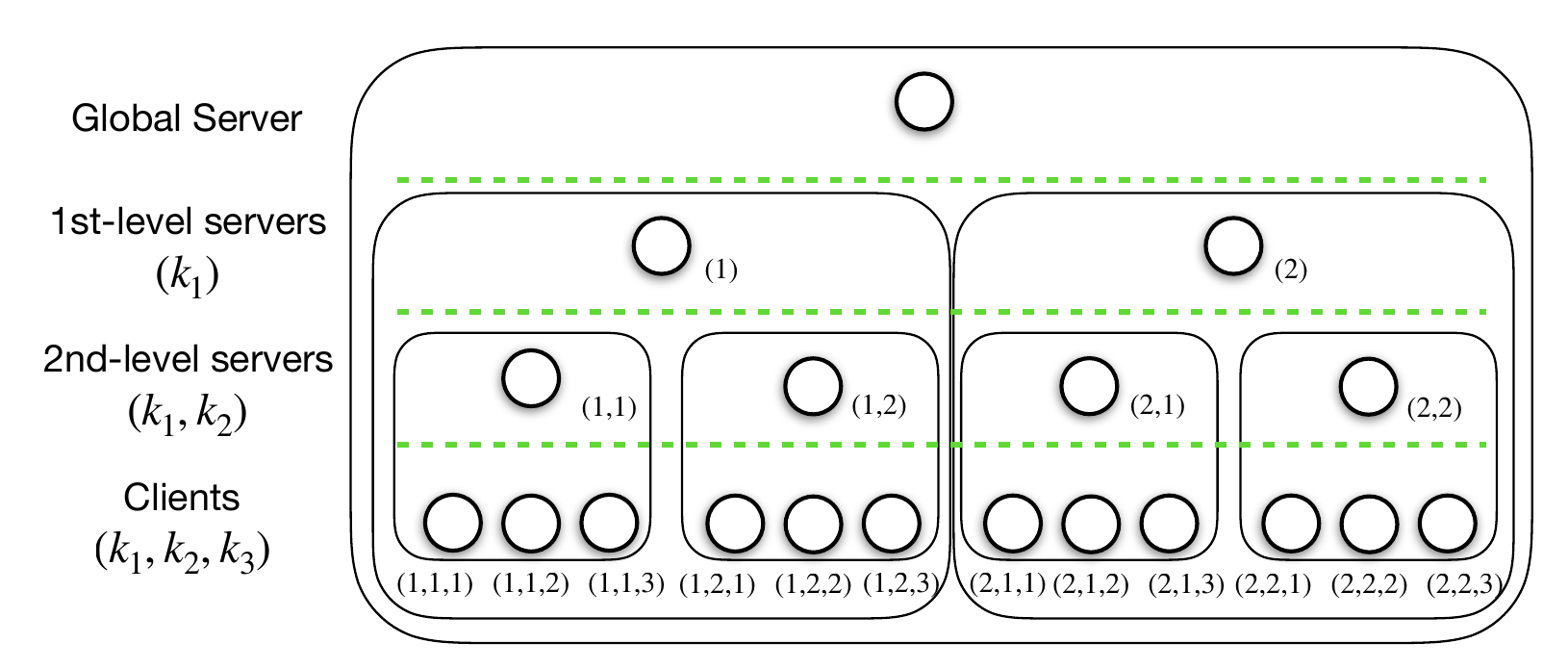}
\caption{{Three-level example with $N_1=N_2=2, N_3=3$.
}}
\label{fig:3-level-case}
\end{figure}

In the two-level H-SGD setting discussed in the previous sections, there is only one level of local servers between the global server and workers. 
In this section, we  extend our analysis to H-SGD with more than one level of local servers. 
Specifically, as shown in Figure~\ref{fig:3-level-case}, we consider a total of $M \geq 2$ levels.
For $\ell=1,2,\ldots ,M-1$, each server at level $\ell-1$ is connected to $N_\ell$ servers at level $\ell$, where we assume that the global server is at a ``dummy'' level $\ell=0$ for convenience. 
For $\ell = M$, each server at level $M-1$ directly connects to $N_M$ workers. 
As a result, we have a total of $n = \Pi_{\ell=1}^M N_\ell$ workers. 
With this notation, a sequence of indices $(k_1, k_2, \ldots , k_\ell)$ denotes a ``path'' from the global server at level $0$ to a local server/worker at level $\ell$, where this path traverses the $k_1$-th server at level $1$, the $k_2$-th server at level $2$ connected to the $k_1$-th server at level $1$, and so on. Hence, the index sequence $(k_1, k_2, \ldots , k_\ell)$ uniquely determines servers (and workers if $\ell=M$) down to level $\ell$. 

We assume in the multi-level case that each server at level $\ell -1$ connects to the same number ($N_\ell$) of servers/workers in the next level $\ell$, for ease of presentation. Our results can be extended to the general case with different group sizes. 

Let $F_{k_1\ldots  k_M}(\mathbf{w})$  denote the objective function of worker $k_1  \ldots  k_M$ and $f_{k_1 \ldots  k_\ell}(\mathbf{w})$ denote the averaged objective function of workers connected to server $k_1 \ldots  k_\ell$,
\begin{align}
\!\!\!\!f_{k_1 \ldots  k_\ell}(\mathbf{w})\! :=\! \frac{1}{\prod_{j=\ell+1}^M N_j}\sum_{k_{\ell+1}=1}^{N_{\ell+1}}\!\!\!\cdots\!\!\!\sum_{k_M=1}^{N_M} \!\!\!F_{k_1 \ldots  k_M}(\mathbf{w}).  \!  
\end{align}
Then the global objective function can be rewritten as
\begin{align}
\label{eq: f}
f(\mathbf{w}) &= \frac{1}{\prod_{j=1}^M N_j}\sum_{k_1=1}^{N_1}\cdots\sum_{k_M=1}^{N_M} F_{k_1 \ldots  k_M}(\mathbf{w})    \notag\\
&= \frac{1}{\prod_{j=1}^\ell N_j}\sum_{k_1=1}^{N_1}\cdots\sum_{k_\ell=1}^{N_{\ell}} f_{k_1 \ldots  k_\ell}(\mathbf{w}).  
\end{align}
Let $P_{\ell}$ ($\ell=1,2,\ldots ,M$) denote the period (expressed as the number of local iterations) 
that the parameters at servers at level $\ell$ are aggregated by their parent server at level $\ell - 1$. 
We require that $P_1\!>\!P_2\!>\!\ldots \!>\!P_M$ and $P_{\ell}$ is an integer multiple of $P_{\ell-1}$. The algorithm is in Appendix~\ref{appendix:MultiLevel}.

In the following, we provide  results for the random grouping case. Results for the fixed grouping case can be found in Appendix~\ref{appendix:MultiLevel}.
We apply the divergence partition idea to each level, so we extend Lemma~\ref{lemma1} and Lemma~\ref{lemma2} to obtain the following lemma. 
\begin{lemma}
\label{lemma3}
Using the uniformly random grouping strategy $\mathsf{S}$, the $\ell$-th level averaged upward and downward divergences are given by
\begin{align}
&\mathbb{E}_\mathsf{S} \bigg[ \frac{1}{n_\ell}\!\sum_{k_1=1}^{N_1}\!\!\!\!\cdots\!\!\!\sum_{k_\ell=1}^{N_\ell}\!\!\|\nabla f_{k_1 \ldots  k_\ell}(\mathbf{w})\!-\!\nabla f(\mathbf{w}) \|^2\bigg] \le\! \bigg(\frac{n_\ell-1}{n-1} \bigg)\tilde{\epsilon},\\
&\mathbb{E}_\mathsf{S} \bigg[\frac{n_\ell}{n}\!\sum_{k_{\ell+1}=1}^{N_{\ell+1}}\!\!\!\!\cdots\!\!\!\sum_{k_M=1}^{N_M}\!\!\!\|\nabla f_{k_1 \ldots  k_\ell}(\mathbf{w})\!-\!\nabla F_{k_1 \ldots  k_M}(\mathbf{w}) \|^2 \bigg]\notag\\
&\le\! \bigg(1-\frac{n_\ell-1}{n-1} \bigg)\tilde{\epsilon}^2 \label{eq:multi_level_local_divergence}
\end{align}
respectively, 
$\forall\mathbf{w}$, $\forall k_1,\ldots ,k_\ell$ in (\ref{eq:multi_level_local_divergence}), where $n_\ell = \Pi_{j=1}^\ell N_j$ and $\tilde{\epsilon}$ is the global divergence. 
\end{lemma}
When the number of servers in the $\ell$th-level $n_\ell$ increases, the $\ell$th-level's upward divergence becomes larger while its downward divergence becomes smaller. The sum of the $\ell$th-level upward and downward divergences are upper bounded by the global divergence. Based on this lemma, we derive the convergence bound for multi-level case as the following. 
\begin{theorem}
\label{thm3}
Consider uniform random grouping strategy $\mathsf{S}$, if $\gamma\le \frac{1}{2\sqrt{6}P_1L}$, then for multi-level case,
\begin{align}
\label{eq: coro 3}
&\frac{1}{T}\sum_{t=0}^{T-1} \mathbb{E}_{\mathsf{S}} \bigg\|\nabla f(\bar{\mathbf{w}}^t) \bigg\|^2 
\le  \frac{2}{\gamma T}\bigg(f^0-f^* \bigg) + \frac{\gamma L \sigma^2}{n} \notag\\
&\quad + C\gamma^2L^2\frac{1}{M-1}\! \cdot \!\sum_{\ell=1}^{M-1}\left[2A_1(\ell)\sigma^2 +3A_2(\ell)\tilde{\epsilon}^2\right]    
\end{align}
where, $A_1(\ell) := P_1\Big(\frac{1}{\Pi_{j=\ell}^M N_j}-\frac{1}{n}\Big) 
+P_{\ell}\Big(1-\frac{1}{\Pi_{j=\ell}^M N_j}\Big)$, $A_2(\ell) := P_1^2\Big(\frac{n_\ell - 1}{n-1}\Big)+P_{\ell}^2\Big(1-\frac{n_\ell - 1}{n-1}\Big)$. 
\end{theorem}
\textbf{Remark 6.} Note that when $M=2$, this bound reduces to the bound given in Theorem~\ref{thm2}. The effect of the $\ell$-th level can be represented as $2A_1(\ell)\sigma^2 +3A_2(\ell)\tilde{\epsilon}^2$, where each $A_i(\ell), i=1,2$ is composed of an upward part ``modulated'' by $P_1$ and a downward part ``modulated" by $P_\ell$. %
Similar to 
(\ref{eq: remark 21}) and (\ref{eq: remark 22}), we can obtain the sandwich behavior as:
\begin{align}
\Big(1 - \frac{1}{n}\Big) P_M &\leq \frac{1}{M - 1}\sum_{\ell=1}^{M-1} A_1(\ell) \leq \Big(1 - \frac{1}{n}\Big) P_1, \!\\
P_M^2 &\leq \frac{1}{M-1}\sum_{\ell=1}^{M-1} A_2(\ell) \leq P_1^2.
\end{align}
It can be seen that the convergence upper bound of H-SGD with more than two levels also takes a value that is between the convergence upper bounds of local SGD with local periods $P_1$ and $P_M$, respectively. Compared to the two-level case, this provides greater freedom to choose  parameters in $A_1(\ell)$ and $A_2(\ell)$, for $\ell = 1,\ldots ,M-1$.

\section{Experiments}
\label{sec:experiments}

\begin{table*}[t]
\centering
\caption{Total time (s) needed to achieve 50\% test accuracy for VGG-11 with CIFAR-10.}
\vspace{-0.2cm}
\label{tab: wallclocktime}
\small
\begin{tabular}{cccccc}
\hline
Case
& $P=5$  & $P=10$ & $P=50$ &$G=50, I=5$& $G=50, I=10$ \\
\hline
Time (s) & $673.5$ & $690.2$ & $944.3$ & $160.3$ & $381.1$ \\
\hline
\end{tabular}
\end{table*}

\begin{figure*}[h]
\centering
\begin{subfigure}{0.32\textwidth}
  \centering
  \includegraphics[width=4.8cm]{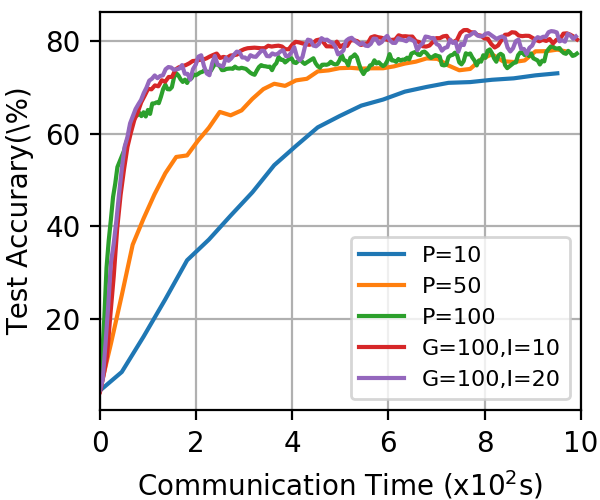}

  \caption{FEMNIST with CNN.}
  \end{subfigure}
  \begin{subfigure}{0.32\textwidth}
  \centering
  \includegraphics[width=4.8cm]{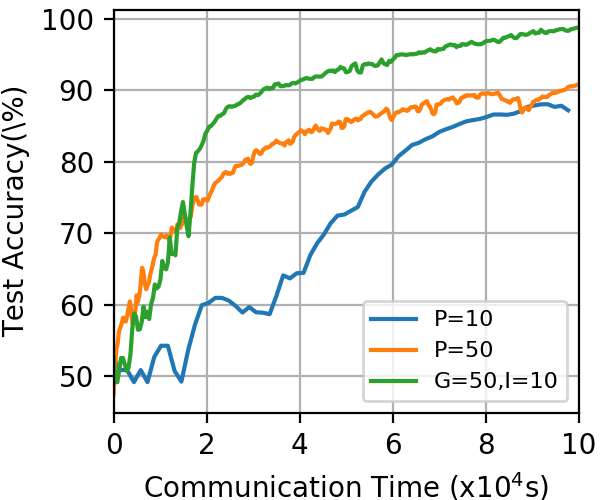}
  \caption{CelebA with VGG-11.}
  \end{subfigure}
  \begin{subfigure}{0.32\textwidth}
  \centering
  \includegraphics[width=4.8cm]{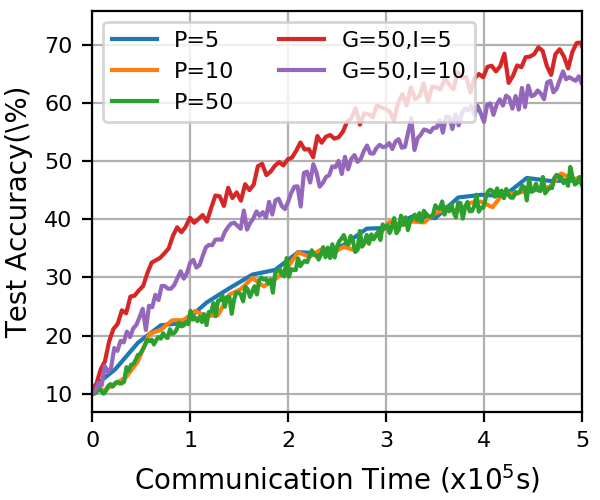}
  \caption{CIFAR-10 with VGG-11.}
  \end{subfigure}
\vspace{-0.2cm}
\caption{Test accuracy v.s. communication time ($N=2$).}
\label{fig:commtime}

\end{figure*}

\begin{figure*}[tb]
  \centering
  \begin{subfigure}{0.32\textwidth}
  \centering
  \includegraphics[width=4.8cm]{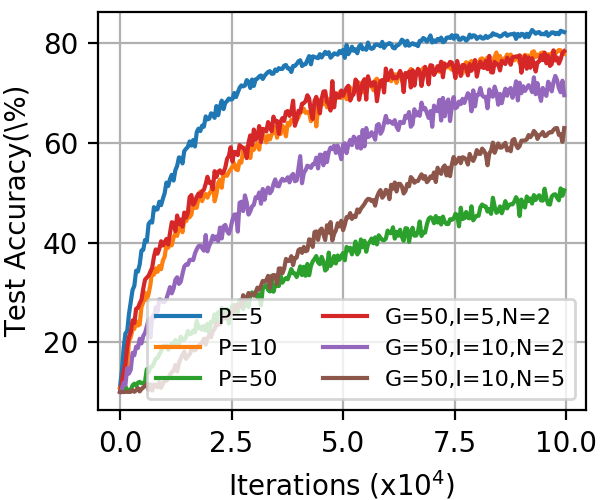}
  \caption{Non-IID ($G=50$ for two-level case).}
  \label{fig:nIID}
  \end{subfigure}
  \begin{subfigure}{0.32\textwidth}
  \centering
  \includegraphics[width=4.8cm]{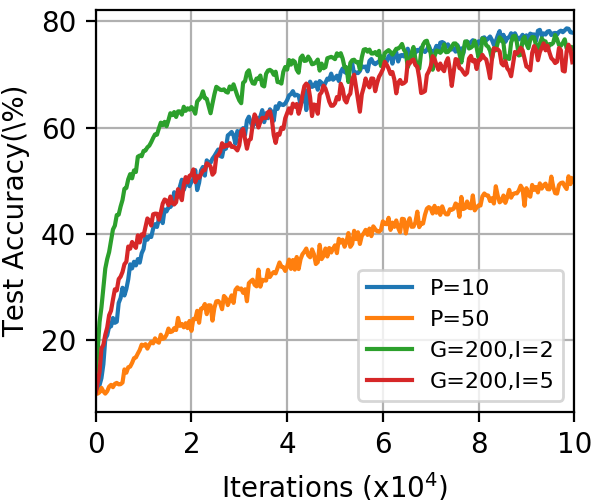}
  \caption{Non-IID ($G=200$ for two-level case).}
  \label{fig:largeG}
  \end{subfigure}
  \begin{subfigure}{0.32\textwidth}
  \centering
  \includegraphics[width=4.8cm]{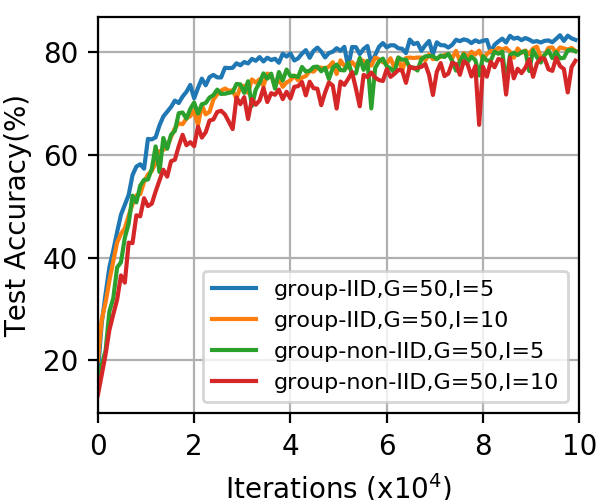}
  \caption{Group-IID/Group-non-IID.}
  \label{fig:gIID}
  \end{subfigure}
  \vspace{-0.1cm}
  \caption{Results with CIFAR-10. Test accuracy v.s. local iterations. 
  By default, $N=2$.}
  \label{fig: cifar 10}
  \vspace{-0.3cm}
\end{figure*}

In this section, we validate our theoretical results with experiments on training the VGG-11 model over CIFAR-10~\citep{krizhevsky2009learning} and CelebA~\citep{liu2015faceattributes}, and training a convolutional neural network (CNN) over FEMNIST~\citep{cohen2017emnist}, all with non-IID data partitioning across workers. The communication time is emulated by measuring the round-trip time of transmitting the model between a device (in a home) and local (near)~/ global (far) Amazon EC2 instances.
For the computation time, we measured the averaged computation time needed for VGG-11 during one iteration, where we ran SGD with VGG-11 on a single GPU for 100 times and then computed the averaged computation time, which is approximately 4 ms per iteration on each worker.
The experiments presented in this section are for \textit{two-level} H-SGD. Additional details of the setup and experiments with \textit{more levels} and \textit{partial worker participation} can be found in Appendix~\ref{appendix:AdditionalExperiments}. Our code is available at \url{https://github.com/C3atUofU/Hierarchical-SGD.git}.

In all plots, curves with $P$ (denoting the aggregation period) are for local SGD and curves with $G$ (global period), $I$ (local period), and $N$ (number of groups) are for H-SGD. Figure~\ref{fig:commtime} shows
that for all the datasets and models evaluated in the experiments, H-SGD can achieve a better accuracy than local SGD given the same communication time. 
In other words, for any given accuracy, H-SGD can achieve this accuracy using shorter or similar communication time. In Table~\ref{tab: wallclocktime}, we provide the total time (including both communication and computation) needed for VGG-11 with CIFAR-10 to achieve $50\%$ test accuracy (since $50\%$ is nearly the best accuracy for $P=50$). We can see H-SGD cases ($G=50, I=5$ and $G=50, I=10$) perform much better than single-level local SGD cases. This is because communication time plays a much more important role and H-SGD can achieve 50\% accuracy within less amount of time in total.

To show the ``sandwich'' behavior, we examine the test accuracy as a function of the total number of local iterations for the non-IID scenario in Figure~\ref{fig:nIID}.  We can observe that the performance of H-SGD with global period $G$ and local period $I$ is between that of local SGD with $P=G$ and with $P=I$. We can also observe that as $N$ becomes larger, the performance becomes worse since the upward divergence becomes larger, which is consistent with our analysis in Remark~4. Comparing Figures~\ref{fig:nIID} and \ref{fig:largeG}, we see that decreasing $I$ while increasing $G$ can even improve the performance, which is consistent with Theorem~\ref{thm2}. 
For example, $G=200,I=2$ gives a better performance than $G=50,I=5$, while $G=50,I=5$ is similar to $P=10$.
This shows that by allowing more local aggregations, H-SGD can reduce the number of global aggregations by $95\%$ ($G=200$ vs. $P=10$) while maintaining a similar performance to local SGD. In Figure~\ref{fig:gIID}, we show the effects of grouping corresponding to our analysis for Theorem~\ref{thm1}. Group-IID is a grouping strategy which makes upward divergence nearly zero while group-non-IID is a grouping strategy with a large upward divergence. We can see that group-IID performs as well as group-non-IID after reducing $I$ by half.

\section{Conclusion}
We have studied H-SGD with multi-level model aggregations. In particular, we have provided a thorough theoretical analysis on the convergence of H-SGD over non-IID data, under both fixed and random worker grouping. We have successfully answered the important question on how local aggregation affects convergence of H-SGD. 
Based on our novel analysis of the local and global divergences, we established
explicit comparisons of the convergence rate 
for H-SGD and local SGD.
Our theoretical analysis provides valuable insights into the design of practical H-SGD systems, including the choice of global and local aggregation periods. Different grouping strategies are considered to best utilize the divergences to reduce communication costs while accelerating learning convergence. In addition, we have extended the analysis approach to H-SGD with arbitrary number of levels.
Future work could theoretically analyze the effect of partial worker participation in H-SGD.

\clearpage

\section*{Acknowledgment}
    This research was partly sponsored by the U.S. Army Research Laboratory and the U.K. Ministry of Defence under Agreement Number W911NF-16-3-0001. The views and conclusions contained in this document are those of the authors and should not be interpreted as representing the official policies, either expressed or implied, of the U.S. Army Research Laboratory, the U.S. Government, the U.K. Ministry of Defence or the U.K. Government. The U.S. and U.K. Governments are authorized to reproduce and distribute reprints for Government purposes notwithstanding any copyright notation hereon. 


\clearpage

\onecolumn

\newcommand{\E}{\mathbb{E}}
\newcommand{\w}{\mathbf{w}}
\newcommand{\g}{\mathbf{g}}

\begin{center}
\LARGE \textbf{Appendix}
\end{center}

\appendix
\vspace{3em}

\numberwithin{equation}{section}
\counterwithin{figure}{section}
\counterwithin{algocf}{section}
\counterwithin{theorem}{section}
\counterwithin{lemma}{section}
\counterwithin{table}{section}

\startcontents[sections]
\printcontents[sections]{l}{1}{\setcounter{tocdepth}{2}}

\clearpage

\section{Additional Discussions}
In this section, we provide some more discussions on the differences between decentralized SGD and H-SGD and also on the justification of the bounded gradient divergence assumptions (Assumption~\ref{assumption:hf-sgd} for H-SGD and Assumption~\ref{assumption:divergence-sf-sgd} for local SGD).

\subsection{Differences between H-SGD and Decentralized SGD}
In decentralized SGD, a doubly stochastic mixing is used to describe how workers exchange models with their neighbors. A necessary assumption for this mixing matrix is that its secondary largest absolute eigenvalue $\xi$ is strictly less than 1. We can see this in classic works for decentralized SGD \citep{CooperativeSGD, lian2017can}. This is because
in their proofs, $1-\xi$ is present in the denominator. 

In H-SGD, local aggregation can also be described by a matrix $\mathbf{A}$. However, $\mathbf{A}$ does not satisfy the assumption for the mixing matrix in decentralized SGD. To see this, consider the following non-trivial example for H-SGD. Suppose there are $K = \frac{n}{N}, 1<N<n$ nodes in each group. Let
$$
\mathbf{A} =
\begin{bmatrix}
    \mathbf{D} & \mathbf{0}  & \dots  & \mathbf{0} \\
    \mathbf{0} & \mathbf{D} & \dots  & \mathbf{0} \\
    \vdots & \vdots  & \ddots & \vdots \\
    \mathbf{0} & \mathbf{0}  & \dots  & \mathbf{D}
\end{bmatrix},
$$
where
$$
\mathbf{D} = 
\begin{bmatrix}
    \frac{1}{K} & \frac{1}{K}  & \dots  & \frac{1}{K} \\
    \frac{1}{K} & \frac{1}{K}  & \dots  & \frac{1}{K} \\
    \vdots & \vdots  & \ddots & \vdots \\
    \frac{1}{K} & \frac{1}{K}  & \dots  & \frac{1}{K}
\end{bmatrix}.
$$
Then we have $\det (\mathbf{A}-\lambda\mathbf{I}) = \det (\mathbf{D}-\lambda\mathbf{I})^N$. Since $\mathbf{D}$ is a rank-1 matrix with eigenvalue 1, there will be $N$ 
eigenvalues with value 1, which does not satisfy the assumption for the mixing matrix in decentralized SGD. 

The above shows that the convergence result of H-SGD cannot be obtained from the approach based on the mixing matrix. Our result provides useful insights on how grouping and local/global periods affect the convergence, which cannot be obtained using standard approaches for decentralized SGD or local SGD.

\subsection{Justification of the Bounded Gradient Divergence Assumptions}
Bounded gradient divergence assumption is widely used and well accepted in literature for theoretical analysis for local SGD \citep{NEURIPS2019_c17028c9,karimireddy2020scaffold,Li2020On,stich2018local,WangJSAC2019,YuAAAI2019,yu2019linear} to describe the degree of data heterogeneity. In this paper, upward and downward divergence are essential for the convergence analysis of H-SGD. First, with a fixed global divergence, there can be different kinds of data distributions in a group. We use upward and downward divergence to capture these characteristics and to show how they can affect the convergence of H-SGD. The standard global divergence assumption cannot capture the heterogeneous characteristics of data distribution within each group and across different groups. 

Upward and downward divergence can also provide us insights on how to group workers. If we group workers with similar data distribution, then the downward divergence can be very small and the benefit of local aggregation can be limited, so it is better to group workers with diverse data. This is one implication of Theorem~\ref{thm1}. 

A classic example that satisfies this bounded gradient divergence assumption is the cross-entropy loss with softmax. Consider the following model:
$$
\mathbf{z} = \mathbf{W}\mathbf{x},
$$
where $\mathbf{x}$ is the input data and $\mathbf{W}$ is a matrix representing the model parameters.
After softmax, we have
$$
s_j = \frac{e^{z_j}}{\sum_i e^{z_i}}, \forall j,
$$
where $z_j$ is the $j$-th component of $\mathbf{z}$.
The cross-entropy loss function is given by 
$$
L(\mathbf{y},\mathbf{s}) = -\sum_i y_i\log (s_i),
$$
where $\mathbf{y}$ is the one-hot encoded ground-truth label.
Then we have 
$$
\frac{\partial L}{\partial \mathbf{z}} = \mathbf{y} - \mathbf{s}.
$$
Because $0\le s_i \le 1, 0\le y_i \le 1 ,\forall i$, we have $|y_i-s_i|\le 1$. That is, the derivative of $L$ with respect to $\mathbf{z}$ is bounded. Since $\mathbf{z}$ is a linear function of $\mathbf{W}$, %
the derivative of $\mathbf{z}$ with respect to $\mathbf{W}$ 
is also bounded as long as the data sample $\mathbf{x}$ is bounded. Then, the gradient of the loss $L$ with respect to the model parameter $\mathbf{W}$ is the product of $\frac{\partial L}{\partial \mathbf{z}}$ and $\frac{\partial \mathbf{z}}{\partial \mathbf{W}}$, which is also bounded and depends on the input data $\mathbf{x}$. This boundedness of derivatives extends to neural network models with sigmoid or ReLU activation functions, since these functions also have bounded derivatives. Hence, the divergence, which is defined as the norm of the difference between gradients, is also bounded.

\clearpage

\section{Proofs for the Two-level Fixed Grouping Case}
We present the proofs of Theorem~\ref{thm1} in this section.
Throughout the proof, we use the following inequalities frequently: 
\begin{equation}\label{l1}
\Bigg\| \sum_{i=1}^M p_i \mathbf{x}_i\Bigg\|^2  \le \sum_{i=1}^M p_i\| \mathbf{x}_i\|^2, 
\end{equation}
and
\begin{equation}\label{l2}
\sum_{i=1}^M p_i \|\mathbf{x}_i-\mathbf{\bar x}\|^2 = \sum_{i=1}^M p_i\|\mathbf{x}_i \|^2 - \|\mathbf{\bar x} \|^2 \le\sum_{i=1}^M p_i\|\mathbf{x}_i \|^2,    
\end{equation}
where $\bar{\mathbf{x}} :=  \sum_{i=1}^M p_i \mathbf{x}_i$, $0 \le p_i\le 1$ and $\sum_{i=1}^M p_i =1$ and (\ref{l1}) is due to Jensen's Inequality.  

For the ease of notations, we define $\mathbf{g}_{j}(\mathbf{w}) := \mathbf{g}(\mathbf{w}, \zeta_{j})$ for any $\mathbf{w}$.

\subsection{Proof of Theorem~\ref{thm1}}
Although the averaged global model $\bar \w^t$ is not observable in the system at $t \neq aG, a=0,1,2,\ldots $, here we use $\bar \w^t$ for analysis. 
\begin{align}\label{E1}
&\E f(\bar \w^{t+1}) = \E f\bigg[\bar \w^t - \gamma \frac{1}{n}\sum_{i=1}^N \sum_{j\in\mathcal{V}_i}\g_{j}(\w_j^t)\bigg] \notag \\
&\overset{(a)}{\le} \E f(\bar \w^t)-\gamma \E \bigg \langle\nabla f(\bar \w^t),\frac{1}{n}\sum_{i=1}^N \sum_{j\in\mathcal{V}_i}\g_{j}(\w_j^t) \bigg\rangle 
+ \frac{\gamma^2 L}{2} \E  \bigg \| \frac{1}{n}\sum_{i=1}^N \sum_{j\in\mathcal{V}_i}\g_j(\w_j^t) \bigg \|^2, 	
\end{align}
where $(a)$ is a proposition of Lipschitz smooth, which is shown in (4.3) of \cite{bottou2018optimization}.  
For the inner product term 
\begin{align}\label{cross1}
&-\gamma \E
\bigg \langle\nabla f(\bar \w^t),\frac{1}{n}\sum_{i=1}^N\sum_{j\in\mathcal{V}_i}\g_j(\w_j^t) \bigg\rangle \notag\\ 
& \overset{(a)}{=} -\gamma \E\left\{\E\left[
\bigg \langle\nabla f(\bar \w^t),\frac{1}{n}\sum_{i=1}^N\sum_{j\in\mathcal{V}_i}\g_j(\w_j^t) \bigg\rangle\Bigg|\{\w_j^t : \forall j\in\mathcal{V}\}\right]\right\} \notag \\ 
& \overset{(b)}{=} -\gamma \E \bigg \langle\nabla f(\bar \w^t),\frac{1}{n}\sum_{i=1}^N\sum_{j\in\mathcal{V}_i}\nabla F_j(\w_j^t) \bigg\rangle \notag \\
&=\frac{\gamma}{2}\left(\E \bigg\|\nabla f(\bar \w^t) - \frac{1}{n}\sum_{i=1}^N\sum_{j\in\mathcal{V}_i}\nabla F_j(\w_j^t)\bigg\|^2-\E \|\nabla f(\bar \w^t)\|^2 - \E \bigg\| \frac{1}{n}\sum_{i=1}^N\sum_{j\in\mathcal{V}_i}\nabla F_j(\w_j^t) \bigg\|^2\right). 
 \end{align}
Recalling that  $\mathbf{g}_{j}(\mathbf{w}_j^t)=  \mathbf{g}(\mathbf{w}_j^t, \zeta_{j}^t)$, we note that 
the expectation operator $\E$ in (\ref{E1}) is over all random  samples $\{\zeta_{j}^{\tau}: \tau=0, 1, \cdots, t, j=1, \cdots, n\}$, where $\zeta_{j}^t$ denotes the random samples used for SGD in iteration $t$. In step $(a)$ of (\ref{cross1}), a conditional expectation is taken for given $\{\w_j^t, j=1, \cdots, n\}$. We note that $\E \left[\g_j(\w_j^t)|\{\w_j^t : \forall j\in\mathcal{V}\}\right] = \nabla F_j(\w_j^t)$ due to the unbiased gradient assumption, from which $(b)$ follows. 
This way of replacing the random gradient $\g_j(\w_j^t)$ by its unbiased average $\nabla F_j(\w_j^t)$ through the conditional expectation is also used later in the proof, where we may not write out the conditional expectation step for compactness.

For the last term of (\ref{E1}), we have
\begin{align}\label{E3}
&\E \bigg\| \frac{1}{n}\sum_{i=1}^N \sum_{j\in\mathcal{V}_i}\g_j(\w_j^t)  \bigg\|^2  \notag \\
&\overset{(a)}{=}\E \bigg\| \frac{1}{n}\sum_{i=1}^N \sum_{j\in\mathcal{V}_i}\left(\g_j(\w_j^t)-\nabla F_j(\w_j^t) \right)\bigg \|^2 + \E \bigg\| \frac{1}{n}\sum_{i=1}^N \sum_{j\in\mathcal{V}_i} \nabla F_j(\w_j^t) \bigg \|^2	\notag \\
&\le   \frac{1}{n} \sigma^2 +\E\bigg \| \frac{1}{n}\sum_{i=1}^N \sum_{j\in\mathcal{V}_i} \nabla F_j(\w_j^t)\bigg \|^2,
\end{align}
where $(a)$ holds due to $\E\| \mathbf{x}\|^2= \E\|\mathbf{x}-\E[\mathbf{x}]\|^2+\|\E[\mathbf{x}]\|^2$. 

Substitute (\ref{cross1}) and (\ref{E3}) into (\ref{E1}), we have
\begin{align}
\E f(\bar \w^{t+1}) &\le \E f(\bar \w^t) + \frac{\gamma^2L}{2n}\sigma^2 -\frac{\gamma}{2}\E \|\nabla f(\bar \w^t)\|^2 \notag \\
&-\bigg(\frac{\gamma}{2}-\frac{\gamma^2L}{2}\bigg)\E\bigg \| \frac{1}{n}\sum_{i=1}^N \sum_{j\in\mathcal{V}_i} \nabla F_j(\w_j^t)\bigg \|^2 
+ \frac{\gamma}{2}\E \bigg\|\nabla f(\bar \w^t) - \frac{1}{n}\sum_{i=1}^N\sum_{j\in\mathcal{V}_i}\nabla F_j(\w_j^t) \bigg \|^2.
\end{align}
Suppose $\gamma \le \frac{1}{L}$, that is, $\frac{\gamma}{2}-\frac{\gamma^2L}{2} \ge 0$. We can obtain
\begin{align}\label{E2}
\E f(\bar \w^{t+1}) &\le \E f(\bar \w^t) + \frac{\gamma^2L}{2n}\sigma^2 -\frac{\gamma}{2}\E \|\nabla f(\bar \w^t)\|^2 \notag\\
 &+ \frac{\gamma}{2}\E \bigg\|\nabla f(\bar \w^t) - \frac{1}{n}\sum_{i=1}^N\sum_{j\in\mathcal{V}_i}\nabla F_j(\w_j^t)\bigg \|^2.
\end{align}
Now we compute the upper bound of the last term of inequality~(\ref{E2})
\begin{align}
&\E\bigg\|\nabla f(\bar \w^t) - \frac{1}{n}\sum_{i=1}^N\sum_{j\in\mathcal{V}_i}\nabla F_j(\w_j^t)\bigg \|^2 \notag \\
&\le \E\bigg\| \nabla f(\bar \w^t) - \sum_{i=1}^N\frac{n_i}{n}\nabla f_i(\bar \w_i^t)+ \sum_{i=1}^N\frac{n_i}{n}\nabla f_i(\bar \w_i^t) -\sum_{i=1}^N\frac{n_i}{n} \frac{1}{n_i}\sum_{j\in\mathcal{V}_i}\nabla F_j(\w_j^t)\bigg \|^2 \notag \\
&\le 2\E \bigg\| \nabla f(\bar \w^t) - \sum_{i=1}^N\frac{n_i}{n}\nabla f_i(\bar \w_i^t)\bigg\|^2 + 2\E \bigg \|\sum_{i=1}^N\frac{n_i}{n}\nabla f_i(\bar \w_i^t) -\sum_{i=1}^N\frac{n_i}{n}\frac{1}{n_i}\sum_{j\in\mathcal{V}_i}\nabla F_j(\w_j^t) \bigg \|^2 \notag \\
&\overset{(a)}{\le} 2L^2\sum_{l=1}^N\frac{n_l}{n}\E\|\bar \w^t-\bar \w_l^t\|^2 
+2L^2\frac{1}{n}\sum_{i=1}^N \sum_{k\in\mathcal{V}_i}\E \| \bar \w_i^t-\w_{k}^t \|^2, 
\label{eq:global_local_divergence_est}
\end{align}
where $(a)$ is due to Lipschitz gradient Assumption~\ref{assumption:hf-sgd}(a) in the main paper and (\ref{l1}).

Substituting (\ref{eq:global_local_divergence_est}) to (\ref{E2}) and rearranging the order, we have 
\begin{align}
\frac{\gamma}{2}\E \|\nabla f(\bar \w^t)\|^2 &\le \E f(\bar \w^t) - \E f(\bar \w^{t+1}) + \frac{\gamma^2L}{2n}\sigma^2 
+ \frac{\gamma}{2}\Big[ 2L^2\sum_{l=1}^N\frac{n_l}{n}\E\|\bar \w^t-\bar \w_l^t\|^2 \notag \\
& +2L^2\frac{1}{n}\sum_{i=1}^N \sum_{k\in\mathcal{V}_i}\E \| \bar \w_i^t-\w_{k}^t \|^2 \Big].
\end{align}
Dividing both sides by $\frac{\gamma}{2}$ and taking the average over time, we have 
\begin{align}\label{final}
&\frac{1}{T} \sum_{t=0}^{T-1}\E\|\nabla f(\bar \w^t)\|^2 \notag \\
&\le \frac{2}{\gamma T}\Big[ f(\bar \w^0) \!-\! \E f(\bar \w^T)\Big]
\!+\! \frac{\gamma L \sigma^2}{n} 
\!+\! 2L^2\frac{1}{T}\!\sum_{t=0}^{T-1}\sum_{l=1}^N\frac{n_l}{n}\E\|\bar \w^t\!-\!\bar \w_l^t\|^2\notag \\
&+\! 2L^2\frac{1}{T}\!\sum_{t=0}^{T-1}\frac{1}{n}\sum_{i=1}^N \sum_{k\in\mathcal{V}_i}\E \| \bar \w_i^t\!-\!\w_{k}^t \|^2 \notag \\
&\le \frac{2}{\gamma T}\Big[f(\bar \w^0) - f^* \Big]
+ \frac{\gamma L\sigma^2}{n}
+ 2L^2\frac{1}{T}\sum_{t=0}^{T-1}\sum_{l=1}^N\frac{n_l}{n}\E\|\bar \w^t-\bar \w_l^t\|^2 \notag \\
&+ 2L^2\frac{1}{T}\sum_{t=0}^{T-1}\frac{1}{n}\sum_{i=1}^N \sum_{k\in\mathcal{V}_i}\E \| \bar \w_i^t-\w_{k}^t\|^2,
\end{align}
where $f^* = \min_{\w} f(\w)$.

Note that (\ref{final}) is the key equation that we will use to prove Theorem 1. It involves two major terms. The first is $\frac{1}{T}\sum_{t=0}^{T-1}\sum_{l=1}^N\frac{n_l}{n}\E\|\bar \w^t-\bar \w_l^t\|^2$, which represents the upward parameter mean-square-error (MSE) between the globally aggregated parameter $\bar \w^t$ and the locally aggregated parameters $\{\bar \w_l^t, l=1, \cdots, N\}$ from $N$ groups. The second term is $\frac{1}{T}\sum_{t=0}^{T-1}\frac{1}{n}\sum_{i=1}^N \sum_{k\in\mathcal{V}_i}\E \| \bar \w_i^t-\w_{k}^t\|^2$, which represents the downward parameter MSE between each local parameter $\w_j^t$ and its locally aggregated parameter $\bar \w_i^t$ for each worker $j$ that belongs to group $i$. Next, we will analyze these two terms separately.

\subsection[Bounding Upward Parameter MSE]{Bounding Upward Parameter MSE
\texorpdfstring{$\frac{1}{T}\sum_{t=0}^{T-1}\sum_{l=1}^N\frac{n_l}{n}\E\|\bar \w^t-\bar \w_l^t\|^2 $}{} in (\ref{eq:global_local_divergence_est})}\label{global-mse}
In this section, we will bound $\frac{1}{T}\sum_{t=0}^{T-1}\sum_{l=1}^N\frac{n_l}{n}\E\|\bar \w^t-\bar \w_l^t\|^2$ in terms of global divergence $\epsilon^2$ and $\sigma^2$. 
Note that when $t=aG, a = 0,1,2,\ldots $,  we have $\bar \w^t = \bar \w^t_{l}, l=1,2,\ldots ,N$. 
Suppose $t=aG+c, c=0,1,\ldots $. When $1 \le c \le G-1$, we have 
\begin{align}\label{E7}
& \sum_{l=1}^N\frac{n_l}{n}\E\|\bar \w^t-\bar \w_l^t\|^2 \notag   \\
&= \sum_{l=1}^N\frac{n_l}{n}\E \Bigg\| \Big(\bar \w^{t-1} -\frac{\gamma}{n}\sum_{i=1}^N \sum_{j\in\mathcal{V}_i} \g_j(\w_j^{t-1})\Big) - \Big( \bar \w_l^{t-1} -\frac{\gamma}{n_l}\sum_{m\in\mathcal{V}_l} \g_m (\w_m^{t-1})   \Big) \Bigg\|^2 \notag \\
&= \gamma^2\sum_{l=1}^N\frac{n_l}{n} \E\left\|\sum^{aG+c-1}_{\tau=aG}\left(\frac{1}{n_l}\sum_{m\in\mathcal{V}_l}\g_m(\w_m^{\tau})- \frac{1}{n}\sum_{i=1}^N \sum_{j\in\mathcal{V}_i} \g_j(\w_j^{\tau}) \right) \right\|^2 \notag   \\
&= \gamma^2\sum_{l=1}^N\frac{n_l}{n} \E \Bigg \| \sum^{aG+c-1}_{\tau=aG} \Bigg(\frac{1}{n_l}\sum_{m\in\mathcal{V}_l}\g_m(\w_m^{\tau})
- \nabla f_l (\bar\w^{\tau}_l)
+\nabla f_l (\bar\w^{\tau}_l) - \sum_{i=1}^N\frac{n_i}{n}\nabla f_i(\bar\w_i^{\tau})+ \sum_{i=1}^N\frac{n_i}{n}\nabla f_i(\bar\w_i^{\tau}) \notag \\
&\quad\quad\quad -\frac{1}{n}\sum_{i=1}^N \sum_{j\in\mathcal{V}_i} \g_j(\w_j^{\tau}) \Bigg) \Bigg\|^2 \notag \\
&\overset{(a)}{\le} 2\gamma^2\sum_{l=1}^N \frac{n_l}{n}\E \Bigg\| \sum^{aG+c-1}_{\tau=aG}\Bigg( \Big(\frac{1}{n_l}\sum_{m\in\mathcal{V}_l}\g_m(\w_m^{\tau})
- \nabla f_l (\bar\w^{\tau}_l) \Big)- \Big( \frac{1}{n}\sum_{i=1}^N \sum_{j\in\mathcal{V}_i} \g_j(\w_j^{\tau})-\sum_{i=1}^N\frac{n_i}{n}\nabla f_i(\bar\w_i^{\tau})
\Big) \Bigg) \Bigg\|^2 \notag \\
&\quad + 2\gamma^2\sum_{l=1}^N \frac{n_l}{n}\E \Bigg\|\sum^{aG+c-1}_{\tau=aG}\Bigg(\nabla f_l (\bar\w^{\tau}_l) - \sum_{i=1}^N\frac{n_i}{n}\nabla f_i(\bar\w_i^{\tau})\Bigg) \Bigg\|^2,
\end{align}
where $(a)$ is due to (\ref{l1}).
Next, we will derive an upper bound on each of the two terms in 
(\ref{E7}) separately. 
For the first term of (\ref{E7}), we have
\begin{align}\label{E8}
&2\gamma^2\sum_{l=1}^N \frac{n_l}{n}\E \Bigg\| \sum^{aG+c-1}_{\tau=aG}\Bigg(\Big(\frac{1}{n_l}\sum_{m\in\mathcal{V}_l}\g_m(\w_m^{\tau})
- \nabla f_l (\bar\w^{\tau}_l)\Big)- \Big( \frac{1}{n}\sum_{i=1}^N \sum_{j\in\mathcal{V}_i} \g_j(\w_j^{\tau})-\sum_{i=1}^N\frac{n_i}{n}\nabla f_i(\bar\w_i^{\tau})
\Big)\Bigg) \Bigg\|^2 \notag \\
&= 2\gamma^2\sum_{l=1}^N \frac{n_l}{n}\E \Bigg\| \sum^{aG+c-1}_{\tau=aG}\Bigg( \Big(\frac{1}{n_l}\sum_{m\in\mathcal{V}_l}\g_m(\w_m^{\tau})
-\frac{1}{n_l}\sum_{m\in\mathcal{V}_l}\nabla F_m(\w_m^{\tau})
+\frac{1}{n_l}\sum_{m\in\mathcal{V}_l}\nabla F_m(\w_m^{\tau})
- \nabla f_l (\bar\w^{\tau}_l)\Big) \notag \\
&\quad\quad - \Big( \frac{1}{n}\sum_{i=1}^N \sum_{j\in\mathcal{V}_i} \g_j(\w_j^{\tau}) 
-\frac{1}{n}\sum_{i=1}^N \sum_{j\in\mathcal{V}_i} \nabla F_j(\w_j^{\tau})
+\frac{1}{n}\sum_{i=1}^N \sum_{j\in\mathcal{V}_i} \nabla F_j(\w_j^{\tau}) 
-\sum_{i=1}^N\frac{n_i}{n}\nabla f_i(\bar\w_i^{\tau})
\Big)\Bigg) \Bigg\|^2 \notag \\
&= 2\gamma^2\sum_{l=1}^N \frac{n_l}{n} \E \Bigg\| \sum^{aG+c-1}_{\tau=aG}\!\Bigg(\!\Big(\frac{1}{n_l}\sum_{m\in\mathcal{V}_l}\!\g_m(\w_m^{\tau})\!-\!\frac{1}{n_l}\sum_{m\in\mathcal{V}_l}\!\nabla F_m(\w_m^{\tau})\!\Big)\!-\!\Big(\frac{1}{n}\sum_{i=1}^N \sum_{j\in\mathcal{V}_i} \g_j(\w_j^{\tau}) 
\!-\!\frac{1}{n}\sum_{i=1}^N \sum_{j\in\mathcal{V}_i} \nabla F_j(\w_j^{\tau})\!\Big) \notag \\
&\quad\quad\quad +\Big(\frac{1}{n_l}\sum_{m\in\mathcal{V}_l}\nabla F_m(\w_m^{\tau})
- \nabla f_l (\bar\w^{\tau}_l)\Big)
-\Big(\frac{1}{n}\sum_{i=1}^N \sum_{j\in\mathcal{V}_i} \nabla F_j(\w_j^{\tau}) 
-\sum_{i=1}^N\frac{n_i}{n}\nabla f_i(\bar\w_i^{\tau})\Big)
\Bigg) \Bigg\|^2 \notag \\
&\overset{(a)}{\le}\! 4\gamma^2\sum_{l=1}^N \frac{n_l}{n} \E \Bigg\| \sum^{aG+c-1}_{\tau=aG}\!\Bigg(\!\Big(\frac{1}{n_l}\!\sum_{m\in\mathcal{V}_l}\!\g_m(\w_m^{\tau})
\!-\!\frac{1}{n_l}\!\sum_{m\in\mathcal{V}_l}\!\nabla F_m(\w_m^{\tau})\!\Big)\!-\!
\Big(\frac{1}{n}\sum_{i=1}^N\! \sum_{j\in\mathcal{V}_i} \!\g_j(\w_j^{\tau}) 
\!-\!\frac{1}{n}\sum_{i=1}^N\! \sum_{j\in\mathcal{V}_i}\! \nabla F_j(\w_j^{\tau})\!\Big)\! \Bigg)\! \Bigg\|^2 \notag \\
&\quad\quad\quad +4\gamma^2\sum_{l=1}^N \frac{n_l}{n} \E \Bigg\| \sum^{aG+c-1}_{\tau=aG}\Bigg(\frac{1}{n_l}\sum_{m\in\mathcal{V}_l}\nabla F_m(\w_m^{\tau})
- \nabla f_l (\bar\w^{\tau}_l)
\Bigg) \Bigg\|^2, 
\end{align} 
where $(a)$ follows from (\ref{l1}) with $M=2$ and (\ref{l2}).  

For the first term of (\ref{E8}), we have
\begin{align}\label{e13}
&4\gamma^2\sum_{l=1}^N \frac{n_l}{n} \E \Bigg\| \sum^{aG+c-1}_{\tau=aG}\!\!\Bigg(\! \Big(\frac{1}{n_l}\!\sum_{m\in\mathcal{V}_l}\!\g_m(\w_m^{\tau})
\!-\!\frac{1}{n_l}\!\sum_{m\in\mathcal{V}_l}\!\nabla F_m(\w_m^{\tau}) \Big)\!
-\!
\Big(\frac{1}{n}\sum_{i=1}^N \sum_{j\in\mathcal{V}_i} \g_j(\w_j^{\tau}) 
\!-\!\frac{1}{n}\sum_{i=1}^N \sum_{j\in\mathcal{V}_i} \nabla F_j(\w_j^{\tau}) \Big)\! \Bigg) \! \Bigg\|^2 \notag \\
&\overset{(a)}{=} 4\gamma^2\Bigg(\sum_{l=1}^N\frac{n_l}{n}\E\Bigg\| \sum^{aG+c-1}_{\tau=aG}\Bigg(\frac{1}{n_l}\sum_{m\in\mathcal{V}_l}\g_m(\w_m^{\tau})-\frac{1}{n_l}\sum_{m\in\mathcal{V}_l}\nabla F_m(\w_m^{\tau})\Bigg) \Bigg\|^2 \notag\\
&\quad\quad\quad - \E \Bigg\|\sum^{aG+c-1}_{\tau=aG}\Bigg(\frac{1}{n}\sum_{i=1}^N \sum_{j\in\mathcal{V}_i} \g_j(\w_j^{\tau}) 
-\frac{1}{n}\sum_{i=1}^N \sum_{j\in\mathcal{V}_i} \nabla F_j(\w_j^{\tau})\Bigg)\Bigg\|^2\Bigg)	\notag \\
&\overset{(b)}{=} 4\gamma^2\sum_{\tau=aG}^{aG+c-1}\Bigg(\sum_{l=1}^N \frac{n_l}{n} \frac{1}{n_l^2}\sum_{m\in\mathcal{V}_l}\E\Big\| \g_m(\w_m^{\tau}) - \nabla F_m(\w_m^{\tau})\Big\|^2 - \frac{1}{n^2}\sum_{i=1}^N  \sum_{j\in\mathcal{V}_i}\E \Big\|\g_j(\w_j^{\tau})-\nabla F_j(\w_j^{\tau})\Big\|^2\Bigg)\notag \\
&\overset{(c)}{=} 4\gamma^2\sum_{\tau=aG}^{aG+c-1} \frac{1}{n}\sum_{i=1}^N\bigg(\frac{1}{n_i} - \frac{1}{n}\bigg) \sum_{j\in\mathcal{V}_i}\E \Big\|\g_j(\w_j^{\tau})-\nabla F_j(\w_j^{\tau})\Big\|^2   \notag \\
&\le 4\gamma^2G \frac{N-1}{n}\sigma^2 \quad \text{(use Assumption 1b)},
\end{align}

where $(a)$ is due to (\ref{l2}), $(b)$ and $(c)$ hold due to the independence of SGD noise over different time, groups and nodes, and $\mathbb{E}[\mathbf{g}_i(\mathbf{w}_{i}^t)] = \mathbb{E}\left[\mathbb{E}[\mathbf{g}_i(\mathbf{w}_{i}^t) | \mathbf{w}^t_i]\right]= \mathbb{E}[\nabla F_{i}(\mathbf{w}^t_i)], \forall i,t$. 
This implies that $\E\langle \g_i(\w_i^t) - \nabla F_i(\w_i^t),\g(\w_j^{t'})-\nabla F_j(\w_j^{t'})\rangle = 0, i\neq j$.

 For the second term of (\ref{E8}), we have 
\begin{align}\label{E14}
&4\gamma^2\sum_{l=1}^N \frac{n_l}{n} \E \Bigg\| \sum^{aG+c-1}_{\tau=aG}\Bigg(\frac{1}{n_l}\sum_{m\in\mathcal{V}_l}\nabla F_m(\w_m^{\tau})
- \nabla f_l (\bar\w^{\tau}_l)
\Bigg) \Bigg\|^2 \notag \\
&\le 4\gamma^2G\sum_{l=1}^N \frac{n_l}{n}\sum^{aG+c-1}_{\tau=aG} \E\Bigg\|\frac{1}{n_l}\sum_{m\in\mathcal{V}_l}\Bigg(\nabla F_m(\w_m^{\tau})-\nabla F_m(\bar\w_l^{\tau})\Bigg)\Bigg\|^2 \quad \text{(use (\ref{l1}) with $M=G$)}\notag \\
&\le 4\gamma^2GL^2\sum^{aG+c-1}_{\tau=aG}\sum_{l=1}^N\frac{n_l}{n}\frac{1}{n_l}\sum_{m\in\mathcal{V}_l}\E\|\w_m^{\tau}-\bar\w_l^{\tau}\|^2  \quad \text{(use (\ref{l1}) with $M=N_{\ell}$ and Assumption \ref{assumption:hf-sgd}a)}. 	
\end{align}\label{15}
For the second term of (\ref{E7}), we have
\begin{align}\label{e15}
&2\gamma^2\sum_{l=1}^N \frac{n_l}{n}\E\Bigg\|\sum^{aG+c-1}_{\tau=aG}\Bigg(\nabla f_l (\bar\w^{\tau}_l) - \sum_{i=1}^N\frac{n_i}{n}\nabla f_i(\bar\w_i^{\tau})\Bigg)\Bigg\|^2	\notag \\
&\le 2\gamma^2\sum_{l=1}^N \frac{n_l}{n}\E\Bigg\|\sum^{aG+c-1}_{\tau=aG}\Bigg(\nabla f_l (\bar\w^{\tau}_l) - \nabla f_l(\bar{\w}^\tau)+\nabla f_l(\bar{\w}^\tau) -\nabla f(\bar{\w}^\tau)+ \nabla f(\bar{\w}^\tau)-\sum_{i=1}^N\frac{n_i}{n}\nabla f_i(\bar\w_i^{\tau})\Bigg)\Bigg\|^2 \notag \\
&\overset{(a)}{\le} 6\gamma^2L^2G\sum^{aG+c-1}_{\tau=aG} \sum_{l=1}^N\frac{n_l}{n}\E\|\bar\w_l^{\tau} - \bar\w^{\tau}\|^2 + 6\gamma^2G^2 \epsilon^2+ 6\gamma^2L^2G\sum^{aG+c-1}_{\tau=aG}\sum_{i=1}^N\frac{n_i}{n}\E\|\bar\w^\tau-\bar\w_i^{\tau}\|^2 \notag \\
&= 6\gamma^2G^2 \epsilon^2 
 + 12\gamma^2L^2G\sum^{aG+c-1}_{\tau=aG}\sum_{i=1}^N\frac{n_i}{n}\E\|\bar\w^{\tau}-\bar\w_i^{\tau}\|^2,
\end{align}
where (a) uses (\ref{l1}), and Assumptions \ref{assumption:hf-sgd}c and \ref{assumption:hf-sgd}a.  
Now back to inequalities (\ref{E7}), with results of inequality (\ref{E8}), (\ref{e13}), (\ref{E14}), (\ref{e15}), we can obtain
\begin{align}\label{E12}
&\sum_{l=1}^N\frac{n_l}{n}\E\| \bar \w^t-\bar \w_l^t\|^2 \notag \\
&\le 4\gamma^2G\frac{N-1}{n}\sigma^2
+ 6\gamma^2G^2\epsilon^2
+ 4\gamma^2GL^2\sum^{t-1}_{\tau=aG}\sum_{l=1}^N\frac{n_l}{n}\frac{1}{n_l}\sum_{m\in\mathcal{V}_l}\E\|\w_m^{\tau}-\bar\w_l^{\tau}\|^2 \notag \\
& \quad\quad\quad\quad + 12\gamma^2L^2G\sum^{t-1}_{\tau=aG} \sum_{l=1}^N\frac{n_l}{n}\E\|\bar\w_l^{\tau} - \bar\w^{\tau}\|^2 \notag \\
&\overset{(a)}{\le} 4\gamma^2G\frac{N-1}{n}\sigma^2
+ 6\gamma^2G^2\epsilon^2
+ 4\gamma^2GL^2\sum^{aG+G-1}_{\tau =aG}\sum_{l=1}^N\frac{n_l}{n}\frac{1}{n_l}\sum_{m\in\mathcal{V}_l}\E\|\w_m^{\tau}-\bar\w_l^{\tau}\|^2  \notag \\
& \quad\quad\quad\quad + 12\gamma^2L^2G\sum^{aG+G-1}_{\tau =aG} \sum_{l=1}^N\frac{n_l}{n}\E\|\bar\w_l^{\tau} - \bar\w^{\tau}\|^2,,
\end{align}
where $(a)$ is due to $1\le c\le G-1$.
Since $\| \bar \w^t - \bar \w_l^t \|^2=0$, $(a)$ also holds when $t = aG, a=0,1,2\ldots $. Suppose $T-1 = dG, d=0,1,2,\ldots $. Take the average over time on both sides of (\ref{E12}), we have
\begin{align}\label{E13}
&\frac{1}{T} \sum_{t=0}^{T-1} \sum_{l=1}^N\frac{n_l}{n}\E\| \bar \w^t-\bar \w_l^t\|^2  \notag \\
& \overset{(a)}{=} \frac{1}{T} \sum_{t=0}^{T-2} \sum_{l=1}^N\frac{n_l}{n}\E\| \bar \w^t-\bar \w_l^t\|^2 \notag \\
&\le 4\gamma^2G\frac{N-1}{n}\sigma^2
+ 6\gamma^2G^2\epsilon^2
+ 4\gamma^2GL^2\frac{1}{T}\sum_{a=0}^{d-1} \Bigg(G\sum_{\tau=aG}^{aG+G-1}\sum_{l=1}^N\frac{n_l}{n}\frac{1}{n_l}\sum_{m\in\mathcal{V}_l}\E\|\w_m^{\tau}-\bar\w_l^{\tau}\|^2\Bigg) \notag \\
&\quad\quad\quad + 12\gamma^2L^2G\frac{1}{T}  \sum_{a=0}^{d-1} \Bigg( G \sum_{\tau=aG}^{aG+G-1}\sum_{l=1}^N\frac{n_l}{n}\E\|\bar\w_l^{\tau} - \bar\w^{\tau}\|^2 \Bigg)\notag \\
&= 4\gamma^2G\frac{N-1}{n}\sigma^2+ 6\gamma^2G^2\epsilon^2+ 4\gamma^2 G^2 L^2 \frac{1}{T} \sum_{t=0}^{T-2} \sum_{l=1}^N\frac{n_l}{n}\frac{1}{n_l}\sum_{m\in\mathcal{V}_l}\E\|\w_m^{t}-\bar\w_l^{t}\|^2 \notag \\
&\quad\quad\quad + 12\gamma^2L^2G^2 \frac{1}{T} \sum_{t=0}^{T-2} \sum_{l=1}^N\frac{n_l}{n}\E\|\bar\w_l^{t} - \bar\w^{t}\|^2 \notag \\
&\overset{(b)}{=} 4\gamma^2G\frac{N-1}{n}\sigma^2+ 6\gamma^2G^2\epsilon^2 
+ 4\gamma^2 G^2 L^2 \frac{1}{T} \sum_{t=0}^{T-1} \sum_{l=1}^N\frac{n_l}{n}\frac{1}{n_l}\sum_{m\in\mathcal{V}_l}\E\|\w_m^{t}-\bar\w_l^{t}\|^2 
\notag \\
&\quad\quad\quad + 12\gamma^2L^2G^2 \frac{1}{T} \sum_{t=0}^{T-1} \sum_{l=1}^N \frac{n_l}{n}\E\|\bar\w_l^{t} - \bar\w^{t}\|^2,  
\end{align}
where $(a)$ and $(b)$ hold due to $\|\w_m^{T-1} - \bar \w_l^{T-1} \|^2 = 0$ and $\|\bar \w_l^{T-1} - \bar \w^{T-1} \|^2=0$.
Suppose $1-12\gamma^2L^2G^2 > 0$, move the last term of (\ref{E13}) to the left , we can obtain
\begin{align}
    &(1-12\gamma^2 L^2 G^2) \frac{1}{T}\sum_{t=0}^{T-1} \sum_{l=1}^N\frac{n_l}{n}\E\| \bar \w^t-\bar \w_l^t\|^2 \notag \\
    &\le 4\gamma^2G\frac{N-1}{n}\sigma^2
+ 6\gamma^2G^2\epsilon^2 + \frac{1}{T} \sum_{t=0}^{T-1} 4\gamma^2G^2L^2\frac{1}{n}\sum_{l=1}^N\sum_{m\in\mathcal{V}_l}\E\|\w_m^{t}-\bar\w_l^{t}\|^2,
\end{align}
then
\begin{align}\label{E15}
&\frac{1}{T}\sum_{t=0}^{T-1} \sum_{l=1}^N\frac{n_l}{n}\E\| \bar \w^t-\bar \w_l^t\|^2 \notag \\
&\le \frac{4\gamma^2G\frac{N-1}{n}\sigma^2}{1-12\gamma^2 L^2 G^2}
+ \frac{6\gamma^2G^2\epsilon^2}{1-12\gamma^2 L^2 G^2} \notag \\
&\quad\quad\quad + \frac{4\gamma^2G^2L^2}{1-12\gamma^2 L^2 G^2}\frac{1}{T} \sum_{t=0}^{T-1}\frac{1}{n}\sum_{l=1}^N\sum_{m\in\mathcal{V}_l}\E\|\w_m^{t}-\bar\w_l^{t}\|^2.
\end{align}
The importance of (\ref{E15}) is that it presents an upper bound on the upward parameter MSE in terms of the downward parameter MSE. Next, we will proceed to analyze the downward parameter MSE.
\subsection[Bounding Downward Parameter MSE]{Bounding Downward Parameter MSE
\texorpdfstring{$\frac{1}{T}\sum_{t=0}^{T-1}\frac{1}{n}\sum_{i=1}^N \sum_{k\in\mathcal{V}_i}\E \| \bar \w_i^t-\w_{k}^t \|^2$}{} in (\ref{eq:global_local_divergence_est})}\label{local-mse}
Now we consider the local aggregation. Suppose $t = b_i I_i + c_i$ for $i=1,2,3\ldots ,N$ where $b_i = 0,1,\ldots ,\frac{G}{I_i}-1$ and $c_i = 0,1,2,\ldots ,I_i-1$. When $c_i=0$, we have $\bar \w_i^t = \w_k^t, k\in\mathcal{V}_i, \forall i$. When $c_i \ge 1$, we have
\begin{align}\label{E16}
&\frac{1}{n}\sum_{i=1}^N  \sum_{k\in\mathcal{V}_i}\E \| \bar \w_i^t-\w_{k}^t \|^2 \notag \\
&= \frac{1}{n}\sum_{i=1}^N \sum_{k\in\mathcal{V}_i}\E \Bigg\|\Big( \bar\w_i^{t-1} - \gamma \frac{1}{n_i} \sum_{j\in\mathcal{V}_i} \g_j(\w_j^{t-1})\Big) -\Big(\w_{k}^{t-1} - \gamma \g_k(\w_k^{t-1}) \Big) \Bigg\|^2 \notag \\
&=  \frac{\gamma^2}{n}\sum_{i=1}^N \sum_{k\in\mathcal{V}_i} \E \Bigg\|\sum^{t-1}_{\tau=b_iI_i}\Bigg( \g_k(\w_k^{\tau}) - \frac{1}{n_i}\sum_{j\in\mathcal{V}_i} \g_j (\w_j^{\tau}) \Bigg) \Bigg\|^2 \notag \\
&=  \frac{\gamma^2}{n}\sum_{i=1}^N \sum_{k\in\mathcal{V}_i}\E \Bigg\|\!\sum^{t-1}_{\tau=b_iI_i}\!\!\!\Bigg(\! \g_k(\w_k^{\tau}) \!-\! \nabla F_k(\w_k^{\tau}) \!+\!\nabla F_k(\w_k^{\tau}) \!-\! \frac{1}{n_i}\sum_{j\in\mathcal{V}_i}\!\nabla F_j(\w_j^{\tau}) \!+\! \frac{1}{n_i}\sum_{j\in\mathcal{V}_i}\!\nabla F_j(\w_j^{\tau}) \!-\! \frac{1}{n_i}\sum_{j\in\mathcal{V}_i} \g_j (\w_j^{\tau}) \!\Bigg)\! \Bigg\|^2 \notag \\
&= \!\frac{\gamma^2}{n}\sum_{i=1}^N\!\sum_{k\in\mathcal{V}_i}\!\E \Bigg\|\!\sum^{t-1}_{\tau=b_iI_i}\!\!\!\Bigg(\!\! \Big(\!\g_k(\w_k^{\tau}) \!-\! \nabla F_k(\w_k^{\tau})\!+\! \frac{1}{n_i}\!\sum_{j\in\mathcal{V}_i}\!\nabla F_j(\w_j^{\tau}) \!-\! \frac{1}{n_i}\!\sum_{j\in\mathcal{V}_i}\! \g_j (\w_j^{\tau})\!\! \Big) \!\!+\!\! \Big(\nabla F_k(\w_k^{\tau}) \!-\! \frac{1}{n_i}\sum_{j\in\mathcal{V}_i}\nabla F_j(\w_j^{\tau}) \!\Big) \!\! \Bigg) \!\Bigg\|^2 \notag \\
&\le  \frac{2\gamma^2}{n}\sum_{i=1}^N\sum_{k\in\mathcal{V}_i}\E \Bigg\|\sum^{t-1}_{\tau=b_iI_i}\Bigg( \Big(\g_k(\w_k^{\tau}) - \nabla F_k(\w_k^{\tau})\Big)-\Big( \frac{1}{n_i}\sum_{j\in\mathcal{V}_i} \g_j (\w_j^{\tau}) - \frac{1}{n_i}\sum_{j\in\mathcal{V}_i}\nabla F_j(\w_j^{\tau}) \Big)\Bigg)\Bigg\|^2  \\
&\quad\quad\quad + \frac{2\gamma^2}{n} \sum_{i=1}^N \sum_{k\in\mathcal{V}_i}\E \Bigg\|\sum^{t-1}_{\tau=b_iI_i}\Bigg( \nabla F_k(\w_k^{\tau}) - \frac{1}{n_i}\sum_{j\in\mathcal{V}_i}\nabla F_j(\w_j^{\tau}) \Bigg) \Bigg\|^2.
\label{E17}
\end{align}
For (\ref{E16}), we have
\begin{align}\label{E18}
& \frac{2\gamma^2}{n}\sum_{i=1}^N \sum_{k\in\mathcal{V}_i}\E \Bigg\|\sum^{t-1}_{\tau=b_iI_i}\Bigg( \Big(\g_k(\w_k^{\tau}) - \nabla F_k(\w_k^{\tau})\Big)- \Big( \frac{1}{n_i}\sum_{j\in\mathcal{V}_i} \g_j (\w_j^{\tau}) - \frac{1}{n_i}\sum_{j\in\mathcal{V}_i}\nabla F_j(\w_j^{\tau}) \Big)\Bigg)\Bigg\|^2\notag \\
&= \frac{2\gamma^2}{n}\sum_{i=1}^N\sum_{k\in\mathcal{V}_i}\E \Bigg\|\sum^{t-1}_{\tau=b_iI_i}\Bigg(\g_k(\w_k^{\tau}) - \nabla F_k(\w_k^{\tau})\Bigg)\Bigg\|^2 - 2\gamma^2 \sum_{i=1}^N\frac{n_i}{n}\E \Bigg\|\sum^{t-1}_{\tau=b_iI_i}\Bigg(\frac{1}{n_i}\sum_{j\in\mathcal{V}_i} \g_j (\w_j^{\tau}) - \frac{1}{n_i}\sum_{j\in\mathcal{V}_i}\nabla F_j(\w_j^{\tau}) \Bigg)\Bigg\|^2 \notag \\
&=  \frac{2\gamma^2}{n}\sum_{i=1}^N\sum_{k\in\mathcal{V}_i}\sum^{t-1}_{\tau=b_iI_i}\E \Big\|\g_k(\w_k^{\tau}) - \nabla F_k(\w_k^{\tau})\Big\|^2 - 2\gamma^2 \sum_{i=1}^N \frac{n_i}{n} \frac{1}{n_i^2}\sum_{j\in\mathcal{V}_i} \sum^{t-1}_{\tau=b_iI_i}\E \Big\| \g_j (\w_j^{\tau}) - \nabla F_j(\w_j^{\tau}) \Big\|^2 \notag \\
&=\frac{2\gamma^2}{n} \sum_{i=1}^N \Big(1 - \frac{1}{n_i}\Big)\sum_{k\in\mathcal{V}_i}\sum^{t-1}_{\tau=b_iI_i}\E \Big\|\g_k(\w_k^{\tau}) - \nabla F_k(\w_k^{\tau})\Big\|^2 \notag\\
&\le 2\gamma^2\sigma^2 \sum_{i=1}^N \frac{(n_i-1)I_i}{n}. 
\end{align}

For (\ref{E17}), we have
\begin{align}\label{E19}
& \frac{2\gamma^2}{n}\sum_{i=1}^N\sum_{k\in\mathcal{V}_i}\E \Bigg\|\sum^{t-1}_{\tau=b_iI_i}\Bigg( \nabla F_k(\w_k^{\tau}) - \frac{1}{n_i}\sum_{j\in\mathcal{V}_i}\nabla F_j(\w_j^{\tau}) \Bigg) \Bigg\|^2 \notag \\
&=  \frac{2\gamma^2}{n}\sum_{i=1}^N\sum_{k\in\mathcal{V}_i}\E \Bigg\|\sum^{t-1}_{\tau=b_iI_i}\Bigg( \nabla F_k(\w_k^{\tau})
- \nabla F_k(\bar \w^{\tau}_i)
+ \nabla F_k(\bar \w^{\tau}_i)
- \nabla f_i(\bar \w^{\tau}_i)
+ \nabla f_i(\bar \w^{\tau}_i)
- \frac{1}{n_i}\sum_{j\in\mathcal{V}_i}\nabla F_j(\w_j^{\tau}) \Bigg) \Bigg\|^2 \notag \\
&\le  \frac{6\gamma^2}{n}\sum_{i=1}^N \sum_{k\in\mathcal{V}_i}\E\Big\|\sum^{t-1}_{\tau=b_iI_i}\Big( \nabla F_k(\w_k^{\tau})
- \nabla F_k(\bar \w^{\tau}_i)\Big)\Big\|^2
+  \frac{6\gamma^2}{n}\sum_{i=1}^N \sum_{k\in\mathcal{V}_i}\E\Big\|\sum^{t-1}_{\tau=b_iI_i}\Big(\nabla F_k(\bar \w^{\tau}_i)
- \nabla f_i(\bar \w^{\tau}_i)\Big)\Big\|^2 \notag \\
&\quad\quad\quad +6\gamma^2 \sum_{i=1}^N\frac{n_i}{n}\E\Big\|\sum^{t-1}_{\tau=b_iI_i}\Big(\nabla f_i(\bar \w^{\tau}_i)
- \frac{1}{n_i}\sum_{j\in\mathcal{V}_i}\nabla F_j(\w_j^{\tau}) \Big)\Big\|^2 \notag \\
&\overset{(a)}{\le} \frac{12\gamma^2}{n} \sum_{i=1}^N \sum_{k\in\mathcal{V}_i}\E\Big\|\sum^{t-1}_{\tau=b_iI_i}\Big( \nabla F_k(\w_k^{\tau})
- \nabla F_k(\bar \w^{\tau}_i)\Big)\Big\|^2
+ 6\gamma^2 \sum_{i=1}^N \frac{n_i}{n}I_i^2\epsilon_i^2\notag \\
&\le \frac{12\gamma^2 L^2}{n} \sum_{i=1}^N \sum_{k\in\mathcal{V}_i}\sum^{t-1}_{\tau=b_iI_i}I_i\E\Big\|\w_k^{\tau}
- \bar \w^{\tau}_i \Big\|^2
+ 6\gamma^2 \sum_{i=1}^N\frac{n_i}{n} I_i^2\epsilon_i^2,
\end{align}
where $(a)$ is due to (\ref{l1}).

Substituting (\ref{E18}) and (\ref{E19}) back to (\ref{E16}) and (\ref{E17}), we have
\begin{align}\label{E20}
&\frac{1}{n}\sum_{i=1}^N  \sum_{k\in\mathcal{V}_i}\E \| \bar \w_i^t-\w_{k}^t \|^2 \notag \\
&\le 2\gamma^2\sigma^2 \sum_{i=1}^N \frac{(n_i-1)I_i}{n}
+ 6\gamma^2 \sum_{i=1}^N\frac{n_i}{n} I_i^2\epsilon_i^2
+ \frac{12\gamma^2 L^2}{n}\sum_{i=1}^N \sum_{k\in\mathcal{V}_i}\sum^{t-1}_{\tau=b_iI_i}I_i\E\Big\|\w_k^{\tau}
- \bar \w^{\tau}_i \Big\|^2 \notag \\
&\overset{(a)}{\le} 2\gamma^2\sigma^2 \sum_{i=1}^N \frac{(n_i-1)I_i}{n}
+ 6\gamma^2 \sum_{i=1}^N\frac{n_i}{n} I_i^2\epsilon_i^2
+ \frac{12\gamma^2 L^2}{n}\sum_{i=1}^N \sum_{k\in\mathcal{V}_i}\sum^{b_iI_i + I_i -1}_{\tau=b_iI_i}I_i\E\Big\|\w_k^{\tau}
- \bar \w^{\tau}_i \Big\|^2
\end{align}
where $(a)$ holds due to $c_i \le I_i-1$. 
(\ref{E20}) also holds when $t=b_iI_i$ since $\frac{1}{n_i} \sum_{k\in\mathcal{V}_i}\E \| \bar \w_i^{b_iI_i}-\w_{k}^{b_iI_i} \|^2 =0$.
Since $T-1=dG, d=0,1,\ldots $, $T-1$ is also an integral multiple of $I_i$. That is, $T-1=f_iI_i$. Taking the average over time yields
\begin{align}
&\frac{1}{T}\sum_{t=0}^{T-1}\frac{1}{n}\sum_{i=1}^N  \sum_{k\in\mathcal{V}_i}\E \| \bar \w_i^t-\w_{k}^t \|^2 \notag \\
&\le 2\gamma^2\sigma^2 \sum_{i=1}^N \frac{(n_i-1)I_i}{n}
+ 6\gamma^2 \sum_{i=1}^N\frac{n_i}{n} I_i^2\epsilon_i^2
+  \frac{12\gamma^2 L^2}{nT} \sum_{i=1}^N \sum_{k\in\mathcal{V}_i}\sum_{b_i=0}^{f_i-1}\sum^{b_iI_i + I_i -1}_{\tau=b_iI_i}I_i^2\E\Big\|\w_k^{\tau}
- \bar \w^{\tau}_i \Big\|^2 \notag \\
&\le 2\gamma^2\sigma^2 \sum_{i=1}^N \frac{(n_i-1)I_i}{n}
+ 6\gamma^2 \sum_{i=1}^N\frac{n_i}{n} I_i^2\epsilon_i^2
+ \frac{1}{T}\sum_{\tau=0}^{T-1} \frac{12\gamma^2 L^2}{n} \sum_{i=1}^N \sum_{k\in\mathcal{V}_i}I_i^2\E\Big\|\w_k^{\tau}
- \bar \w^{\tau}_i \Big\|^2 \notag \\
&\le 2\gamma^2\sigma^2 \sum_{i=1}^N \frac{(n_i-1)I_i}{n}
+ 6\gamma^2 \sum_{i=1}^N\frac{n_i}{n} I_i^2\epsilon_i^2
+ \frac{1}{T}\sum_{\tau=0}^{T-1}\frac{12\gamma^2 L^2 I_{max}^2}{n} \sum_{i=1}^N\sum_{k\in\mathcal{V}_i}\E\Big\|\w_k^{\tau}
- \bar \w^{\tau}_i \Big\|^2,
\end{align}
where $I_{max} = \max_i I_i$. 
Suppose $1-12\gamma^2L^2I_{max}^2 > 0$, we have
\begin{align}\label{E22}
&\frac{1}{T}\sum_{t=0}^{T-1}\frac{1}{n}\sum_{i=1}^N  \sum_{k\in\mathcal{V}_i}\E \| \bar \w_i^t-\w_{k}^t \|^2 \notag \\
&\le \frac{2\gamma^2\sigma^2 \sum_{i=1}^N \frac{(n_i-1)I_i}{n}}{1-12\gamma^2L^2I_{max}^2} + \frac{6\gamma^2 \sum_{i=1}^N\frac{n_i}{n} I_i^2\epsilon_i^2}{1-12\gamma^2L^2I_{max}^2}.  
\end{align}

\subsubsection{Obtaining the Final Result}

Substituting (\ref{E22}) into (\ref{E15}), we obtain an upper bound on the upward parameter MSE as
\begin{align}\label{E23}
&\frac{1}{T}\sum_{t=0}^{T-1} \sum_{l=1}^N\frac{n_l}{n}\E\| \bar \w^t-\bar \w_l^t\|^2 \notag \\
&\le \frac{4\gamma^2G\frac{N-1}{n}\sigma^2}{1-12\gamma^2 L^2 G^2}
+ \frac{6\gamma^2G^2\epsilon^2}{1-12\gamma^2 L^2 G^2} \notag \\
&\quad\quad\quad + \frac{4\gamma^2G^2L^2}{1-12\gamma^2 L^2 G^2}\Bigg(\frac{2\gamma^2\sigma^2 \sum_{i=1}^N \frac{(n_i-1)I_i}{n}}{1-12\gamma^2L^2I_{max}^2} + \frac{6\gamma^2 \sum_{i=1}^N\frac{n_i}{n} I_i^2\epsilon_i^2}{1-12\gamma^2L^2I_{max}^2} \Bigg).
\end{align}
Taking the average over time and substitute (\ref{E22}) and (\ref{E23}) to (\ref{final}), we obtain the final convergence upper bound:
\begin{align}\label{result}
&\frac{1}{T}\sum_{t=0}^{T-1} \E \|\nabla f(\bar \w^t)\|^2   \notag \\
&\le \frac{2}{\gamma T}\Big(\E f(\bar \w^0)- f^*\Big) + \frac{\gamma L \sigma^2}{n}\notag \\
&\quad\quad\quad +\frac{8\gamma^2L^2G\frac{N-1}{n}\sigma^2}{1-12\gamma^2 L^2 G^2}
+ \frac{12\gamma^2L^2G^2\epsilon^2}{1-12\gamma^2 L^2 G^2} \notag \\ 
&\quad\quad\quad +\Bigg(1+ \frac{4\gamma^2G^2L^2}{1-12\gamma^2 L^2 G^2}\Bigg)\Bigg(\frac{4\gamma^2L^2\sigma^2 \sum_{i=1}^N \frac{(n_i-1)I_i}{n}}{1-12\gamma^2L^2I_{max}^2} + \frac{12\gamma^2L^2 \sum_{i=1}^N\frac{n_i}{n} I_i^2\epsilon_i^2}{1-12\gamma^2L^2I_{max}^2} \Bigg).
\end{align}

When $0<\gamma \le \frac{1}{2\sqrt{6}GL}$, we have
\begin{equation}
0<\frac{4\gamma^2G^2L^2}{1-12\gamma^2 L^2 G^2} \le \frac{2}{3},
\label{eq:c1}
\end{equation}
and 
\begin{equation}
0<\frac{1}{1-12\gamma^2L^2I_{max}^2} \le \frac{1}{1-12\gamma^2L^2G^2} \le 2. 
\label{eq:c2}
\end{equation}
By substituting (\ref{eq:c1}) and (\ref{eq:c2}) into (\ref{E23}) of the main paper and  setting
$C=\frac{40}{3}$, we obtain the upper bound in Theorem~\ref{thm1}. Note that the constant in front of the term $\sum_{i=1}^N\frac{n_i}{n} \Big(1-\frac{1}{n_i}\Big)I_i$ is intentionally increased by a factor of $2$ in order to better match the constants given in Theorem~\ref{thm1}.

\clearpage

\section{Proofs for the Two-level Random Grouping Case}

In the following, we will present the proofs of Lemmas~\ref{lemma1} and \ref{lemma2} and Theorem~\ref{thm2}. %
As discussed in Section~\ref{sec: HF-SGD random grouping} of the main paper, we consider uniformly random grouping here where the groups $\{\mathcal{V}_i:i=1,2,\ldots ,N\}$ are random sets of equal size, i.e., $n_i = |\mathcal{V}_i| = \frac{n}{N}$ for all $i$. For simplicity, we use $K := \frac{n}{N} = n_i$ to denote the size of each group. The uniformly random partitioning of workers into groups is equivalent to the following problem. Assume that there are $n$ bins, where the first $K$ bins contain workers in group $i=1$, the next $K$ bins contain workers in group $i=2$, and so on. We randomly and uniformly assign the $n$ workers to the $n$ bins, such that each bin contains one (and only one) worker and each worker is assigned to one (and only one) bin. A random realization $\mathsf{S}$ corresponds to a mapping between workers and bins.

Let $\mathsf{k}_j$ denote the worker assigned to bin $j$, where $\mathsf{k}_j$ is a random variable due to random assignment. Also let $[K] := \{1,2,\ldots ,K\}$ and $[n] := \{1,2,\ldots ,n\} = \mathcal{V}$ (the last equality is by definition of $\mathcal{V}$). We proceed with the proofs in the following.

We define
\begin{equation}
\epsilon_\mathbf{w}^2 := \E_{\mathsf{S}} \Big( \| \nabla f(\mathbf{w}) - \nabla F_{\mathsf{k}_j}(\mathbf{w}) \|^2 \Big) \overset{(a)}{=} \frac{1}{n}\sum_{k=1}^n\| \nabla f(\w) - \nabla F_k(\w)\|^2  \overset{(b)}{\le} \tilde{\epsilon}^2.
\label{eq:epsilon_w}
\end{equation}
where $(a)$ is because for any given bin $j$, the probability of any worker being assigned to it is $\frac{1}{n}$, and $(b)$ is due to Assumption~\ref{assumption:divergence-sf-sgd} in the main paper.

\subsection{Proof of Lemma~\ref{lemma1}}
We have 
\begin{align}\label{s1}
&\E_{\mathsf{S}} \left[\sum_{i=1}^N\frac{n_i}{n} \bigg\| \nabla f(\mathbf{w}) - \frac{1}{K}\sum_{k\in\mathcal{V}_i} \nabla F_{k}(\mathbf{w}) \bigg\|^2\right]\notag \\
&= \frac{1}{N} \sum_{i=1}^N  \E_{\mathsf{S}}  \bigg\| \nabla f(\mathbf{w}) - \frac{1}{K}\sum_{k\in\mathcal{V}_i} \nabla F_{k}(\mathbf{w}) \bigg\|^2 \notag \\
& \overset{(a)}{=}  \E_{\mathsf{S}} \bigg\| \nabla f(\mathbf{w}) - \frac{1}{K}\sum_{k\in\mathcal{V}_1} \nabla F_{k}(\mathbf{w})\bigg \|^2\notag \\
& \overset{(b)}{=}  \E_{\mathsf{S}} \bigg\| \nabla f(\mathbf{w}) - \frac{1}{K}\sum_{j=1}^K \nabla F_{\mathsf{k}_j}(\mathbf{w})\bigg \|^2\notag \\
& = \frac{1}{K^2}  \E_{\mathsf{S}} \bigg\|  \sum_{j=1}^K \left( \nabla f(\mathbf{w}) - \nabla F_{\mathsf{k}_j}(\mathbf{w})\right)\bigg \|^2\notag \\
&= \frac{1}{K^2} \E_{\mathsf{S}} \Big( \sum_{j=1}^K\| \nabla f(\mathbf{w}) - \nabla F_{\mathsf{k}_j}(\mathbf{w}) \|^2 \Big) + \frac{1}{K^2}  \E_{\mathsf{S}} \Big( \sum_{\substack{j,j' \in [K]\\ j \neq j'}} \Big \langle \nabla f(\w) - \nabla F_{\mathsf{k}_{j}}(\w), \nabla f(\w) - \nabla F_{\mathsf{k}_{j'}}(\w) \Big \rangle \Big),  
\end{align}
where we note that $\mathcal{V}_i$ is a random set of workers in group $i$ and $\mathsf{k}_j$ is a random worker index assigned to the $j$-th bin, both of which capture the outcome of random grouping. Equality $(a)$ is due to symmetry, so that the expectation involving any group $i$ is equal to the expectation involving the first group $i=1$. Equality $(b)$ is from the fact that the first group $\mathcal{V}_1$ includes workers in bins $j=1,2,\ldots ,K$, hence $\sum_{k\in\mathcal{V}_1} \nabla F_{k}(\mathbf{w}) = \sum_{j=1}^K \nabla F_{\mathsf{k}_j}(\mathbf{w})$. We analyze the two terms in (\ref{s1}) in the following.

For the first term in (\ref{s1}), we have
\begin{align}
\frac{1}{K^2} \E_{\mathsf{S}} \Big( \sum_{j=1}^K\| \nabla f(\mathbf{w}) - \nabla F_{\mathsf{k}_j}(\mathbf{w}) \|^2 \Big) & = \frac{1}{K^2} \sum_{j=1}^K \E_{\mathsf{S}} \Big( \| \nabla f(\mathbf{w}) - \nabla F_{\mathsf{k}_j}(\mathbf{w}) \|^2 \Big)  = \frac{\epsilon_\mathbf{w}^2}{K}, 
\label{eq:proof_lemma1_first_term}
\end{align}
where the last inequality is from (\ref{eq:epsilon_w}).

For the second term in (\ref{s1}), we have
\begin{align}
& \frac{1}{K^2}  \E_{\mathsf{S}} \Big( \sum_{\substack{j,j' \in [K]\\ j \neq j'}} \Big \langle \nabla f(\w) - \nabla F_{\mathsf{k}_{j}}(\w), \nabla f(\w) - \nabla F_{\mathsf{k}_{j'}}(\w) \Big \rangle \Big) \notag \\
& = \frac{1}{K^2}   \sum_{\substack{j,j' \in [K]\\ j \neq j'}} \E_{\mathsf{S}} \Big \langle \nabla f(\w) - \nabla F_{\mathsf{k}_{j}}(\w), \nabla f(\w) - \nabla F_{\mathsf{k}_{j'}}(\w) \Big \rangle  \notag \\
& \overset{(a)}{=} \frac{1}{K^2}  \sum_{\substack{j,j' \in [K]\\ j \neq j'}}  \sum_{\substack{k,k' \in [n]\\ k\neq k'}}\frac{1}{n(n-1)} \Big \langle \nabla f(\w) - \nabla F_{k}(\w), \nabla f(\w) - \nabla F_{k'}(\w) \Big \rangle  \notag \\
& = \frac{1}{K^2}  \sum_{\substack{j,j' \in [K]\\ j \neq j'}} \frac{1}{n(n-1)} \Bigg[\Big\|\sum_{k=1}^n\Big(\nabla f(\w) - \nabla F_k(\w) \Big)\Big\|^2 - \sum_{k=1}^n\Big\| \nabla f(\w) - \nabla F_k(\w)\Big\|^2 \Bigg]  \notag \\
& \overset{(b)}{=} -\frac{1}{K^2}  \sum_{\substack{j,j' \in [K]\\ j \neq j'}} \frac{1}{n(n-1)} \sum_{k=1}^n\Big\| \nabla f(\w) - \nabla F_k(\w)\Big\|^2  \notag \\
& = -\frac{K-1}{K} \cdot \frac{1}{n-1}\cdot\frac{1}{n} \sum_{k=1}^n\Big\| \nabla f(\w) - \nabla F_k(\w)\Big\|^2  \notag \\
& \overset{(c)}{=} -\frac{K-1}{K(n-1)} \epsilon_\mathbf{w}^2,
\label{eq:proof_lemma1_second_term}
\end{align}
where $(a)$ is because the probability of assigning worker $k$ to bin $j$ and at the same time a different worker $k'$ to a different bin $j'$ is $\frac{1}{n(n-1)}$, $(b)$ is because $\sum_{k=1}^n\Big(\nabla f(\w) - \nabla F_k(\w) \Big) = 0$ according to the definition in (\ref{eq:objective}) and the fact that $n_i$ is equal for all $i$, and $(c)$ is from (\ref{eq:epsilon_w}).

Substituting (\ref{eq:proof_lemma1_first_term}) and (\ref{eq:proof_lemma1_second_term}) into (\ref{s1}), we obtain
\begin{align}
&\E_{\mathsf{S}} \left[\sum_{i=1}^N\frac{n_i}{n} \bigg\| \nabla f(\mathbf{w}) - \frac{1}{K}\sum_{k\in\mathcal{V}_i} \nabla F_{k}(\mathbf{w})\bigg\|^2\right]\notag \\
&=  \frac{1}{K}\Bigg[ \epsilon_{\w}^2 - \frac{K-1}{n-1} \epsilon_{\mathbf{w}}^2\Bigg] \notag \\
&= \frac{1}{K}\left(\frac{n-K}{n-1}\right) \epsilon_{\mathbf{w}}^2 = \Big(\frac{N-1}{n-1}\Big) \epsilon_{\mathbf{w}}^2
 \overset{(a)}{\le} \Big(\frac{N-1}{n-1}\Big) \tilde{\epsilon}^2,
 \label{eq:lemma1-final-result}
\end{align}
where $(a)$ is from (\ref{eq:epsilon_w}). This proves the lemma.

\subsection{Proof of Lemma~\ref{lemma2}}
We have 
\begin{align}
& \E_{\mathsf{S}}\left[\frac{1}{K} \sum_{k\in\mathcal{V}_i} \bigg \|\nabla F_{k}(\mathbf{w}) - \frac{1}{K} \sum_{k' \in\mathcal{V}_i} \nabla F_{k'}(\mathbf{w}) \bigg\|^2 \right]\notag\\
& \overset{(a)}{=} \E_{\mathsf{S}}\left[\frac{1}{K} \sum_{k\in\mathcal{V}_1} \bigg \|\nabla F_{k}(\mathbf{w}) - \frac{1}{K} \sum_{k' \in\mathcal{V}_1} \nabla F_{k'}(\mathbf{w}) \bigg\|^2 \right]\notag\\
& \overset{(b)}{=} \E_{\mathsf{S}}\left[\frac{1}{K} \sum_{j=1}^K \bigg \|\nabla F_{\mathsf{k}_j}(\mathbf{w}) - \frac{1}{K} \sum_{j' =1}^K \nabla F_{\mathsf{k}_{j'}}(\mathbf{w}) \bigg\|^2 \right]\notag\\
& = \frac{1}{K} \sum_{j=1}^K \E_{\mathsf{S}}\bigg \|\nabla F_{\mathsf{k}_j}(\mathbf{w}) - \nabla f(\mathbf{w})+\nabla f(\mathbf{w}) - \frac{1}{K} \sum_{j' =1}^K \nabla F_{\mathsf{k}_{j'}}(\mathbf{w}) \bigg\|^2 \notag\\
&= \frac{1}{K} \sum_{j=1}^K  \E_{\mathsf{S}}\left[\bigg \| \nabla F_{\mathsf{k}_j}(\mathbf{w}) - \nabla f(\mathbf{w})\bigg \|^2 \right] 
+ \frac{1}{K} \sum_{j=1}^K \E_{\mathsf{S}} \left\|\nabla f(\mathbf{w}) -\frac{1}{K} \sum_{j' =1}^K \nabla F_{\mathsf{k}_{j'}}(\mathbf{w})\right\|^2 \notag \\
&\quad\quad\quad + \frac{2}{K}\sum_{j=1}^K\E_{\mathsf{S}} \left\langle\nabla F_{\mathsf{k}_j}(\mathbf{w})-\nabla f(\mathbf{w}),
\nabla f(\mathbf{w}) -\frac{1}{K} \sum_{j' =1}^K \nabla F_{\mathsf{k}_{j'}}(\mathbf{w}) \right\rangle \notag \\
&= \frac{1}{K} \sum_{j=1}^K  \E_{\mathsf{S}}\left[\bigg \| \nabla F_{\mathsf{k}_j}(\mathbf{w}) - \nabla f(\mathbf{w})\bigg \|^2 \right] 
+ \E_{\mathsf{S}} \left\|\nabla f(\mathbf{w}) -\frac{1}{K} \sum_{j' =1}^K \nabla F_{\mathsf{k}_{j'}}(\mathbf{w})\right\|^2 \notag \\
&\quad\quad\quad + 2\E_{\mathsf{S}} \left\langle\frac{1}{K}\sum_{j=1}^K\nabla F_{\mathsf{k}_j}(\mathbf{w})-\nabla f(\mathbf{w}),
\nabla f(\mathbf{w}) -\frac{1}{K} \sum_{j' =1}^K \nabla F_{\mathsf{k}_{j'}}(\mathbf{w}) \right\rangle \notag \\
&= \frac{1}{K} \sum_{j=1}^K  \E_{\mathsf{S}}\left[\bigg \| \nabla F_{\mathsf{k}_j}(\mathbf{w}) - \nabla f(\mathbf{w})\bigg \|^2 \right] 
- \E_{\mathsf{S}} \left\|\nabla f(\mathbf{w}) -\frac{1}{K} \sum_{j' =1}^K \nabla F_{\mathsf{k}_{j'}}(\mathbf{w})\right\|^2 \notag \\
&\overset{(c)}{=} \epsilon_{\mathbf{w}}^2 - \Big(\frac{N-1}{n-1}\Big)\epsilon_{\mathbf{w}}^2 \notag \\
&\overset{(d)}{\leq} \left(1-\frac{N-1}{n-1}\right)\tilde{\epsilon}^2,
\end{align}
where $(a)$ is due to symmetry of groups so that the expectation over $\mathcal{V}_i$ is equal to the expectation over $\mathcal{V}_1$, $(b)$ is because the workers in group $i=1$ are in the first $K$ bins, the first term in $(c)$ is from (\ref{eq:epsilon_w}), the second term in $(c)$ is from the intermediate result in (\ref{eq:lemma1-final-result}) of Lemma~\ref{lemma1}'s proof and also the symmetry of groups, and $(d)$ is from (\ref{eq:epsilon_w}) and $1-\frac{N-1}{n-1} \geq 0$.

\subsection{Proof of Theorem~\ref{thm2}}
\label{sec:proof-of-thm2}
From Lemma~\ref{lemma1} and Lemma~\ref{lemma2}, we have the expected global divergence and local divergence:
\begin{equation}
\E_{\mathsf{S}} \left[\sum_{i=1}^N\frac{n_i}{n} \| \nabla f_i(\mathbf{w}) - \nabla f(\w) \|^2\right] \le \Big(\frac{N-1}{n-1}\Big) \tilde{\epsilon}^2,  \forall \w, 
\label{eq:div_global}
\end{equation}
\begin{equation}
\E_{\mathsf{S}}\left[\frac{1}{n_i}\sum_{k\in \mathcal{V}_i}\|\nabla F_{k}(\mathbf{w}) - \nabla f_i(\mathbf{w})\|^2\right] \le \left(1- \frac{N-1}{n-1}\right)\tilde{\epsilon}^2, \forall \w, \forall i
\label{eq:div_local}
\end{equation}
where we recall that $n_i = K = \frac{n}{N}$ and $f_i(\mathbf{w})=\frac{1}{n_i}\sum_{j \in \mathcal{V}_i} F_{j}(\mathbf{w})$ by definition.

The proof of Theorem~\ref{thm2} follows similar steps as  that of Theorem~\ref{thm1}, with the following differences: 1) we take an additional expectation over $\mathsf{S}$ (i.e., all possible groupings) and consider the expectations currently in Theorem~\ref{thm1}, and their proofs as being conditioned on a given groping realization $\mathsf{S}$; 2) we substitute (\ref{eq:div_global}) and (\ref{eq:div_local}) into 
(\ref{e15}) and (\ref{E19}) to replace the original upper bounds $\epsilon^2$ and $\epsilon_i^2$ with the expected divergence upper bounds $\Big(\frac{N-1}{n-1}\Big) \tilde{\epsilon}^2$ and $\left(1- \frac{N-1}{n-1}\right)\tilde{\epsilon}^2$, respectively. Due to the  similarity, we do not write the full proof of Theorem~\ref{thm2}, and refer readers to the proofs of Theorem~\ref{thm1} while noting the above minor changes. 

\clearpage

\section{Extension to the Multi-Level Case
(\texorpdfstring{$M\geq 2$}{M >= 2}) }

\label{appendix:MultiLevel}

The multi-level SGD algorithm is shown in Algorithm~\ref{alg:multi-level-SGD}. 

\begin{algorithm}[h]
\small
 \caption{Multi-level H-SGD}
 \label{alg:multi-level-SGD}
\KwIn{$\gamma$, $\bar{\mathbf{w}}^0$, $\{P_i : i\in \{1,2,\ldots ,M\}\}$}
\KwOut{Global aggregated model $\mathbf{\bar w}^{T}$}
\For{$t\leftarrow 0$ to $T-1$}{

\For{Each worker $(k_1,\ldots , k_M) \in \mathcal{V}$, in parallel}{
Compute $\mathbf{g}_{k_1 \ldots  k_M}(\mathbf{w}_{k_1 \ldots  k_M}^t,\zeta^t_{k_1 \ldots  k_M})$;\\
$\mathbf{w}_{k_1 \ldots  k_M}^{t+1} \leftarrow \mathbf{w}_{k_1 \ldots  k_M}^t - \gamma \mathbf{g}_{k_1 \ldots  k_M}(\mathbf{w}_{k_1 \ldots  k_M}^t,\zeta^t_{k_1 \ldots  k_M})$;\\
}
\For{level $i\leftarrow 1$ to $M$}{
\If{$P_i \mid t+1$}{
$i$-th level aggregate: $\mathbf{\bar w}^{t+1}_{k_1 \ldots  k_i} \leftarrow \frac{1}{N_i\ldots  N_M} \sum^{N_i}_{k_i=1}\ldots  \sum_{k_M=1}^{N_M} \mathbf{w}_{k_1 \ldots  k_M}^{t+1},\forall k_1,\ldots  k_M $;\\
Distribute: $\mathbf{w}_{k_1 \ldots  k_M}^{t+1} \leftarrow \mathbf{\bar w}_{k_1 \ldots  k_i}^{t+1}, \forall k_1,\ldots  k_M$;\\
\textbf{break}\tcp*{Lower level aggregation not needed when aggregated at higher level}
}
}
}
\end{algorithm}

First, we formally state an assumption for multi-level upward divergence and downward divergence as follows.
\begin{assumption}
\label{ml-sgd}
For any server at level $\ell$, the $\ell$-th level upward and downward divergences are bounded by
\begin{align}
&\frac{1}{n_\ell}\sum_{k_1=1}^{N_1}\cdots\sum_{k_\ell=1}^{N_\ell}\|\nabla f_{k_1 \ldots  k_\ell}(\mathbf{w})-\nabla f(\mathbf{w}) \|^2 \le \hat{\epsilon}_\ell^2,\label{eq:multi_level_upward_divergence}\\
&\!\!\!\!\!\frac{n_\ell}{n}\!\sum_{k_{\ell+1}=1}^{N_{\ell+1}}\!\!\!\!\cdots\!\!\!\sum_{k_M=1}^{N_M}\!\!\!\|\nabla f_{k_1 \ldots  k_\ell}(\mathbf{w})\!-\!\nabla F_{k_1 \ldots  k_M}(\mathbf{w}) \|^2 \!\le\! \hat{\epsilon}_\ell'^2 \label{eq:multi_level_downward_divergence}
\end{align}
respectively, 
$\forall\mathbf{w}$, $\forall k_1,\ldots ,k_\ell$ in (\ref{eq:multi_level_local_divergence}), where $n_\ell = \Pi_{j=1}^\ell N_j$. %
\end{assumption}

Next, we provide the theorem for multi-level H-SGD under the fixed grouping case and its proof. We will extend the proof to the uniform random grouping case in Sections~\ref{sec:proof-lemma3} and \ref{sec:proof-uniform-multi}. 

\begin{theorem}\label{thm:fixed-multi-level}
For any fixed worker grouping that satisfies Assumption~\ref{ml-sgd}, if the learning rate in Algorithm~\ref{alg:multi-level-SGD} satisfies $\gamma < \frac{1}{2\sqrt{3}P_1 L}$, then for any $T \ge 1$, we have
\begin{align*}
&\frac{1}{T}\sum_{t=0}^{T-1} \E \bigg\|\nabla f(\bar\w^t) \bigg\|^2 \notag \\
&\le  \frac{2}{\gamma T}\bigg(f^0-f^* \bigg) + \frac{\gamma L \sigma^2}{n} \notag \\
&\quad + \frac{1}{M-1}\sum_{l=1}^{M-1} \Bigg[ \frac{8\gamma^2L^2P_1\left(1-\frac{1}{\prod_{j=1}^{\ell} N_j}\right)\frac{1}{\prod_{j=\ell+1}^M N_j} \sigma^2}{1-12\gamma^2L^2P_1^2}
+ \frac{12\gamma^2L^2P_1^2\hat{\epsilon}_\ell^2}{1-12\gamma^2L^2P_1^2} \notag \\
&\quad + \bigg(1+\frac{8\gamma^2P_1^2L^2}{1-12\gamma^2L^2P_1^2} \bigg) \bigg( \frac{4\gamma^2L^2P_{\ell}\left(1-\frac{1}{\prod_{j=\ell+1}^M N_j}\right) \sigma^2}{1-12\gamma^2L^2P_{\ell}^2} + \frac{12\gamma^2L^2 P_{\ell}^2 \hat{\epsilon}'^2_{\ell}}{1-12\gamma^2L^2P_{\ell}^2}\bigg) \Bigg].    
\end{align*}
\end{theorem}

\subsection{Proof of Theorem~\ref{thm:fixed-multi-level}}
Similar to (\ref{E1})--(\ref{E2}), when $\gamma \le \frac{1}{L}$, that is, $\frac{\gamma}{2}-\frac{\gamma^2L}{2} \ge 0$, we can obtain
\begin{align*}
\E f(\bar \w^{t+1}) \le \E f(\bar \w^t) + \frac{\gamma^2L\sigma^2}{2n} -\frac{\gamma}{2}\E \|\nabla f(\bar \w^t)\|^2 
 + \frac{\gamma}{2}\E\bigg\|\nabla f(\bar \w^t) - \frac{1}{n} \sum_{i=1}^n \nabla F_i(\w_j^t)\bigg \|^2.
\end{align*}

Next, we will upper bound the term $\E\bigg\|\nabla f(\bar \w^t) - \frac{1}{n} \sum_{i=1}^n \nabla F_i(\w_j^t)\bigg \|^2$ for the multi-level case. 
In order to bound this term, we %
divide it into upward  part as in (\ref{top-down}) and downward part in (\ref{bottom-up}). The upward %
part means the difference between the global gradient and $i$-th ($1\le i \le M-1$) level averaged gradient. The downward means the difference between the gradients of workers and $i$-th ($1\le i \le M-1$) level averaged gradient, which is similar to (\ref{eq:global_local_divergence_est}). Then each part is divided into $M-1$ terms corresponding to $M-1$ levels between the global server and workers. The details are as follows. 
\begin{align}
    &\E\bigg\|\nabla f(\bar \w^t) - \frac{1}{n} \sum_{i=1}^n \nabla F_i(\w_j^t)\bigg \|^2 \notag \\
    &= \E\bigg\|\nabla f(\bar \w^t) 
    -\frac{1}{M-1} \bigg[ \frac{1}{N_1}\sum_{k_1=1}^{N_1} \nabla f_{k_1}(\bar{\w}_{k_1}^t) - \frac{1}{N_1}\sum_{k_1=1}^{N_1} \nabla f_{k_1}(\bar{\w}_{k_1}^t) + \frac{1}{N_1N_2}\sum_{k_1=1}^{N_1}\sum_{k_2 =1}^{N_2} \nabla f_{k_1k_2 }(\bar{\w}_{k_1k_2 }^t)\notag \\
    &\quad -\frac{1}{N_1N_2}\sum_{k_1=1}^{N_1}\sum_{k_2 =1}^{N_2} \nabla f_{k_1k_2 }(\bar{\w}_{k_1k_2 }^t) \notag \\
    &\quad + \ldots  \notag \\
    &\quad + \frac{1}{\prod_{j=1}^{M-1} N_j}\sum_{k_1=1}^{N_1}\ldots  \sum_{k_{M-1}=1}^{N_{M-1}} \nabla f_{k_1 \ldots  k_{M-1}}(\bar{\w}_{k_1 \ldots  k_{M-1}}^t)
    - \frac{1}{\prod_{j=1}^{M-1} N_j}\sum_{k_1=1}^{N_1}\ldots  \sum_{k_{M-1}=1}^{N_{M-1}} \nabla f_{k_1 \ldots  k_{M-1}}(\bar{\w}_{k_1 \ldots  k_{M-1}}^t)
    )\bigg] \notag \\
    &\quad - \frac{1}{\prod_{j=1}^M N_j}\sum_{k_1=1}^{N_1}  \ldots  \sum_{k_M=1}^{N_M}  \nabla F_{k_1 \ldots  k_M}(\w_{k_1 \ldots  k_M}^t) \bigg \|^2 \notag \\
    &\overset{(a)}{\le} 2\E \bigg\| \frac{1}{M-1} \bigg[\nabla f(\bar \w^t) - \frac{1}{N_1}\sum_{k_1=1}^{N_1} \nabla f_{k_1}(\bar{\w}_{k_1}^t) + \nabla f(\bar \w^t) - \frac{1}{N_1N_2}\sum_{k_1=1}^{N_1}\sum_{ k_2=1}^{N_2} \nabla f_{k_1k_2 }(\bar{\w}_{k_1k_2 }^t) + \ldots  \notag  \\
    &\quad + \nabla f(\bar \w^t) -  \frac{1}{\prod_{j=1}^{M-1} N_j}\sum_{k_1=1}^{N_1}\ldots  \sum_{k_{M-1}=1}^{N_{M-1}} \nabla f_{k_1 \ldots  k_{M-1}}(\bar{\w}_{k_1 \ldots  k_{M-1}}^t) \bigg] \bigg\|^2  %
    \label{top-down} 
    \\
    &\quad + 2\E \bigg\|\frac{1}{M-1}\bigg[\frac{1}{N_1}\sum_{k_1=1}^{N_1} \nabla f_{k_1}(\bar{\w}_{k_1}^t) - \frac{1}{\prod_{j=1}^M N_j}\sum_{k_1=1}^{N_1} \ldots  \sum_{k_M=1}^{N_M}  \nabla F_{k_1 \ldots  k_M}(\w_{k_1 \ldots  k_M}^t) \notag \\ 
    &\quad  + \frac{1}{N_1N_2}\sum_{k_1=1}^{N_1}\sum_{ k_2=1}^{N_2} \nabla f_{k_1k_2 }(\bar{\w}_{k_1k_2}^t)  - \frac{1}{\prod_{j=1}^M N_j}\sum_{k_1=1}^{N_1}  \ldots  \sum_{k_M=1}^{N_M}  \nabla F_{k_1 \ldots  k_M}(\w_{k_1 \ldots  k_M}^t) + \ldots  %
    \notag\\
    &\quad +\frac{1}{\prod_{j=1}^{M-1} N_j}\sum_{k_1=1}^{N_1}  \ldots    \sum_{k_{M-1}=1}^{N_{M-1}} \nabla f_{k_1 \ldots  k_{M-1}}(\bar{\w}_{k_1 \ldots  k_{M-1}}^t) 
    - \frac{1}{\prod_{j=1}^M N_j}\sum_{k_1=1}^{N_1} \ldots  \sum_{k_M=1}^{N_M}  \nabla F_{k_1 \ldots  k_M}(\w_{k_1 \ldots  k_M}^t)  \bigg] \bigg\|^2   \label{bottom-up} \\
    &\overset{(b)}{\le} \frac{1}{M-1}\bigg[ 2\E\bigg\|\nabla f(\bar \w^t) - \frac{1}{N_1}\sum_{k_1=1}^{N_1} \nabla f_{k_1}(\bar{\w}_{k_1}^t) \bigg\|^2 \notag \\
    &\quad + 2\E\bigg\|\frac{1}{N_1}\sum_{k_1=1}^{N_1} \nabla f_{k_1}(\bar{\w}_{k_1}^t) - \frac{1}{\prod_{j=1}^M N_j}\sum_{k_1=1}^{N_1}  \ldots  \sum_{k_M=1}^{N_M}  \nabla F_{k_1 \ldots  k_M}(\w_{k_1 \ldots  k_M}^t) \bigg\|^2 \bigg] \notag \\
    &\quad +\frac{1}{M-1} \bigg[ 2\E\bigg\|\nabla f(\bar \w^t) - \frac{1}{N_1N_2}\sum_{k_1=1}^{N_1}\sum_{k_2=1}^{N_2} \nabla f_{k_1k_2 }(\bar{\w}_{k_1k_2}^t) \bigg\|^2 \notag \\ &\quad +2\E\bigg\|\frac{1}{N_1N_2}\sum_{k_1=1}^{N_1}\sum_{k_2=1}^{N_2} \nabla f_{k_1k_2}(\bar{\w}_{k_1k_2}^t) - \frac{1}{\prod_{j=1}^M N_j}\sum_{k_1=1}^{N_1}  \ldots  \sum_{k_M=1}^{N_M}  \nabla F_{k_1 \ldots  k_M}(\w_{k_1 \ldots  k_M}^t) \bigg\|^2 \bigg] \notag \\
    &\quad + \ldots  \notag \\
    &\quad + \frac{1}{M-1}\bigg[ 2\E\bigg\|\nabla f(\bar \w^t) - \frac{1}{\prod_{j=1}^{M-1} N_j}\sum_{k_1=1}^{N_1}  \ldots     \sum_{k_{M-1}=1}^{N_{M-1}} \nabla f_{k_1 \ldots  k_{M-1}}(\bar{\w}_{k_1 \ldots  k_{M-1}}^t) \bigg\|^2 \notag \\
    &\quad + 2\E\bigg\|\frac{1}{\prod_{j=1}^{M-1} N_j}\sum_{k_1=1}^{N_1}  \ldots     \sum_{k_{M-1}=1}^{N_{M-1}} \nabla f_{k_1 \ldots  k_{M-1}}(\bar{\w}_{k_1 \ldots  k_{M-1}}^t) 
    - \frac{1}{\prod_{j=1}^M N_j}\sum_{k_1=1}^{N_1}  \ldots  \sum_{k_M=1}^{N_M}  \nabla F_{k_1 \ldots  k_M}(\w_{k_1 \ldots  k_M}^t) \bigg\|^2 \bigg], \label{multi-layer-decomposition}
\end{align}
where $(a)$ and $(b)$ are due to (\ref{l1}).\\
For any level $\ell$, $1 \leq \ell \leq M-1$, we have
\begin{align}\label{multi-layer-divergence1}
&\E\bigg\|\nabla f(\bar \w^t) - \frac{1}{\prod_{j=1}^{\ell} N_j}\sum_{k_1=1}^{N_1} \ldots  \sum_{k_{\ell}=1}^{N_{\ell}} \nabla f_{k_1 \ldots  k_{M-1}}(\bar{\w}_{k_1 \ldots  k_{M-1}}^t) \bigg\|^2 \notag \\
&\overset{(a)}{\leq} L^2 \frac{1}{\prod_{j=1}^{\ell} N_j}\sum_{k_1=1}^{N_1} \ldots   \sum_{k_{\ell}=1}^{N_{\ell}}\E \|\bar{\w}^t - \bar{\w}_{k_1 \ldots  k_{\ell}}^t \|^2
\end{align}
and
\begin{align}\label{multi-layer-divergence2}
&\E\bigg\|\frac{1}{\prod_{j=1}^{\ell} N_j}\sum_{k_1=1}^{N_1}  \ldots  \sum_{k_{\ell}=1}^{N_{\ell}} \nabla f_{k_1 \ldots  k_{M-1}}(\bar{\w}_{k_1 \ldots  k_{M-1}}^t) - \frac{1}{\prod_{j=1}^M N_j}\sum_{k_1=1}^{N_1}  \ldots  \sum_{k_M=1}^{N_M}  \nabla F_{k_1 \ldots  k_M}(\w_{k_1 \ldots  k_M}^t) \bigg\|^2 \notag \\
&\overset{(b)}{\leq} L^2 \frac{1}{\prod_{j=1}^M N_j}\sum_{k_1=1}^{N_1}  \ldots  \sum_{k_{\ell}=1}^{N_{\ell}} \sum_{k_{\ell+1}=1}^{N_{\ell+1}}\ldots  \sum_{k_M=1}^{N_M} \E\|\bar{\w}_{k_1 \ldots  k_{\ell}}^t - \w_{k_1  \ldots  k_{\ell}\ldots  k_M}^t  \|^2,
\end{align}
where $(a)$ and $(b)$ are due to (\ref{l1}) and Lipschitz gradient assumption.
Similar to Sections~\ref{global-mse} and \ref{local-mse}, we will bound averaged (\ref{multi-layer-divergence1}) and (\ref{multi-layer-divergence2}) over time, denoted as the upward  parameter MSE and downward parameter MSE in multi-level case, respectively. 

\subsubsection{Bounding Upward Parameter MSE} 
In this section, we will provide a general way to bound the upward parameter MSE for level $\ell$.
\begin{align}\label{G_MSE}
&\frac{1}{\prod_{j=1}^{\ell} N_j}\sum_{k_1=1}^{N_1}   \ldots  \sum_{k_{\ell}=1}^{N_{\ell}}\E \|\bar{\w}^t - \bar{\w}_{k_1 \ldots  k_{\ell}}^t \|^2 \notag \\
&=  \frac{\gamma^2}{\prod_{j=1}^{\ell}N_j}\sum_{k_1=1}^{N_1}   \ldots  \sum_{k_{\ell}=1}^{N_{\ell}}\E \bigg\| \sum_{\tau=aP_1}^{aP_1+P_1-1} \bigg( \frac{1}{\prod_{j=\ell+1}^M N_j} \sum_{k_{\ell+1}=1}^{N_{\ell+1}}\ldots  \sum_{k_M=1}^{N_M} \g_{k_1 \ldots k_M}(\w_{k_1 \ldots k_M}^\tau) \notag \\
&\quad - \frac{1}{\prod_{j=1}^M N_j}\sum_{k'_1=1}^{N_1}   \ldots  \sum_{k'_M=1}^{N_M} \g_{k'_1\ldots  k'_M}(\w_{k'_1\ldots  k'_M}^\tau)  \bigg)  \bigg\|^2 \notag \\
&\overset{(a)}{\le}  \frac{2\gamma^2}{\prod_{j=1}^{\ell}N_j}\sum_{k_1=1}^{N_1}   \ldots  \sum_{k_{\ell}=1}^{N_{\ell}}\E \bigg\|  \sum_{\tau=aP_1}^{aP_1+P_1-1}\bigg[ \bigg( \frac{1}{\prod_{j=\ell+1}^M N_j} \sum_{k_{\ell+1}=1}^{N_{\ell+1}}\ldots  \sum_{k_M=1}^{N_M} \g_{k_1 \ldots k_M}(\w_{k_1 \ldots k_M}^\tau)-\nabla f_{k_1 ..k_{\ell}}(\bar{\w}_{k_1 \ldots k_{\ell}}^\tau) \bigg)  \notag \\
&\quad -\bigg( \frac{1}{\prod_{j=1}^M N_j}\sum_{k'_1=1}^{N_1}   \ldots  \sum_{k'_M=1}^{N_M}\g_{k'_1\ldots  k'_M}(\w_{k'_1\ldots  k'_M}^\tau) - \frac{1}{\prod_{j=1}^M N_j}\sum_{k'_1=1}^{N_1}   \ldots  \sum_{k'_M=1}^{N_M} \nabla f_{k'_1\ldots  k'_M} (\bar{\w}_{k'_1\ldots  k'_M}^\tau) \bigg) \bigg] \bigg\|^2\notag \\
&\quad +  \frac{2\gamma^2}{\prod_{j=1}^{\ell}N_j}\sum_{k_1=1}^{N_1}   \ldots  \sum_{k_{\ell}=1}^{N_{\ell}}\E \bigg\|\sum_{\tau=aP_1}^{aP_1+P_1-1}\bigg(\frac{1}{\prod_{j=1}^M N_j}\sum_{k'_1=1}^{N_1}   \ldots  \sum_{k'_M=1}^{N_M} \nabla f_{k'_1\ldots  k'_M} (\bar{\w}_{k'_1\ldots  k'_M}^\tau) - \nabla f_{k_1 ..k_{\ell}}(\bar{\w}_{k_1 \ldots k_{\ell}}^\tau)\bigg) \bigg\|^2, 
\end{align}
where $(a)$ is due to (\ref{l1}).
Similar to (\ref{e13}), for the first term of (\ref{G_MSE}), we have
\begin{align}\label{global-param-noise-decom}
&\frac{2\gamma^2}{\prod_{j=1}^{\ell}N_j}\sum_{k_1=1}^{N_1}   \ldots  \sum_{k_{\ell}=1}^{N_{\ell}}\E \bigg\|  \sum_{\tau=aP_1}^{aP_1+P_1-1}\bigg[ \bigg( \frac{1}{\prod_{j=\ell+1}^M N_j} \sum_{k_{\ell+1}=1}^{N_{\ell+1}}\ldots  \sum_{k_M=1}^{N_M} \g_{k_1 \ldots k_M}(\w_{k_1 \ldots k_M}^\tau)-\nabla f_{k_1 ..k_{\ell}}(\bar{\w}_{k_1 \ldots k_{\ell}}^\tau) \bigg)  \notag \\
&-\bigg( \frac{1}{\prod_{j=1}^M N_j}\sum_{k'_1=1}^{N_1}   \ldots  \sum_{k'_M=1}^{N_M}\g_{k'_1\ldots  k'_M}(\w_{k'_1\ldots  k'_M}^\tau) - \frac{1}{\prod_{j=1}^M N_j}\sum_{k'_1=1}^{N_1}   \ldots  \sum_{k'_M=1}^{N_M} \nabla f_{k'_1\ldots  k'_M} (\bar{\w}_{k'_1\ldots  k'_M}^\tau) \bigg) \bigg] \bigg\|^2\notag \\
& \overset{(a)}{\le}  \frac{4\gamma^2}{\prod_{j=1}^{\ell}N_j}\sum_{k_1=1}^{N_1}  \ldots  \sum_{k_{\ell}=1}^{N_{\ell}}\E \bigg\|\sum_{\tau=aP_1}^{aP_1+P_1-1}\bigg[\bigg(\frac{1}{\prod_{j=\ell+1}^M N_j} \sum_{k_{\ell+1}=1}^{N_{\ell+1}}\ldots  \sum_{k_M=1}^{N_M} \g_{k_1 \ldots k_M}(\w_{k_1 \ldots k_M}^\tau) \notag\\
&\quad - \frac{1}{\prod_{j=\ell+1}^M N_j} \sum_{k_{\ell+1}=1}^{N_{\ell+1}}\ldots  \sum_{k_M=1}^{N_M} \nabla F_{k_1 \ldots k_M}(\w_{k_1 \ldots k_M}^\tau) \bigg) \notag \\
&\quad -\bigg(\frac{1}{\prod_{j=1}^M N_j}\sum_{k'_1=1}^{N_1}   \ldots   \sum_{k'_M=1}^{N_M}\g_{k'_1\ldots  k'_M}(\w_{k'_1\ldots  k'_M}^\tau) 
- \frac{1}{\prod_{j=1}^M N_j}\sum_{k'_1=1}^{N_1}   \ldots  \sum_{k'_M=1}^{N_M}\nabla F_{k'_1\ldots  k'_M}(\w_{k'_1\ldots  k'_M}^\tau) \bigg)  \bigg]\bigg\|^2\notag \\
&\quad +  \frac{4\gamma^2}{\prod_{j=1}^{\ell}N_j}\sum_{k_1=1}^{N_1}  \ldots  \sum_{k_{\ell}=1}^{N_{\ell}} \E \bigg\| \sum_{\tau=aP_1}^{aP_1+P_1-1}\bigg(\frac{1}{\prod_{j=\ell+1}^M N_j} \sum_{k_{\ell+1}=1}^{N_{\ell+1}}\ldots  \sum_{k_M=1}^{N_M} \nabla F_{k_1 \ldots k_M}(\w_{k_1 \ldots k_M}^\tau) - \nabla f_{k_1 ..k_{\ell}}(\bar{\w}_{k_1 \ldots k_{\ell}}^\tau)  \bigg) \bigg\|^2, 
\end{align}
where $(a)$ is due to (\ref{l1}).
For the first term of (\ref{global-param-noise-decom}), we have
\begin{align}
&\frac{4\gamma^2}{\prod_{j=1}^{\ell}N_j}\sum_{k_1=1}^{N_1}  \ldots  \sum_{k_{\ell}=1}^{N_{\ell}}\E \bigg\|\sum_{\tau=aP_1}^{aP_1+P_1-1}\bigg[\bigg(\frac{1}{\prod_{j=\ell+1}^M N_j} \sum_{k_{\ell+1}=1}^{N_{\ell+1}}\ldots  \sum_{k_M=1}^{N_M} \g_{k_1 \ldots k_M}(\w_{k_1 \ldots k_M}^\tau) \notag\\
&- \frac{1}{\prod_{j=\ell+1}^M N_j} \sum_{k_{\ell+1}=1}^{N_{\ell+1}}\ldots  \sum_{k_M=1}^{N_M} \nabla F_{k_1 \ldots k_M}(\w_{k_1 \ldots k_M}^\tau) \bigg) \notag \\
&-\bigg(\frac{1}{\prod_{j=1}^M N_j}\sum_{k'_1=1}^{N_1}   \ldots   \sum_{k'_M=1}^{N_M}\g_{k'_1\ldots  k'_M}(\w_{k'_1\ldots  k'_M}^\tau) 
- \frac{1}{\prod_{j=1}^M N_j}\sum_{k'_1=1}^{N_1}   \ldots  \sum_{k'_M=1}^{N_M}\nabla F_{k'_1\ldots  k'_M}(\w_{k'_1\ldots  k'_M}^\tau) \bigg)  \bigg]\bigg\|^2\notag \\
&= \frac{4\gamma^2}{\prod_{j=1}^{\ell}N_j}\sum_{k_1=1}^{N_1}  \ldots  \sum_{k_{\ell}=1}^{N_{\ell}}\E \bigg\|\sum_{\tau=aP_1}^{aP_1+P_1-1}\bigg(\frac{1}{\prod_{j=\ell+1}^M N_j} \sum_{k_{\ell+1}=1}^{N_{\ell+1}}\ldots  \sum_{k_M=1}^{N_M} \g_{k_1 \ldots k_M}(\w_{k_1 \ldots k_M}^\tau) \notag\\
&\quad - \frac{1}{\prod_{j=\ell+1}^M N_j} \sum_{k_{\ell+1}=1}^{N_{\ell+1}}\ldots  \sum_{k_M=1}^{N_M} \nabla F_{k_1 \ldots k_M}(\w_{k_1 \ldots k_M}^\tau) \bigg) \bigg\|^2\notag \\ 
&\quad -4\gamma^2\E \bigg\|\sum_{\tau=aP_1}^{aP_1+P_1-1}\bigg(\frac{1}{\prod_{j=1}^M N_j}\sum_{k_1=1}^{N_1}  \ldots  \sum_{k_M=1}^{N_M}\g_{k_1 \ldots k_M}(\w_{k_1 \ldots k_M}^\tau)
- \frac{1}{\prod_{j=1}^M N_j}\sum_{k_1=1}^{N_1}  \ldots  \sum_{k_M=1}^{N_M}\nabla F_{k_1 \ldots k_M}(\w_{k_1 \ldots k_M}^\tau) \bigg)  \bigg\|^2 \notag \\
&=  4\gamma^2\left(1-\frac{1}{\prod_{j=1}^{\ell} N_j}\right)\E \bigg\|\sum_{\tau=aP_1}^{aP_1+P_1-1}\bigg(\frac{1}{\prod_{j=\ell+1}^M N_j} \sum_{k_{\ell+1}=1}^{N_{\ell+1}}\ldots  \sum_{k_M=1}^{N_M} \g_{k_1 \ldots k_M}(\w_{k_1 \ldots k_M}^\tau)\notag \\
&\quad -\frac{1}{\prod_{j=\ell+1}^M N_j} \sum_{k_{\ell+1}=1}^{N_{\ell+1}}\ldots  \sum_{k_M=1}^{N_M} \nabla F_{k_1 \ldots k_M}(\w_{k_1 \ldots k_M}^\tau) \bigg) \bigg\|^2 \notag \\
&\overset{(a)}{\le} 4P_1\gamma^2\left(1-\frac{1}{\prod_{j=1}^{\ell} N_j}\right)\frac{1}{\prod_{j=\ell+1}^M N_j} \sigma^2, 
\end{align}
where $(a)$ is due to (\ref{l1}) and bounded variance in Assumption~\ref{assumption:hf-sgd}.
For the second term of (\ref{global-param-noise-decom}), we have
\begin{align}
&\frac{4\gamma^2}{\prod_{j=1}^{\ell}N_j}\sum_{k_1=1}^{N_1}  \ldots  \sum_{k_{\ell}=1}^{N_{\ell}} \E \bigg\| \sum_{\tau=aP_1}^{aP_1+P_1-1}\bigg(\frac{1}{\prod_{j=\ell+1}^M N_j} \sum_{k_{\ell+1}=1}^{N_{\ell+1}}\ldots  \sum_{k_M=1}^{N_M} \nabla F_{k_1 \ldots k_M}(\w_{k_1 \ldots k_M}^\tau) - \nabla f_{k_1 ..k_{\ell}}(\bar{\w}_{k_1 \ldots k_{\ell}}^\tau)  \bigg) \bigg\|^2\notag \\
&\le  \frac{4\gamma^2 L^2 P_1}{\prod_{j=1}^M N_j}\sum_{\tau=aP_1}^{aP_1+P_1-1}\sum_{k_1=1}^{N_1} \ldots  \sum_{k_M=1}^{N_M} \E \|  \w_{k_1 \ldots k_M}^\tau - \bar{\w}_{k_1\ldots k_{\ell}}^\tau \|^2, 
\end{align}
Similar to (\ref{e15}), for the second term of (\ref{G_MSE}), we have
\begin{align}
&\frac{2\gamma^2}{\prod_{j=1}^{\ell}N_j}\sum_{k_1=1}^{N_1}   \ldots  \sum_{k_{\ell}=1}^{N_{\ell}}\E \bigg\|\sum_{\tau=aP_1}^{aP_1+P_1-1}\bigg(\frac{1}{\prod_{j=1}^M N_j}\sum_{k'_1=1}^{N_1}   \ldots  \sum_{k'_M=1}^{N_M} \nabla f_{k'_1\ldots  k'_M} (\bar{\w}_{k'_1\ldots  k'_M}^\tau) - \nabla f_{k_1 ..k_{\ell}}(\bar{\w}_{k_1 \ldots k_{\ell}}^\tau)\bigg) \bigg\|^2\notag\\
&=  \frac{\gamma^2}{\prod_{j=1}^{\ell}N_j}\sum_{k_1=1}^{N_1}  \ldots  \sum_{k_{\ell}=1}^{N_{\ell}}\E \bigg\|\sum_{\tau=aP_1}^{aP_1+P_1-1}\bigg(\frac{1}{\prod_{j=1}^{\ell} N_j}\sum_{k'_1=1}^{N_1}   \ldots  \sum_{k'_l=1}^{N_{\ell}} \nabla f_{k'_1\ldots k'_l} (\bar{\w}_{k_1\ldots k_{\ell}}^\tau) - \nabla f(\bar\w^\tau)+\nabla f(\bar\w^\tau)\notag\\
&\quad -\nabla f(\bar{\w}_{k_1\ldots k_{\ell}}^\tau) + \nabla f(\bar{\w}_{k_1\ldots k_{\ell}}^\tau) - \nabla f_{k_1 ..k_{\ell}}(\bar{\w}_{k_1\ldots k_{\ell}}^\tau)\bigg) \bigg\|^2   \notag \\
&\overset{(a)}{\le}  6P_1^2\gamma^2\hat{\epsilon}_\ell^2 + \frac{12P_1\gamma^2L^2}{\prod_{j=1}^{\ell} N_j}\sum_{\tau=aP_1}^{aP_1+P_1-1}\sum_{k_1=1}^{N_1} \ldots   \sum_{k_{\ell}=1}^{N_{\ell}}\E\|\bar\w^\tau - \bar{\w}_{k_1\ldots k_{\ell}}^\tau \|^2, 
\end{align}
where $(a)$ is due to bounded divergence in Assumption~\ref{assumption:hf-sgd} and (\ref{l1}).\\
Similar to (\ref{E13})--(\ref{E15}), 
the upper bound for upward parameter MSE can be derived as follows. 
\begin{align}\label{general-top-downper-bound}
&\frac{1}{T}\sum_{t=0}^{T-1}\frac{1}{\prod_{j=1}^{\ell} N_j}\sum_{k_1=1}^{N_1}  \ldots  \sum_{k_{\ell}=1}^{N_{\ell}}\E \|\bar{\w}^t - \bar{\w}_{k_1 \ldots  k_{\ell}}^t \|^2    \notag \\
&\le  \frac{4P_1\gamma^2\left(1-\frac{1}{\prod_{j=1}^{\ell} N_j}\right)\frac{1}{\prod_{j=\ell+1}^M N_j} \sigma^2}{1-12\gamma^2L^2P_1^2}
+ \frac{6P_1^2\gamma^2\hat{\epsilon}_\ell^2}{1-12\gamma^2L^2P_1^2} \notag \\
&\quad + \frac{4\gamma^2 L^2 P_1^2}{1-12\gamma^2L^2P_1^2}\frac{1}{T}\sum_{t=0}^{T-1}\frac{1}{\prod_{j=1}^M N_j}\sum_{k_1=1}^{N_1}  \ldots  \sum_{k_M=1}^{N_M} \E \|  \w_{k_1 \ldots k_M}^\tau - \bar{\w}_{k_1\ldots k_{\ell}}^\tau \|^2.
\end{align}

\subsubsection{Bounding Downward Parameter MSE}
In this section, we will upper bound the downward parameter MSE for level $\ell$.
\begin{align}\label{general-local-parameter-mse}
&\frac{1}{\prod_{j=1}^M N_j}\sum_{k_1=1}^{N_1}  \ldots  \sum_{k_M=1}^{N_M} \E\|\bar{\w}_{k_1 \ldots  k_{\ell}}^t - \w_{k_1  \ldots  k_M}^t  \|^2    \notag \\
&=  \frac{\gamma^2}{\prod_{j=1}^M N_j}\sum_{k_1=1}^{N_1}  \ldots \sum_{k_M=1}^{N_M} \E\bigg\|\sum_{\tau=aP_{\ell}}^{aP_{\ell} + P_{\ell} -1}\bigg( \g_{k_1 \ldots  k_M}(\w_{k_1 \ldots  k_M}^\tau) \notag\\
& \quad - \frac{1}{\prod_{j=\ell+1}^M N_j} \sum_{k'_{l+1}=1}^{N_{\ell+1}}\ldots  \sum_{k'_M=1}^{N_M}\g_{k_1\ldots  k'_{l+1} \ldots  k'_M}(\w_{k_1\ldots  k'_{l+1} \ldots  k'_M}^\tau)  \bigg) \bigg\|^2 \notag \\
&\overset{(a)}{\le}  \frac{2\gamma^2}{\prod_{j=1}^M N_j}\sum_{k_1=1}^{N_1} \ldots  \sum_{k_M=1}^{N_M} \E\bigg\|\sum_{\tau=aP_{\ell}}^{aP_{\ell} + P_{\ell} -1}\bigg( \g_{k_1 \ldots  k_M}(\w_{k_1 \ldots  k_M}^\tau) - \nabla F_{k_1 \ldots  k_M}(\w_{k_1 \ldots  k_M}^\tau) \notag\\
&\quad + \frac{1}{\prod_{j=\ell+1}^M N_j} \sum_{k'_{l+1}=1}^{N_{\ell+1}}\ldots  \sum_{k'M=1}^{N_M}\nabla F_{k_1\ldots  k'_{l+1}\ldots  k'_M}(\w_{k_1\ldots  k'_{l+1}\ldots  k'_M}^\tau)  \notag\\
&\quad - \frac{1}{\prod_{j=\ell+1}^M N_j} \sum_{k'_{l+1}=1}^{N_{\ell+1}}\ldots  \sum_{k'M=1}^{N_M}\g_{k_1\ldots  k'_{l+1}\ldots  k'_M}(\w_{k_1\ldots  k'_{l+1}\ldots  k'_M}^\tau)  \bigg) \bigg\|^2 \notag \\
&\quad + \frac{2\gamma^2}{\prod_{j=1}^M N_j}\sum_{k_1=1}^{N_1} \ldots  \sum_{k_M=1}^{N_M} \E\bigg\|\sum_{\tau=aP_{\ell}}^{aP_{\ell} + P_{\ell} -1}\bigg(\nabla F_{k_1\ldots  k_M}(\w_{k_1\ldots  k_M}^\tau) \notag\\
&\quad - \frac{1}{\prod_{j=\ell+1}^M N_j} \sum_{k'_{l+1}=1}^{N_{\ell+1}}\ldots  \sum_{k'_M=1}^{N_M}\nabla F_{k_1\ldots  k'_{l+1}\ldots  k_M}(\w_{k_1\ldots  k'_{l+1}\ldots  k_M}^\tau)\bigg)\bigg\|^2
\end{align}
where $(a)$ is due to (\ref{l1}).
Similar to (\ref{E18}), for the first term of (\ref{general-local-parameter-mse}), we have
\begin{align}
&\frac{2\gamma^2}{\prod_{j=1}^M N_j}\sum_{k_1=1}^{N_1} \ldots  \sum_{k_M=1}^{N_M} \E\bigg\|\sum_{\tau=aP_{\ell}}^{aP_{\ell} + P_{\ell} -1}\bigg( \g_{k_1 \ldots  k_M}(\w_{k_1 \ldots  k_M}^\tau) - \nabla F_{k_1 \ldots  k_M}(\w_{k_1 \ldots  k_M}^\tau) \notag\\
&+ \frac{1}{\prod_{j=\ell+1}^M N_j} \sum_{k'_{l+1}=1}^{N_{\ell+1}}\ldots  \sum_{k'M=1}^{N_M}\nabla F_{k_1\ldots  k'_{l+1}\ldots  k'_M}(\w_{k_1\ldots  k'_{l+1}\ldots  k'_M}^\tau)  \notag\\
& - \frac{1}{\prod_{j=\ell+1}^M N_j} \sum_{k'_{l+1}=1}^{N_{\ell+1}}\ldots  \sum_{k'M=1}^{N_M}\g_{k_1\ldots  k'_{l+1}\ldots  k'_M}(\w_{k_1\ldots  k'_{l+1}\ldots  k'_M}^\tau)  \bigg) \bigg\|^2\notag \\
&= \frac{2\gamma^2}{\prod_{j=1}^M N_j}\sum_{k_1=1}^{N_1}  \ldots  \sum_{k_M=1}^{N_M} \E\bigg\|\sum_{\tau=aP_{\ell}}^{aP_{\ell} + P_{\ell} -1}\bigg( \g_{k_1 \ldots  k_M}(\w_{k_1 \ldots  k_M}^\tau) - \nabla F_{k_1 \ldots  k_M}(\w_{k_1 \ldots  k_M}^\tau) \bigg)\bigg\|^2   \notag \\
&\quad -  \frac{\gamma^2}{\prod_{j=1}^{\ell}N_j}\sum_{k_1=1}^{N_1} \ldots  \sum_{k_{\ell}=1}^{N_{\ell}}\E\bigg\|\sum_{\tau=aP_{\ell}}^{aP_{\ell} + P_{\ell} -1}\bigg( \frac{1}{\prod_{j=\ell+1}^M N_j} \sum_{k_{\ell+1}=1}^{N_{\ell+1}}\ldots  \sum_{k_M=1}^{N_M}\nabla F_{k_1 \ldots  k_M}(\w_{k_1 \ldots  k_M}^\tau)\notag\\
&\quad - \frac{1}{\prod_{j=\ell+1}^M N_j} \sum_{k_{\ell+1}=1}^{N_{\ell+1}}\ldots  \sum_{k_M=1}^{N_M}\g_{k_1 \ldots  k_M}(\w_{k_1 \ldots  k_M}^\tau) \bigg)\bigg\|^2 \notag \\
&\overset{(a)}{\le} 2\gamma^2P_{\ell}\left(1-\frac{1}{\prod_{j=\ell+1}^M N_j}\right) \sigma^2
\end{align}
where $(a)$ is due to bounded variance in Assumption~\ref{assumption:divergence-sf-sgd} and (\ref{l1}).
Similar to (\ref{E19}), for the second term of (\ref{general-local-parameter-mse}), we have
\begin{align}
&\frac{2\gamma^2}{\prod_{j=1}^M N_j}\sum_{k_1=1}^{N_1} \ldots  \sum_{k_M=1}^{N_M} \E\bigg\|\sum_{\tau=aP_{\ell}}^{aP_{\ell} + P_{\ell} -1}\bigg(\nabla F_{k_1\ldots  k_M}(\w_{k_1\ldots  k_M}^\tau) \notag\\
&- \frac{1}{\prod_{j=\ell+1}^M N_j} \sum_{k'_{l+1}=1}^{N_{\ell+1}}\ldots  \sum_{k'_M=1}^{N_M}\nabla F_{k_1\ldots  k'_{l+1}\ldots  k_M}(\w_{k_1\ldots  k'_{l+1}\ldots  k_M}^\tau)\bigg)\bigg\|^2\notag \\
&=\frac{2\gamma^2}{\prod_{j=1}^M N_j}\sum_{k_1=1}^{N_1}  \ldots  \sum_{k_M=1}^{N_M} \E\bigg\|\sum_{\tau=aP_{\ell}}^{aP_{\ell} + P_{\ell} -1}\bigg(\nabla F_{k_1\ldots  k_M}(\w_{k_1\ldots  k_M}^\tau) -\nabla F_{k_1\ldots  k_M}(\bar\w^\tau_{k_1\ldots  k_M})\notag \\
&\quad + \nabla F_{k_1\ldots  k_M}(\bar\w^\tau_{k_1\ldots  k_M}) - \nabla f_{k_1\ldots  k_{\ell}}(\bar\w^\tau_{k_1\ldots  k_{\ell}}) + \nabla f_{k_1\ldots  k_{\ell}}(\bar\w^\tau_{k_1\ldots  k_{\ell}}) \notag\\
&\quad - \frac{1}{\prod_{j=\ell+1}^M N_j} \sum_{k'_{l+1}=1}^{N_{\ell+1}}\ldots  \sum_{k'_M=1}^{N_M}\nabla F_{k_1\ldots  k'_{l+1} \ldots  k'_M}(\w_{k_1\ldots  k'_{l+1} \ldots  k'_M}^\tau)\bigg)\bigg\|^2   \notag \\
&\overset{(a)}{\le} 6\gamma^2 P_{\ell}^2 \hat{\epsilon}'^2_\ell + \frac{12\gamma^2L^2P_{\ell}}{\prod_{j=1}^M N_j}\sum_{\tau=aP_{\ell}}^{aP_{\ell} + P_{\ell} -1}\sum_{k_1=1}^{N_1} \ldots  \sum_{k_M=1}^{N_M} \E\|\w_{k_1 \ldots  k_M}^\tau -\bar\w^\tau_{k_1 \ldots  k_{\ell}}\|^2
\end{align}
where $(a)$ is due to bounded divergence in Assumption~{\ref{assumption:hf-sgd}} and (\ref{l1}).
Combining the above two inequalities, applying the same technique in (\ref{E20}), we can obtain
\begin{align}\label{general-loal-upper-bound}
&\frac{1}{T}\sum_{t=0}^{T-1} \frac{1}{\prod_{j=1}^M N_j}\sum_{k_1=1}^{N_1} \ldots  \sum_{k_M=1}^{N_M} \E\|\bar{\w}_{k_1 \ldots  k_{\ell}}^t - \w_{k_1 \ldots  k_M}^t  \|^2 \notag \\
&\le  \frac{2\gamma^2P_{\ell}\left(1-\frac{1}{\prod_{j=\ell+1}^M N_j}\right) \sigma^2}{1-12\gamma^2L^2P_{\ell}^2} + \frac{6\gamma^2 P_{\ell}^2 \hat{\epsilon}'^2_\ell}{1-12\gamma^2L^2P_{\ell}^2}.
\end{align}

\subsubsection{Obtaining the Final Result.}

Substituting (\ref{general-top-downper-bound}) and (\ref{general-loal-upper-bound}) into (\ref{multi-layer-decomposition}) and combining it with (\ref{E2}), since $\gamma < \frac{1}{\sqrt{12} L P_1}$, we have 
\begin{align}\label{ori-resulf-thm3}
&\frac{1}{T}\sum_{t=0}^{T-1} \E \bigg\|\nabla f(\bar\w^t) \bigg\|^2 \notag \\
&\le  \frac{2}{\gamma T}\bigg(f^0-f^* \bigg) + \frac{\gamma L \sigma^2}{n} \notag \\
&\quad + \frac{1}{M-1}\sum_{l=1}^{M-1} \Bigg[ \frac{8\gamma^2L^2P_1\left(1-\frac{1}{\prod_{j=1}^{\ell} N_j}\right)\frac{1}{\prod_{j=\ell+1}^M N_j} \sigma^2}{1-12\gamma^2L^2P_1^2}
+ \frac{12\gamma^2L^2P_1^2\hat{\epsilon}_\ell^2}{1-12\gamma^2L^2P_1^2} \notag \\
&\quad + \bigg(1+\frac{8\gamma^2P_1^2L^2}{1-12\gamma^2L^2P_1^2} \bigg) \bigg( \frac{4\gamma^2L^2P_{\ell}\left(1-\frac{1}{\prod_{j=\ell+1}^M N_j}\right) \sigma^2}{1-12\gamma^2L^2P_{\ell}^2} + \frac{12\gamma^2L^2 P_{\ell}^2 \hat{\epsilon}'^2_{\ell}}{1-12\gamma^2L^2P_{\ell}^2}\bigg) \Bigg].
\end{align}

\subsection{Proof of Lemma~\ref{lemma3}}\label{sec:proof-lemma3}
Note that by directly substituting $K$ as $K = \prod_{j=\ell+1}^M N_j$ in the proofs Lemma~\ref{lemma1} and Lemma~\ref{lemma2}, %
we extend Lemma~\ref{lemma1} and Lemma~\ref{lemma2} to the $\ell$-th level case in Lemma~\ref{lemma3} as follows:
\begin{align}\label{multi-level-random-diver1}
&\E_\mathsf{S} \bigg[\frac{1}{\prod_{j=1}^{\ell} N_j}\sum_{k_1=1}^{N_1}  \ldots  \sum_{k_{\ell}=1}^{N_{\ell}} \bigg\| \nabla f(\w) - \frac{1}{\prod_{j=\ell+1}^M N_j} \sum_{k_{\ell+1}=1}^{N_{\ell+1}}\ldots  \sum_{k_M=1}^{N_M} \nabla F_{k_1 \ldots  k_M} (\w)\bigg\|^2  \bigg]  \notag\\ %
&\le \bigg(\frac{\prod_{j=1}^{\ell} N_j-1}{n-1} \bigg)\tilde{\epsilon} \\
\label{multi-level-random-diver2}
&\E_\mathsf{S}\bigg[\frac{1}{\prod_{j=\ell+1}^M N_j} \sum_{k_{\ell +1}=1}^{N_{\ell+1}}\ldots \sum_{k_M=1}^{N_M}\bigg\| \frac{1}{\prod_{j=\ell+1}^M N_j} \sum_{k'_{\ell +1}=1}^{N_{\ell+1}}\ldots  \sum_{k'_M=1}^{N_M} \nabla F_{k_1\ldots  k'_{\ell+1}\ldots  k'_M} (\w) - \nabla F_{k_1 \ldots  k_M}(\w) \bigg\|^2 \bigg] \notag\\
&\le  \bigg(1-\frac{\prod_{j=1}^{\ell} N_j-1}{n-1} \bigg)\tilde{\epsilon}^2.
\end{align}

\subsection{Proof of Theorem \ref{thm3}}\label{sec:proof-uniform-multi}
From (\ref{ori-resulf-thm3}) to Theorem~\ref{thm3}, we first take expectation over $\mathsf{S}$ for both sides.
Then, similar to Section~\ref{sec:proof-of-thm2}, substitute $\frac{8\gamma^2P_1^2L^2}{1-12\gamma^2 L^2 P_1^2} \le \frac{2}{3}$ and $ \frac{1}{1-12\gamma^2L^2P_1^2} \le 2$ to (\ref{ori-resulf-thm3}) and
set $C=\frac{40}{3}$.
Then applying (\ref{multi-level-random-diver1}) and (\ref{multi-level-random-diver2}) instead of $\hat{\epsilon}^2_\ell$ and $\hat{\epsilon}'^2_\ell$ in (\ref{ori-resulf-thm3}), respectively, we obtain the final result.

\clearpage

\section{Additional Details and Results of Experiments}
\label{appendix:AdditionalExperiments}

In this section, we will provide additional details of our experiments and more results are shown for both two-level and multi-level (with $M=3$) cases.   

\textbf{Environment and Parameters.\,\,\,} All our experiments are implemented in PyTorch and run on a server with four NVIDIA 2080Ti GPUs. Without specific tuning, we set the learning rate $\gamma = 0.02$ for VGG-11 and $\gamma=0.1$ for CNN. 
We use an SGD mini-batch size of $20$.
Each experiment took around 6 hours.
We run each experiment 10 times then plot their average. %

For each experiment, we plot both results of training loss and test accuracy. The training loss and test accuracy are  computed over the whole training set and the whole test set, respectively. The communication time is emulated by measuring the round-trip time of transmitting the model between a worker device (in a home) and local (near) / global (far) Amazon EC2 instances, as summarized in Table~\ref{tab:actual-time}.
\\

\textbf{Datasets.\,\,\,} In our experiments, we study the performance of our algorithm on three real datasets: CIFAR-10, FEMNIST, and CelebA. The details are as follows. 
\begin{itemize}
    \item \textbf{CIFAR-10.\,\,\,} The CIFAR-10 dataset \citep{krizhevsky2009learning} consists of 60000 $32\times 32$ color images in 10 classes, with 6,000 images per class. There are 50,000 training images and 10,000 test images. There are the same number of data points on each worker. %
    This dataset (\url{https://www.cs.toronto.edu/~kriz/cifar.html}) has a citation requirement which we have satisfied.
    
    \item \textbf{FEMNIST.\,\,\,} FEMNIST is a federated version of the EMNIST dataset proposed by LEAF \citep{LEAF}, which consists of $28\times 28$ gray images in 62 classes. We use 156 workers as training data which consists of 34,659 data points and the test dataset consists of 4,973 data points. This corresponds to $5\%$ of the entire dataset (which is the one of the standard settings of LEAF), while ignoring workers that have less than $100$ data samples. This dataset is from LEAF (\url{https://github.com/TalwalkarLab/leaf}) which has a BSD 2-Clause license. %
    
    \item \textbf{CelebA.\,\,\,} The CelebA dataset \citep{liu2015faceattributes} consists of $32\times 32$ color images in binary classes. We use the data partition for CelebA proposed by LEAF \citep{LEAF}. We use 472 workers as training data which consists of 8,855 data points and the test dataset consists of 1,159 data points. This corresponds to $5\%$ of the entire dataset (which is the one of the standard settings of LEAF), while ignoring workers that have less than $5$ data samples. This dataset is also from LEAF (\url{https://github.com/TalwalkarLab/leaf}) which has a BSD 2-Clause license. %
\end{itemize}
The reason that we use $5\%$ of the FEMNIST and CelebA datasets is because processing and storing the entire dataset requires excessive computation and disk space.
We list out the statistics of the number of data points on each worker of FEMNIST and CelebA datasets in Table~\ref{tab:statistics-dataset}.

\begin{table}[ht]
\centering
\caption{Per-round communication time (ms)}
\small
\begin{tabular}{ccc}
\hline
Model 
& Local (near) EC2  & Global (far) EC2 \\
\hline
CNN & $0.29 \pm 0.14$ & $4.53 \pm 0.66$ \\ \hline
VGG-11 & $27.81 \pm 4.61$ & $291.82 \pm 15.11$ \\
\hline
\end{tabular}
\label{tab:actual-time}
\end{table}
\begin{table}[h]
\centering
\caption{Statistics of the number of data points}
\small
\begin{tabular}{>{\centering\arraybackslash}p{0.1\linewidth} | >{\centering\arraybackslash}p{0.17\linewidth} | >{\centering\arraybackslash}p{0.17\linewidth} 
>{\centering\arraybackslash}p{0.2\linewidth} 
>{\centering\arraybackslash}p{0.17\linewidth}}
\hline
\multirow{2}{*}{Dataset} & \multirow{2}{*}{Number of workers} & \multicolumn{3}{c}{Number of data points} \\
 &  & Total (all workers) & Mean (each worker)  & Standard deviation (each worker) \\
\hline
FEMNIST & 156 & 34,659 & 222 & 92 \\ \hline
CelebA &472 & 8,855 & 19 & 7 \\
\hline
\end{tabular}
\label{tab:statistics-dataset}
\end{table}

\textbf{Models.\,\,\,} We train VGG-11 models on CIFAR-10 and CelebA datasets. For FEMNIST dataset, we train a 9-layer CNN model. The structure of the CNN is $5\times 5\times 32$ Convolutional $\to$ $2 \times 2$ MaxPool $\to$ Local Response Normalization $\to$ $5 \times 5 \times 32$ Convolutional $\to$ Local Response Normalization $\to$ $2\times2$ MaxPool $\to$ $1568\times 256$ Fully connected $\to$ $256 \times 62$ Fully connected $\to$ Softmax.\\

\textbf{Data Partitioning.\,\,\,} For experiments with FEMNIST and CelebA, we use the data partition proposed by LEAF. We manually partition CIFAR-10 in a non-IID manner.  The assigned label for each worker is different. For H-SGD with $N=2$, we partition 10 workers into two groups. Group 1 contains only data with labels $\{0,1,2,3,4\}$ and Group 2  contains only data with labels $\{5,6,7,8,9\}$. In Group-IID/Group-non-IID experiments, we consider a different way of partitioning. For the ``\textit{group-IID}'' case, the $5$ workers in each group have labels $\{0,1\},\{2,3\},\{4,5\},\{6,7\},\{8,9\}$, respectively. The basic idea is to make the upward divergence nearly zero in this case. In the ``\textit{group-non-IID}'' case, workers in the first group have labels $\{0,1\},\{2,3\},\{4,5\},\{0,1\},\{2,3\}$, respectively, while workers in the second group have labels $\{4,5\},\{6,7\},\{8,9\},\{6,7\},\{8,9\}$, respectively.
Experimental results for CIFAR-10 with more workers and more groups are also shown in Figures~\ref{fig:D7}--\ref{fig:D9} later.

\subsection{Additional Results for Two-level Case}
\begin{figure}[H]
  \centering
  \begin{subfigure}{0.32\textwidth}
  \centering
  \includegraphics[width=5cm]{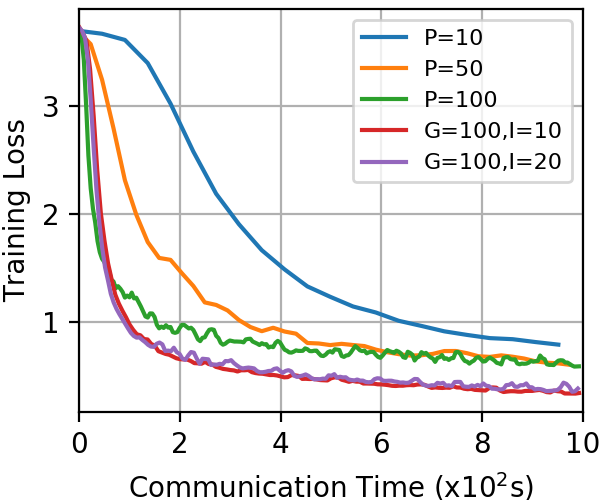}
  \caption{}
  \label{fig:femnist-training-time}
  \end{subfigure}
  \begin{subfigure}{0.32\textwidth}
  \centering
  \includegraphics[width=5cm]{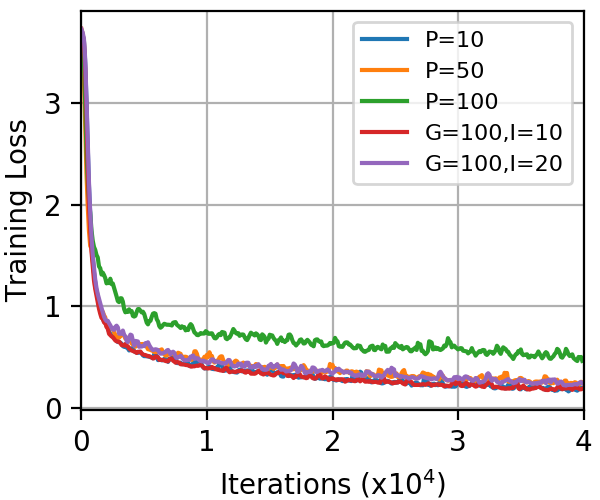}
  \caption{}
  \label{fig:femnist-training-iter}
  \end{subfigure}
  \begin{subfigure}{0.32\textwidth}
  \centering
  \includegraphics[width=5cm]{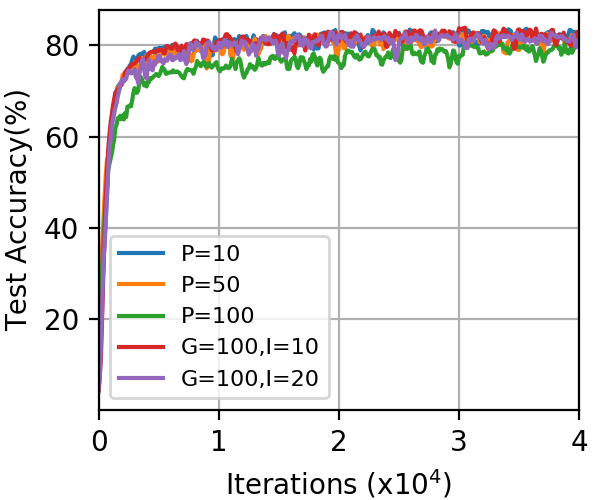}
  \label{fig:femnist-test-iter}
  \caption{}
  \end{subfigure}
  \caption{Results with FEMNIST; (a), (b) training loss v.s. communication time and iterations, (c) test accuracy v.s. iterations.}
  \label{fig:femnist-supp-results}
 \end{figure}

\begin{figure}[H]
  \centering
  \begin{subfigure}{0.32\textwidth}
  \centering
  \includegraphics[width=5cm]{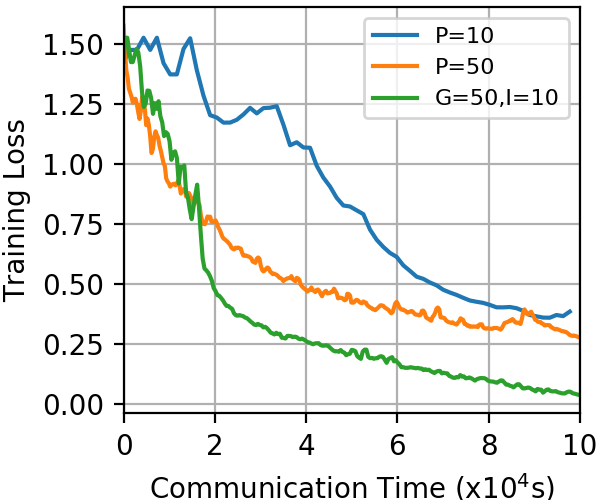}
  \caption{}
  \label{fig:celeba-training-time}
  \end{subfigure}
  \begin{subfigure}{0.32\textwidth}
  \centering
  \includegraphics[width=5cm]{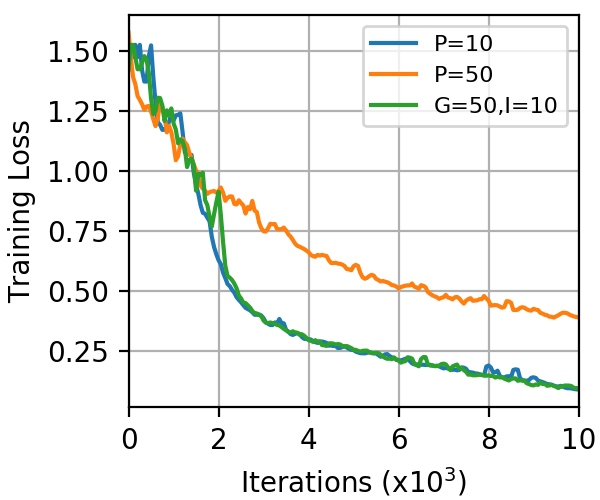}
  \caption{}
  \label{fig:celeba-training-iter}
  \end{subfigure}
  \begin{subfigure}{0.32\textwidth}
  \centering
  \includegraphics[width=5cm]{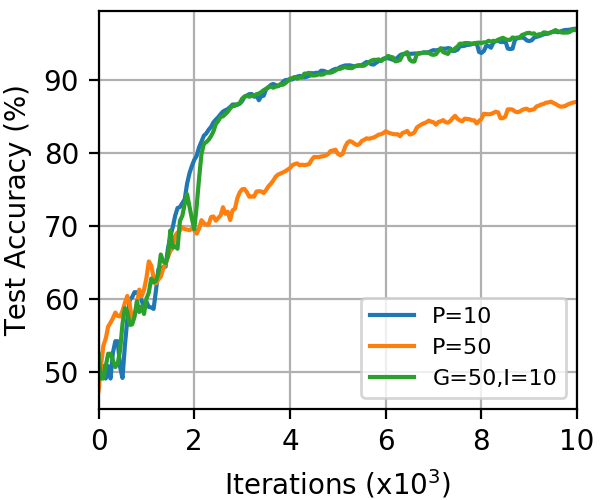}
  \caption{}
  \label{fig:celeba-test-iter}
  \end{subfigure}
  \caption{Results with CelebA; (a), (b) training loss v.s. communication time and iterations, (c) test accuracy v.s. iterations.} 
  \label{fig: celeba-supp-results}
\end{figure}

\begin{figure}[H]
  \centering
    \begin{subfigure}{0.49\textwidth}
  \centering
  \includegraphics[width=5cm]{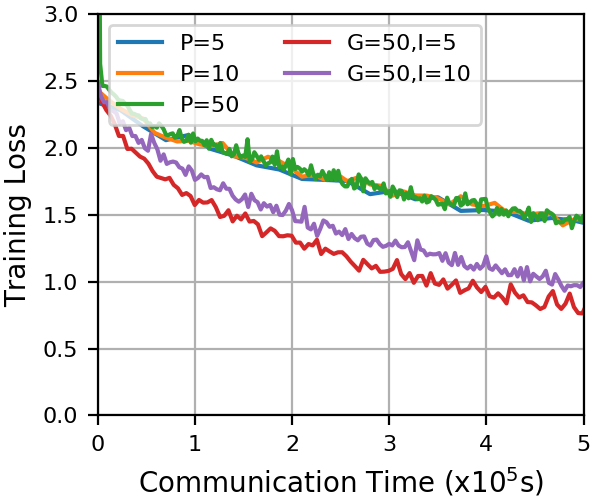}
  \caption{Non-IID with comm. time.}
  \label{fig:com-training}
  \end{subfigure}
  \begin{subfigure}{0.49\textwidth}
  \centering
  \includegraphics[width=5cm]{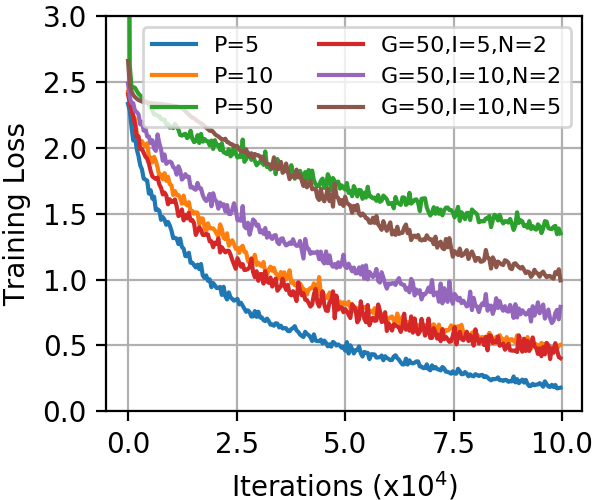}
  \caption{Non-IID case.}
  \label{fig:nIID-training}
  \end{subfigure}\\
  \begin{subfigure}{0.49\textwidth}
  \centering
  \includegraphics[width=5cm]{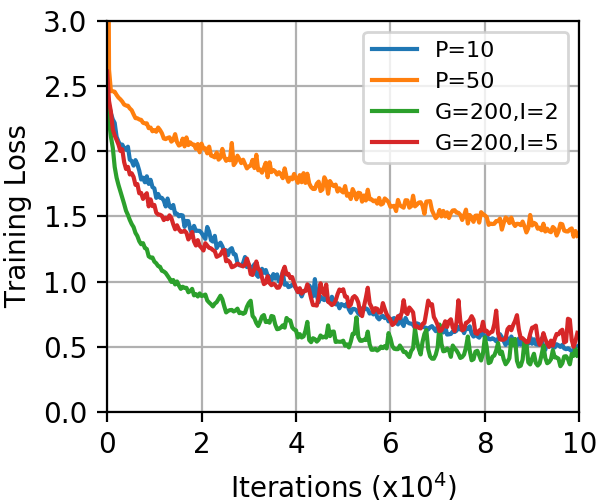}
  \caption{Non-IID with a larger $G$.}
  \label{fig:largeG-training}
  \end{subfigure}
  \begin{subfigure}{0.49\textwidth}
  \centering
  \includegraphics[width=5cm]{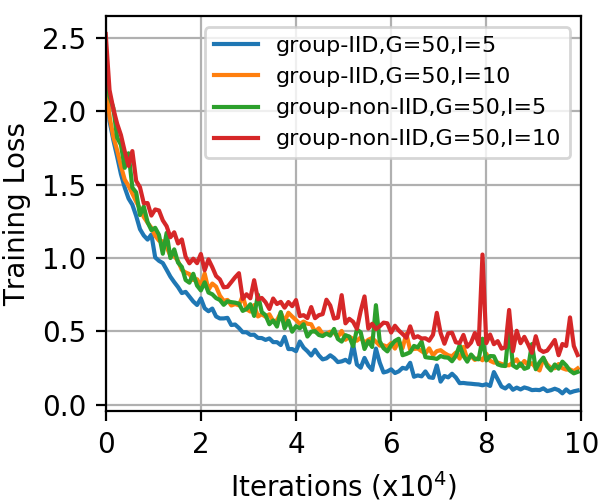}
  \caption{Group-IID/Group-Non-IID.}
  \label{fig:gIID-training}
  \end{subfigure}
  \caption{Results with CIFAR-10. Training loss v.s.  (a) communication time, (b)--(d) local iterations. 
  By default, $N=2$.  } 
  \label{fig: cifar 10-training}
\end{figure}

In this section, we will provide all the training loss results and results based on iterations for FEMNIST and CelebA datasets of experiments in Section~\ref{sec:experiments}, as shown in Figures~\ref{fig:femnist-supp-results}, \ref{fig: celeba-supp-results}, and \ref{fig: cifar 10-training}. It can be observed that the behavior of the training loss results are consistent with that of test accuracy shown in Section~\ref{sec:experiments}. We summarize the final test accuracy in Table~\ref{tab: variance}.

\begin{table}[H]
    \centering
        \caption{Final test accuracy (\%) for two-level cases.}
    \begin{tabular}{c|c|c|c|c|c}
    \hline
         \multirow{2}*{CIFAR-10}&$P=5$&$P=10$&$P=50$&$G=50,I=5$&$G=50,I=10$\\
    \cline{2-6}
      ~&$82.3\pm 0.5$  &$77.9\pm 1.2$ & $50.5\pm 1.8$ &$78.5\pm 0.4$& $69.6\pm 1.5$ \\
    \hline
      \multirow{2}*{FEMNIST}&$P=10$&$P=50$&$P=100$&$G=100,I=10$&$G=100,I=20$\\
     \cline{2-6}
      ~&$83.5\pm 0.3$  &$80.6\pm 1.1$ & $79.6\pm 1.6$ &$82.6\pm 0.6$& $81.7 \pm 1.1$ \\
    \hline
      \multirow{2}*{CelebA}&$P=10$&$P=50$&$G=50,I=10$\\
    \cline{2-4}
      ~&$96.9 \pm 0.2$  &$86.9\pm 0.5$ & $96.8\pm 0.3$  \\
    \hline
    \end{tabular}

    \label{tab: variance}
\end{table}


\begin{figure}[H]
  \begin{subfigure}{0.49\textwidth}
  \centering
  \includegraphics[width=5cm]{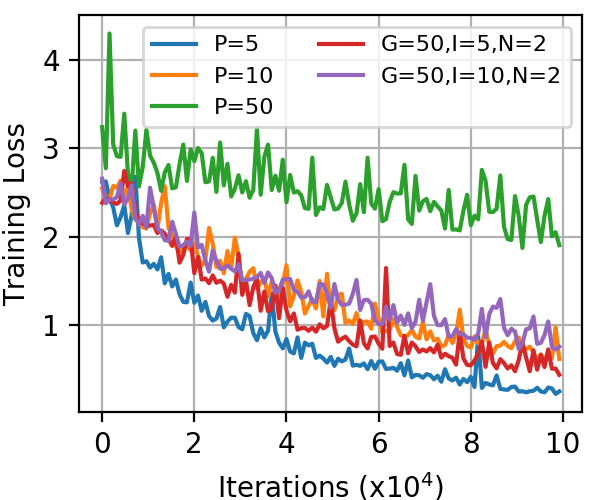}
  \caption{Training loss v.s. iterations. 
  }
  \label{fig:client-selection-1}
  \end{subfigure}
  \begin{subfigure}{0.49\textwidth}
  \centering
  \includegraphics[width=5cm]{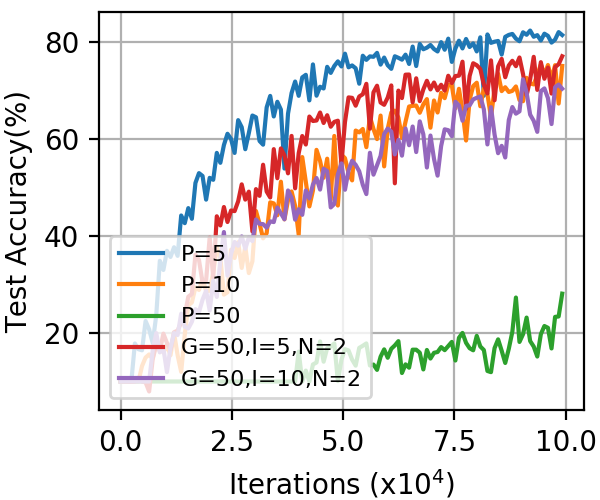}
  \caption{Test accuracy v.s. iterations.}
  \label{fig:client-selection-1-test}
  \end{subfigure}\\
  \begin{subfigure}{0.49\textwidth}
  \centering
  \includegraphics[width=5cm]{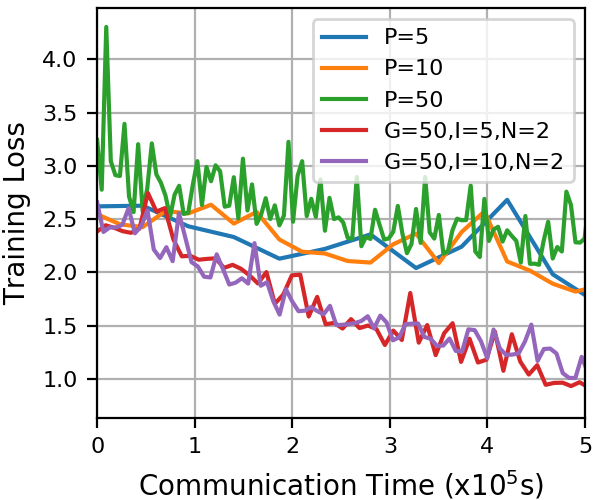}
  \caption{Training loss v.s. communication time. }
  \label{fig:client-selection-time}
  \end{subfigure}
  \begin{subfigure}{0.49\textwidth}
  \centering
  \includegraphics[width=5cm]{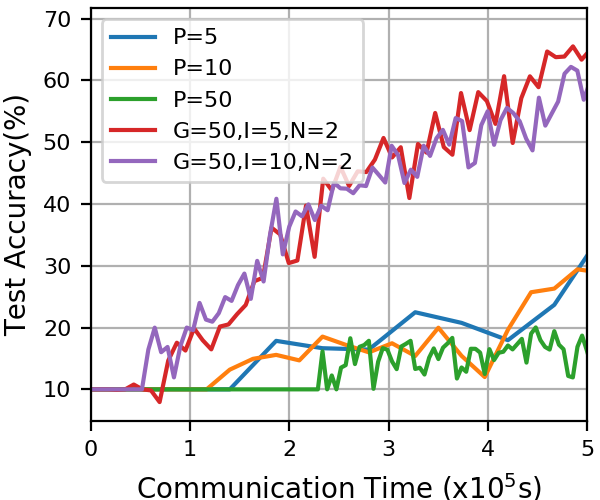}
  \caption{Test accuracy v.s. communication time.}
  \label{fig:client-selection-time-test}
  \end{subfigure}
  \caption{Comparison between local SGD and H-SGD.  
  }
  \label{fig:D7}
\end{figure}

\begin{figure}
  \begin{subfigure}{0.5\textwidth}
  \centering
  \includegraphics[width=5cm]{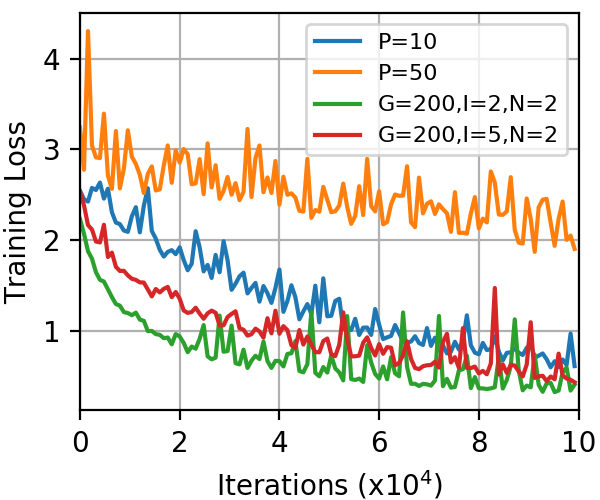}
  \caption{Training loss.}
  \label{fig:client-selection-2}
  \end{subfigure}
  \begin{subfigure}{0.5\textwidth}
  \centering
  \includegraphics[width=5cm]{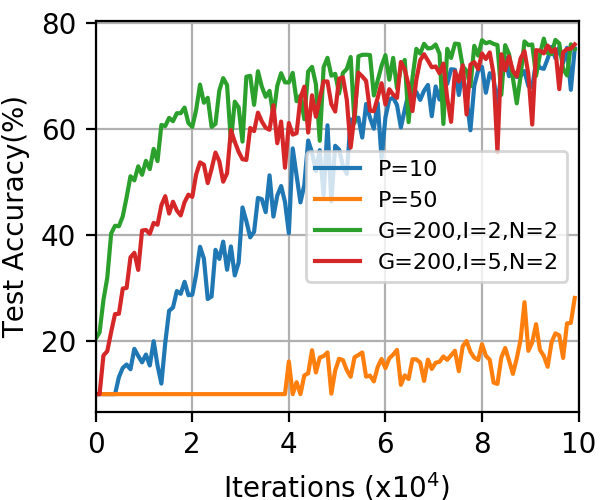}
  \caption{Test accuracy.}
  \label{fig:client-selection-2-test}
  \end{subfigure}
  \caption{Comparison between SF-SGD and HF-SGD with a large $G$.  
  }
\end{figure}

\begin{figure}[H]
  \begin{subfigure}{0.5\textwidth}
  \centering
  \includegraphics[width=5cm]{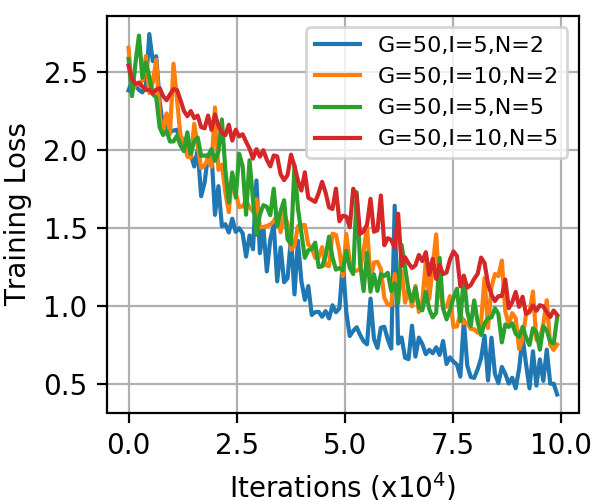}
  \caption{Training loss.}
  \label{fig:client-selection-3}
  \end{subfigure}
  \begin{subfigure}{0.5\textwidth}
  \centering
  \includegraphics[width=5cm]{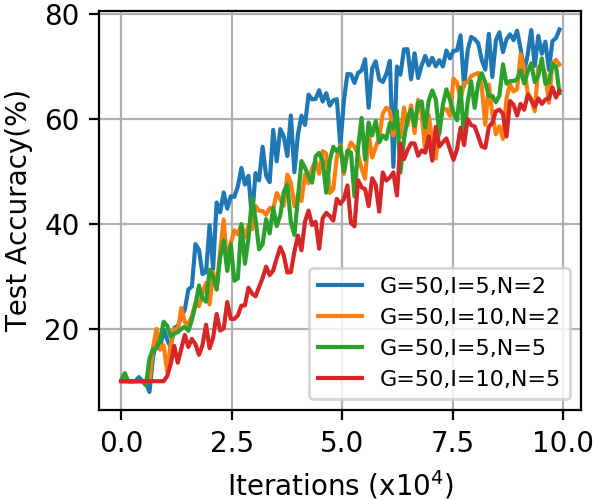}
  \caption{Test accuracy.}
  \label{fig:client-selection-3-test}
  \end{subfigure}
  \caption{Performance comparison for different group sizes.  }
  \label{fig:D9}
\end{figure}

Further, we provide experiments with CIFAR-10 dataset for partial worker participation in two-level HF-SGD with random worker sampling in each round, as shown in Figures~\ref{fig:D7}--\ref{fig:D9}. We separate data into 50 workers in a non-IID manner. There are the same number of workers in each group. For each HG-SGD round, we uniformly sample $20\%$ of workers in each group. The results show that the same insights as described in Section~\ref{sec:experiments} of the main paper can be observed here as well.

\subsection{Multi-level Experiments with
(\texorpdfstring{$M=3$}{M = 3}) }
\begin{figure}[H]
    \centering
    \includegraphics[width=10cm]{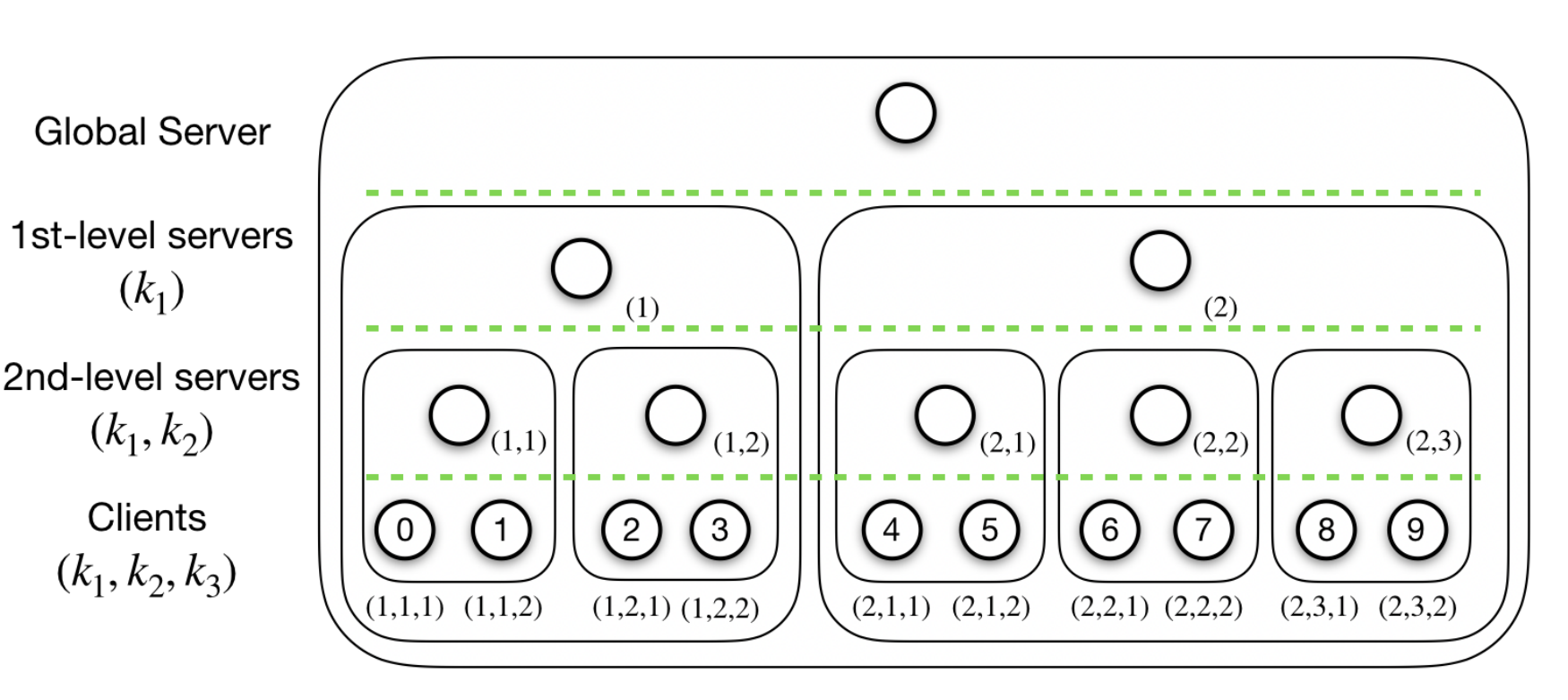}
    \caption{The 3-level structure we used in experiments. The number inside each circle representing the worker is the label available at this worker.}
    \label{fig:3level-structure}
\end{figure}

In order to show the advantage of the multi-level HF-SGD and demonstrate the effectiveness of our analysis shown in Theorem~\ref{thm3}, in this section, we provide experimental results of the 3-level case using the CIFAR-10 dataset. We construct the 3-level system shown in Figure~\ref{fig:3level-structure}, where for example, server $(1,1)$ in level 2 computes the local average of SGD from workers $(1,1,1)$ and $(1,1,2)$, and  server $(1)$ in level 1 computes the average of SGD from servers $(1,1)$ and $(1,2)$ in level 2. Here, 
the CIFAR-10 dataset is partitioned into 10 workers such that each worker has data of one class. For example, the worker $(1,1,1)$ has the data of label 0. 

\begin{figure}[H]
  \begin{subfigure}{0.5\textwidth}
  \centering
  \includegraphics[width=7cm]{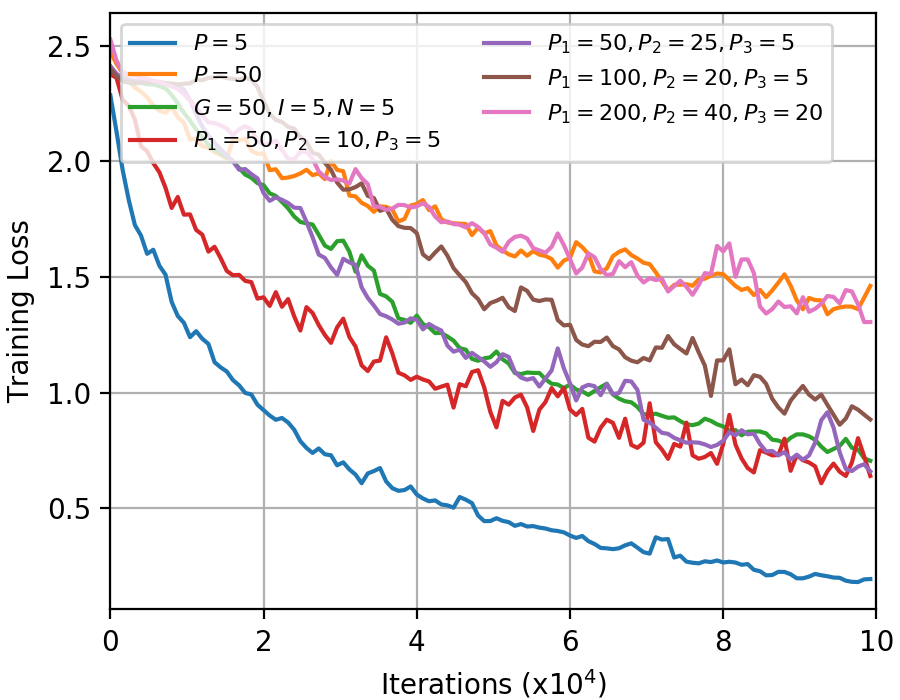}
  \caption{Training loss.}
  \label{fig:3-level-training}
  \end{subfigure}
  \begin{subfigure}{0.5\textwidth}
  \centering
  \includegraphics[width=7cm]{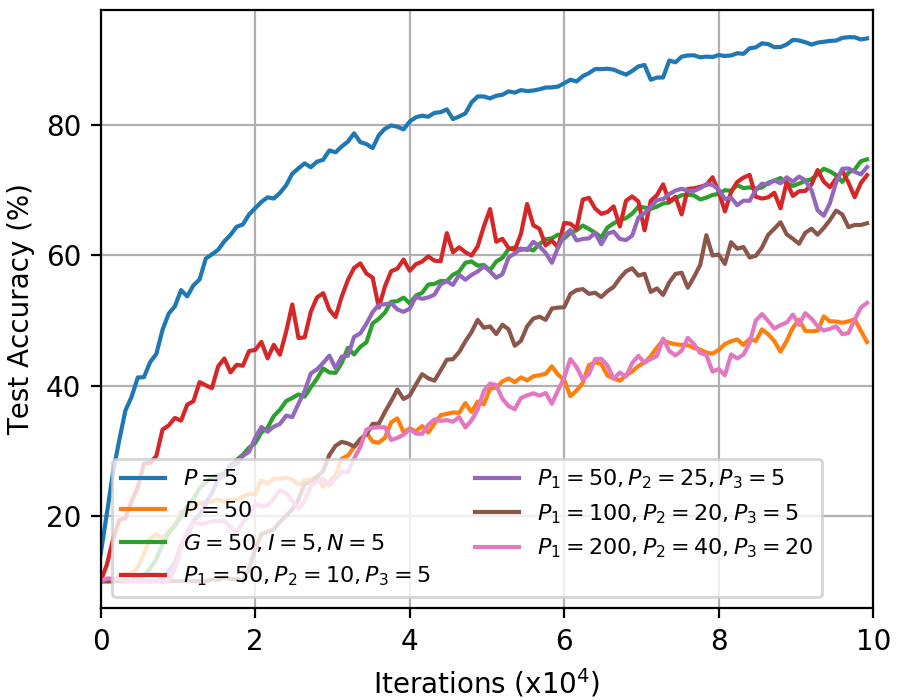}
  \caption{Test accuracy.}
  \label{fig:3-level-test}
  \end{subfigure}
  \caption{3-level experiments v.s. iterations. For 3-level case, we set $N_1=2,N_2=5$ by default. 
  }
  \label{fig:3-level-exp-iteration}
\end{figure}
From Figure~\ref{fig:3-level-exp-iteration}, first, we can observe that second level aggregation helps. For example, compare the 2-level case with $G=50,I=5$ (green curve) to the 3-level cases with  $P_1=50,P_2=10,P_3=5$ (red curve) and $P_1=50,P_2=25,P_3=5$ (purple curve), more second level aggregation (smaller $P_2$) can improve the performance. Note that the two-level case with $G,I$ is equivalent to three-level case with $P_1 = P_2 = G$, $P_3=I$.  Second, compared with single-layer case, 3-level cases also shows the same ``sandwich" behavior as in 2-level cases, e.g., the red curve ($P_1=50,P_2=10,P_3=5$) is between the blue curve ($P=5$) and the orange curve ($P=50$) in Figure~\ref{fig:3-level-exp-iteration}. Third, it can be seen that although the case with $P_1=100,P_2=20,P_3=5$ has a larger global period than single-layer case $P=50$, its performance (brown curve) is still better than single-layer case (orange curve). Similar behavior can be observed even if we increase $P_1,P_2,P_3$ at the same time for the case with $P_1=200,P_2=40,P_3=20$ (magenta curve). %
This implies that, by doing aggregations at different levels with a larger global period, we can reduce the communication time while keeping the same performance.

\begin{figure}[H]
  \begin{subfigure}{0.5\textwidth}
  \centering
  \includegraphics[width=7cm]{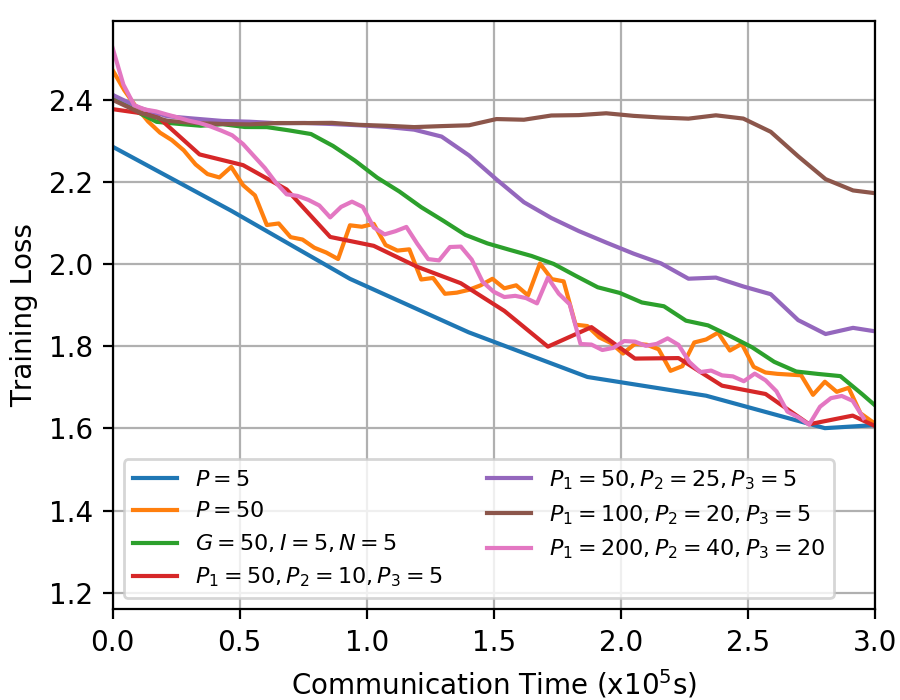}
  \caption{Training loss.}
  \label{fig:3-level-training-time}
  \end{subfigure}
  \begin{subfigure}{0.5\textwidth}
  \centering
  \includegraphics[width=7cm]{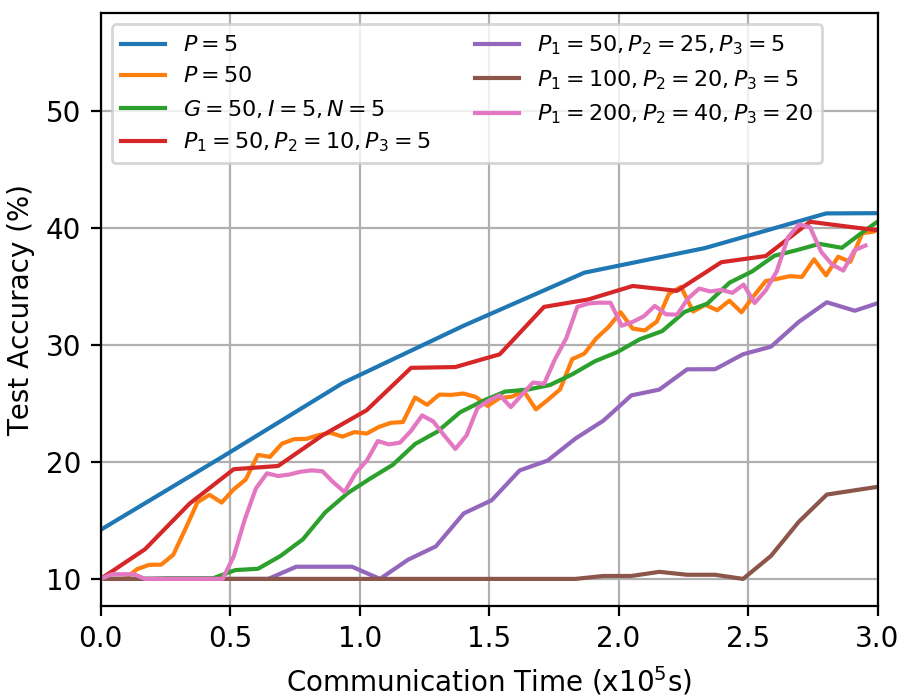}
  \caption{Test accuracy.}
  \label{fig:3-level-test-time}
  \end{subfigure}
  \caption{3-level experiments v.s. communication time. For 3-level case, we set $N_1=2,N_2=5$ by default.  
  }
  \label{fig:3-level-exp-time}
\end{figure}

In Figure~\ref{fig:3-level-exp-time}, we plot the training loss/test accuracy of 3-level cases v.s. communication time. %
In this experiment, we use the communication time measured for VGG-11 model shown in Table~\ref{tab:actual-time}. Here we use the local EC2 time in Table~\ref{tab:actual-time} as communication time needed for aggregation by second-level servers. The global EC2 time in Table~\ref{tab:actual-time} is used as communication time needed for aggregation by the global server. We set the ratio of the communication times between the first-level servers to the workers and second-level to the workers to be $2:1$.\footnote{Note that if the communication time of the second-level server to the worker is too long, then this additional level of aggregation is not needed.}
We can see from the figure that the cases of $P_1=50,P_2=10,P_3=5$ and $P_1=200,P_2=40,P_3=20$ can achieve better performance compared to others under the same total communication time.

\clearpage

\renewcommand\bibname{Additional References for Appendix}
\renewcommand\refname{Additional References for Appendix}

\end{document}